\documentclass[a4paper,10pt,oneside,fleqn,review]{cas-dc}
\usepackage{cas-common}
\usepackage{graphicx}
\usepackage{svg}
\usepackage{lineno}
\usepackage{setspace}
\usepackage[authoryear]{natbib}
\usepackage{subcaption}
\usepackage{caption}
\usepackage{placeins}
\usepackage{float}
\usepackage{hyperref}

\def\tsc#1{\csdef{#1}{\textsc{\lowercase{#1}}\xspace}}
\tsc{WGM}
\tsc{QE}
\begin{document}
\let\WriteBookmarks\relax
\def\floatpagepagefraction{1}
\def\textpagefraction{.001}

% Short title
\shorttitle{Asymptomatic Sugarcane Disease Detection With Multispectral Satellite Imaging}    

% Short author
\shortauthors{Ethan Kane Waters, Carla Chia-ming Chen and Mostafa Rahimi Azghadi}  

% Main title of the paper
\title [mode = title]{Machine Learning for Asymptomatic Ratoon Stunting Disease Detection With Freely Available Satellite Based Multispectral Imaging}  

\author[1,2,3]{Ethan Kane Waters}[orcid=0000-0003-0624-4877]
\cormark[1]
\ead{ethan.waters@dpi.qld.gov.au}
\credit{Conceptualization, Methodology, Formal analysis, Investigation, Software, Data Curation, Visualization, Project administration, Funding acquisition, Writing – original draft, Writing – review \& editing}

% Address/affiliation
\affiliation[1]{organization={College of Science and Engineering},
            addressline={James Cook University}, 
            city={Townsville},
            postcode={4818}, 
            state={QLD},
            country={Australia}}

\affiliation[2]{organization={Agriculture Technology and Adoption Centre},
            addressline={James Cook University}, 
            city={Townsville},
            postcode={4814}, 
            state={QLD},
            country={Australia}}

\affiliation[3]{organization={Department Of Primary Industries},
            addressline={203 Tor Street}, 
            city={Toowoomba},
            postcode={4350}, 
            state={QLD},
            country={Australia}}

\author[1,2]{Carla Chia-Ming Chen}[orcid=0000-0001-9718-4464]
\ead{carla.ewels@jcu.edu.au}
\credit{Conceptualization, Methodology, Supervision,  Writing – review \& editing}

\author[1,2]{Mostafa {Rahimi Azghadi}}[orcid=0000-0001-7975-3985]
\ead{mostafa.rahimiazghadi@jcu.edu.au}
 \credit{Conceptualization, Methodology, Supervision, Project administration, Funding acquisition, Writing – review \& editing}

\cortext[1]{Corresponding author}

\begin{abstract}
Disease detection in sugarcane, particularly the identification of asymptomatic infectious diseases such as Ratoon Stunting Disease (RSD), is critical for effective crop management. This study employed various machine learning techniques to detect the presence of RSD in different sugarcane varieties, using vegetation indices derived from freely available satellite-based spectral data. Our results show that the Support Vector Machine with a Radial Basis Function Kernel (SVM-RBF) was the most effective algorithm, achieving classification accuracy between 85.64\% and 96.55\%, depending on the variety. Gradient Boosting and Random Forest also demonstrated high performance achieving accuracy between 83.33\% to 96.55\%, while Logistic Regression and Quadratic Discriminant Analysis showed variable results across different varieties. The inclusion of sugarcane variety and vegetation indices was important in the detection of RSD. This agreed with what was identified in the current literature. Our study highlights the potential of satellite-based remote sensing as a cost-effective and efficient method for large-scale sugarcane disease detection alternative to traditional manual laboratory testing methods.
\end{abstract}

% Keywords
% Each keyword is seperated by \sep
% Sugarcane Health Monitoring; Remote Sensing; Satellite-based Spectroscopy; Machine Learning; Vegetation Indices; Disease Detection

\maketitle

\section{Introduction}
Disease presents a formidable hurdle to maximising yield in the sugarcane industry with Ratoon stunting disease (RSD) emerging as a key contributor globally \citep{RSD3}. The bacterium \emph{Leifsonia xyli subsp. xyli (Lxx)} is the primary causal agent of RSD, which is primarily propagated with contaminated cutting implements \citep{RN474}. RSD infection can result in substantial yield losses (up to 60\%), depending on sugarcane variety and water availability during growth periods \citep{RN11, RSD3, RSD2, RSD1}. This results in pronounced economic repercussions, with an estimated annual loss of \$25 million observed across 87,000 hectares monitored in Australia in 2019\citep{RN12}. The lack of external symptoms \citep{RSD3, RSD2, RSD1, RSD4} has made the detection and the management of RSD a significant challenge. It relies heavily on laboratory diagnostic techniques such as Polymerase Chain Reaction (PCR), Quantitative Polymerase Chain Reaction (qPCR),  Loop-Mediated Isothermal Amplification (LAMP), or Leaf Sheet Biopsy Quantitative Polymerase Chain Reaction (LSB-qPCR) for RSD detection \citep{RN477, RN478, RN476, RN475}. Collecting the field samples required for these techniques is both time-consuming and costly, especially for large-area detection. This underscores the need to develop a more efficient method for large-scale detection of RSD.

Previous research has demonstrated success in using spectroscopy to detect disease and pests in sugarcane by leveraging subtle differences often imperceptible to the naked eye \citep{ RN479}. Spectroscopy measures the electromagnetic radiation reflected from an object, providing insights into its chemical composition, molecular structure, and physical properties \citep{RN560} that may indicate the presence of disease. However, most of the studies in the literature utilised handheld spectrometers \citep{Bao2, RN15, RN225, RN479, RN93, RN96, RN216}, which are challenging to scale up for large field detection. Hence, recent studies have shifted towards drone or satellite-based observations \citep{RN92, RN18, RN20, RN224, RN95, RN213}, acknowledging the need for larger-scale diagnostic methods. Despite this, no studies to date have yet utilised satellite-based multispectral imaging for disease detection in sugarcane. Although three previous studies have explored the diagnosis of asymptomatic sugarcane diseases with machine learning and spectroscopy, both relied on handheld spectrometers \citep{RN479, Bao2, RN15}. No current research has applied large-scale remote sensing techniques to detect asymptomatic diseases in sugarcane.

\begin{table*}[H]
\fontsize{8}{8}\selectfont
\caption{Current literature for disease and pest detection in sugarcane with ML and spectroscopy.\label{tab_lit}}
\begin{tabular}{p{3.75cm}p{3cm}p{2cm}p{1.75cm}p{2cm}p{2.5cm}}
\hline\hline
\textbf{Reference} & \textbf{Health Condition}	& \textbf{Spectroscopy} & \textbf{Category} & \textbf{ML Algorithm} & \textbf{Best Classification Accuracy}  \\
\hline
\citet{RN18, RN19} & Orange Rust & Hyperspectral & Satellite & LDA & 96.90\% \\
\midrule
\citet{RN92} & Mosaic & Hyperspectral & Drone & SID & 92.50\% \\
\midrule
\citet{RN15} & SCYLV & Hyperspectral & Handheld Spectrometer & LDA & 73\% \\
\midrule
\citet{RN20} & White Leaf Disease & Multispectral & Drone & RF, DT, KNN, XGB & 92\%, 91\%, 92\%, 92\% \\
\midrule
\citet{RN224} & Orange \& Brown Rust & Multispectral & Drone & RF, KNN, SVM  & 90\%, 90\%, 90\%; 86\%, 83\%, 88\%\\
\midrule
\citet{RN225} & Brown Stripe \& Ring Spot & Hyperspectral & Handheld Spectrometer &  RF, SVM, NB & 95\%, 85\%, 77\% \\
\midrule
\citet{RN95} & Cane Grub & Multispectral & Satellite & GEOBIA & 79\% \\
\midrule
\citet{RN213} & Cane Grub & Multispectral & Satellite & GEOBIA & 98.7\% \\
\midrule
\citet{RN479, Bao2} & Smut \& Mosaic & Hyperspectral & Handheld Spectrometer & CNN & >90\% \\
\midrule
\citet{RN93} & Diatraea saccharalis & Hyperspectral \& Multispectral & Handheld Spectrometer \& Satellite & N/A & 79.8\% \& 85.5\% \\
\midrule
\citet{RN96} & Thrips & Hyperspectral & Handheld Spectrometer & N/A & N/A \\
\midrule
\citet{RN216} & Orange \& Brown Rust & Hyperspectral \& Multispectral & Handheld Spectrometer \& Drone & N/A & N/A \\
\hline\hline
\end{tabular}
\end{table*}

Freely available satellite-based remote sensing offers a cost-effective and efficient alternative to the traditional, resource-intensive methods of identifying and managing disease in sugarcane \citep{Waters}. In particular, free publicly accessible multispectral satellite data can alleviate the financial burden of purchasing expensive spectrometers images, and promote the widespread adoption of this advanced technology. 

Spectroscopy data can be converted into a 'Vegetation Index' to accentuate the important characteristics of vegetation for the specific task \citep{RN45, RN71}. The suitability of a vegetation index for a particular application depends on both the spectrometer and the project objective \citep{RN45}. The majority of the sugarcane disease and pest detection studies \citep{RN20, RN18, RN224, RN93, RN216, RN95, RN213} leverage vegetation indices to improve performance. This was particularly prominent in the study by \citet{RN18} which developed several new vegetation indices for sugarcane disease detection with a focus on water and vegetation stress.

The application of machine learning (ML) algorithms to spectroscopy data has emerged as a promising approach to aid disease management in sugarcane (Table \ref{tab_lit}). A wide range of algorithms has been employed in this field, including XGBoost (XGB), Random Forest (RF), Decision Trees (DT), K-Nearest Neighbors (KNN), Support Vector Machines (SVM), Linear Discriminant Analysis (LDA), and Neural Networks (NN) \citep{RN20, RN15, RN479, RN18, RN19, RN20, RN225, RN224}. Despite the diversity in approaches, only a handful of studies have systematically compared ML performance \citet{RN20, RN225, RN224}. Among the algorithms, RF is the most common and consistently achieves high classification accuracy \citet{RN20, RN225, RN224}. Similarly, SVM has demonstrated strong performance, achieving promising results \citet{RN20, RN224}. However, despite the variety of methods applied, the limited exploration of each method is insufficient to draw robust conclusions about the relative effectiveness of these algorithms. This highlights the need for continued comprehensive comparisons across multiple datasets and conditions to better understand their relative effectiveness in sugarcane disease management. Additionally, few studies have accounted for the impact of different sugarcane varieties on model performance, warranting further exploration given the increased variation introduced by classifying several varieties \citet{RN15, RN96, RN224}.

The primary objective of this research is to evaluate the effectiveness of various machine learning algorithms in classifying asymptomatic RSD in multiple varieties of sugarcane with freely available multispectral satellite data. Additionally, the study seeks to identify which vegetation indices are most important for diagnosing RSD. To the best of our knowledge, this study represents the first effort to classify an asymptomatic disease in sugarcane with satellite-based spectroscopy and is the first instance of diagnosing RSD with machine learning and spectroscopy \citep{Waters}.

\section{Methodology}
\subsection{Overall Process Summary}
The project comprises three stages: ground-truth data collection, data preprocessing, and ML development. 
In stage one, RSD data is collected from sugarcane blocks. For stage two, sentinel-2 images with pre-applied atmospheric correction undergo preprocessing before pixelwise values are extracted and labeled from raw spectral bands and vegetation indices. In the third stage, the dataset is divided into an 80:20 train-test split, with 10-fold validation applied to the training set for hyperparameter tuning. The top-performing models are subsequently evaluated on the test set by performing bootstrapping with 5000 samples. Permutation testing is then performed to ensure model performance is significantly better than a null distribution. The steps are detailed in the following sections. 

\subsection{Study Location \& Ground Truth Data Collection}
Sampling to develop the ground truth dataset for this study was conducted by trained field agronomists at Herbert Cane Productivity Services Limited (HCPSL) between February and March 2022 across 76 sugarcane blocks in the Herbert region of Queensland, Australia. The dataset details the RSD status and variety of the sugarcane in each of the 76 blocks, which are comprised of five sugarcane varieties, Q200, Q208, Q240, Q253 and SRA14.
 
\begin{figure}[ht]
    \centering
    \includegraphics[width=0.5\textwidth]{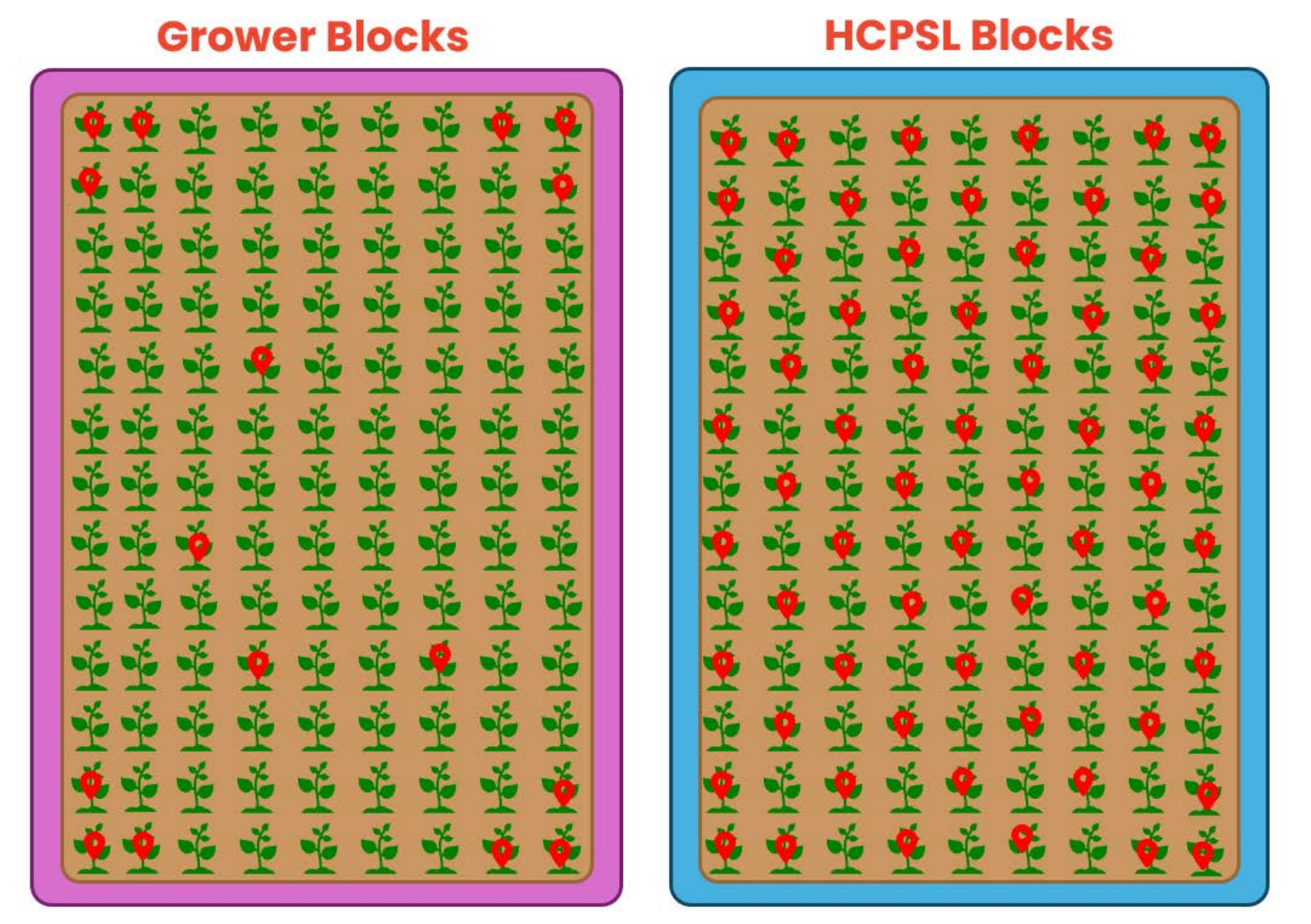} 
    \caption{Visual representation of the field sampling for both block types where the red marker indicates a sampling location.}
    \label{fig:sampling}
\end{figure}

Two types of sugarcane blocks were sampled; farmer-owned blocks and HCPSL-owned seed production blocks. HCPSL-owned seed production blocks exist to provide farmers the opportunity to purchase disease-free sugarcane seeds due to strict hygiene and disease testing requirements. HCPSL followed their standard RSD sampling protocol for both types of blocks. In seed production plots samples from a sugarcane stick was collected every 20m in a grid pattern. In Farmer-owned blocks, samples from twelve sugarcane sticks were collected from the four-corners of the block, and four from randomly selected locations within the block. The xylem extraction method was employed, in which juice from each sugarcane stick was extracted into a vial using pressure generated by an air compressor. The extracted juice from all samples was then frozen and subsequently analysed with qPCR testing at Sugar Research Australia (SRA) laboratories. The entire block is classified as RSD Positive if any sample within that block returned as RSD positive, and only labelled RSD Negative if all samples return a negative result. A visualisation of the field sampling method for the two block types is shown in Figure~\ref{fig:sampling}.

To establish reliable ground truth data for model training and validation, RSD negative samples were sourced exclusively from HCPSL seed cane blocks that returned negative qPCR test results, while RSD positive samples were sourced from grower blocks that returned positive qPCR test results. The more intensive sampling protocols implemented by HCPSL staff to meet strict disease control requirements led to different sampling densities between the two block types. However, this does not introduce bias into the dataset. The study labels data at a block level and consequently does not analyse within block variation which could be influenced by different sampling densities. Additionally, RSD Negative samples were only drawn from HCPSL seed blocks with the higher sampling density to increase confidence that these blocks were truly absent of RSD. This conservative approach to labelling negative samples is appropriate given the asymptomatic and contagious nature of RSD, where undetected infections pose a risk of mislabelling. In contrast, RSD positive samples were drawn only from grower blocks where qPCR testing confirmed disease presence. While these blocks were sampled less intensively, this does not compromise the validity of the positive labels, as only confirmed cases were included. It is possible that some infected blocks were missed due to lower sampling density and therefore excluded from the dataset entirely. Overall, the differing sampling strategies were intentionally structured to minimise label noise and enhance the reliability of both positive and negative class assignments.
 
A shapefile corresponding to the sampled blocks with the sugarcane variety and RSD status determined in the laboratory testing was produced by HCPSL GIS officer Rod Nielson. However, due to a nondisclosure agreement with the farmers, the shapefile and precise maps of the sampled areas cannot be published.

\subsection{Multispectral Satellite Image Acquisition and Data Labelling}
This study utilized the European Space Agency’s (ESA) Sentinel-2 satellite series, which offers freely available products for the wavelengths necessary to calculate the vegetation indices listed in Table \ref{tab2}. “Level-2A” products are provided for all required bands at a 20m resolution and have had atmospheric correction applied by ESA with the Sen2Cor process \citep{RN46}. To streamline data processing, a Python script was developed as a pipeline to extract all spectral bands from the “Level-2A” satellite image data products provided by Sentinel-2 via the “Sentinel Hub Process API.”

The multispectral data used in the study were captured by Sentinel-2 on 27th February 2022, as it had the least cloud coverage during the sampling period. Since RSD is transmitted through contaminated tools during planting or harvesting, and the study period was during the plant growth period, we assumed that the infection status remained unchanged during the study period. Only one day of image data was used in this study. 

\begin{figure}[htbp]
    \centering

    \begin{subfigure}[b]{0.15\textwidth}
        \includegraphics[width=\textwidth]{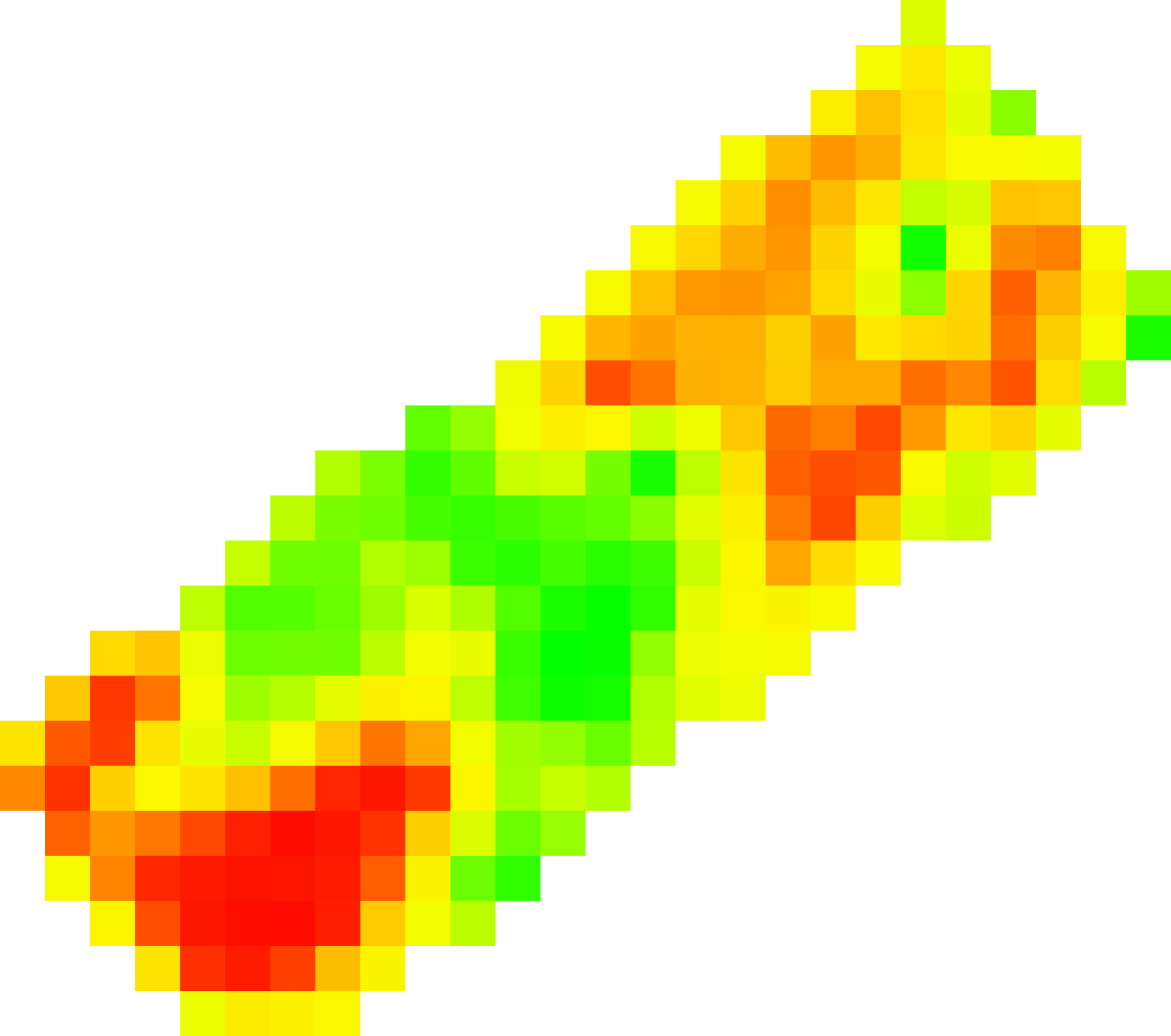}
        \caption{B02 (490 nm)}
        \label{fig:b02}
    \end{subfigure}
    \hspace{0.25em}
    \begin{subfigure}[b]{0.15\textwidth}
        \includegraphics[width=\textwidth]{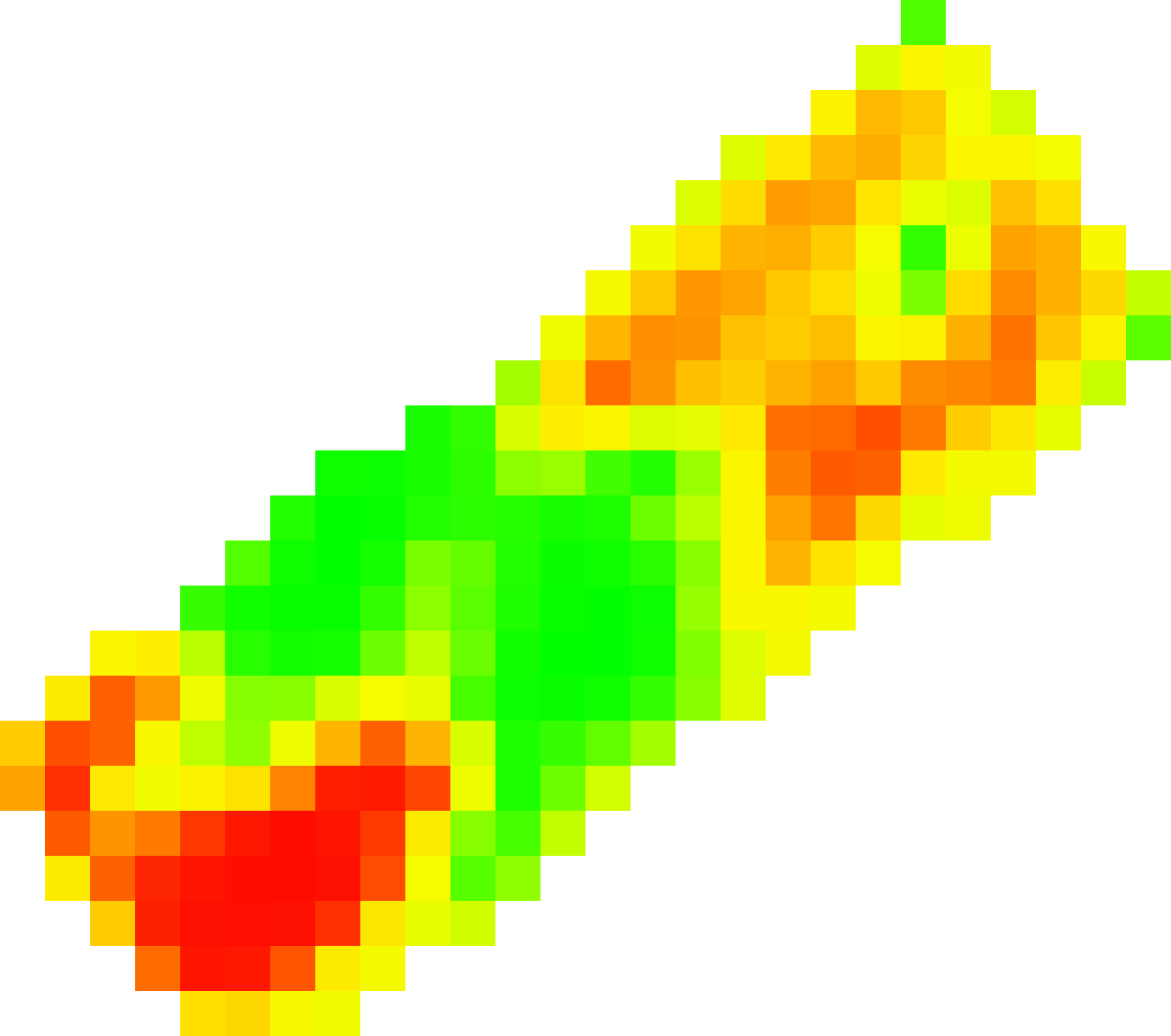}
        \caption{B03 (560 nm)}
        \label{fig:b03}
    \end{subfigure}
    \hspace{0.25em}
    \begin{subfigure}[b]{0.15\textwidth}
        \includegraphics[width=\textwidth]{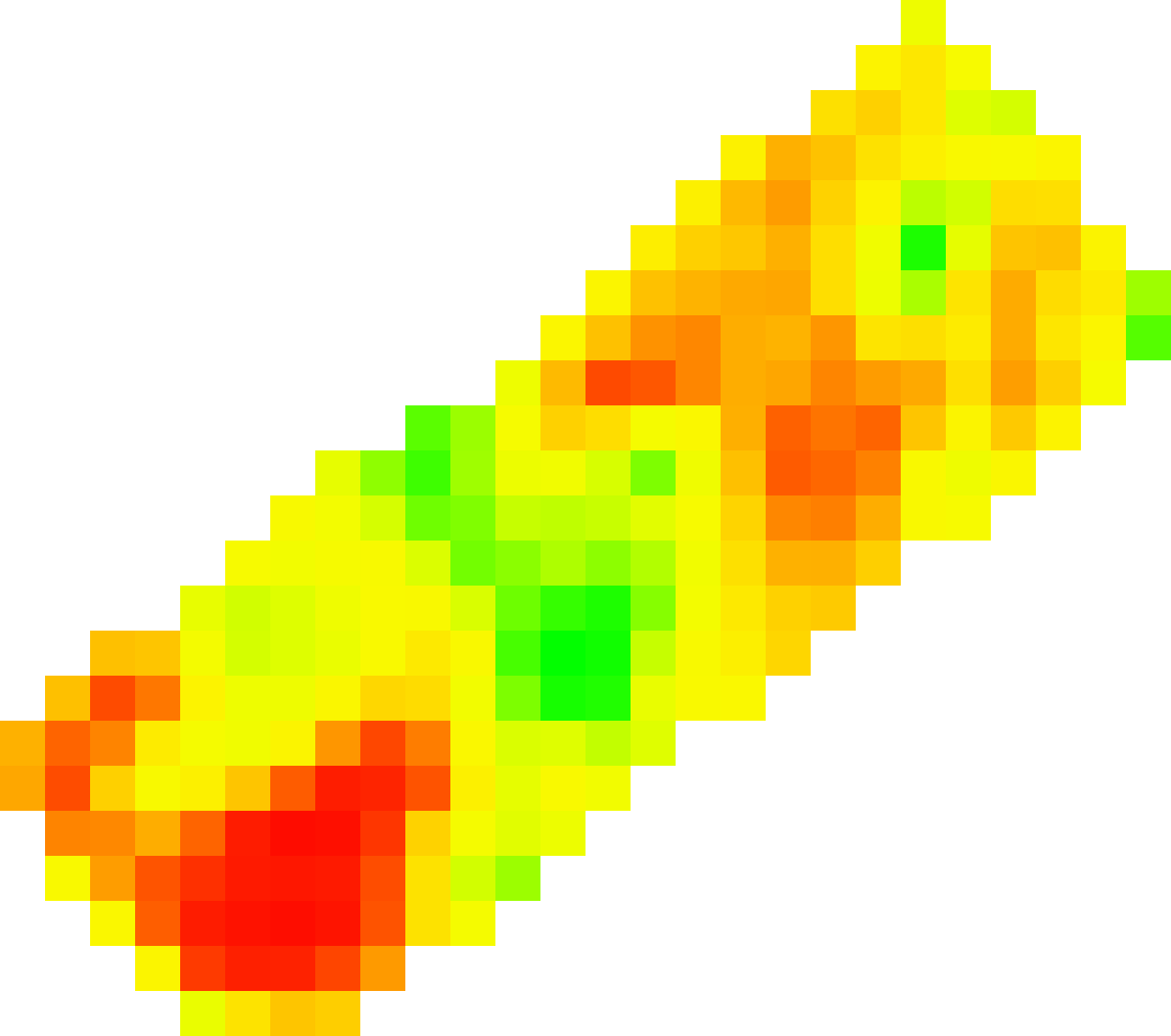}
        \caption{B04 (665 nm)}
        \label{fig:b04}
    \end{subfigure}
    
    \vspace{1em}
    
    \begin{subfigure}[b]{0.15\textwidth}
        \includegraphics[width=\textwidth]{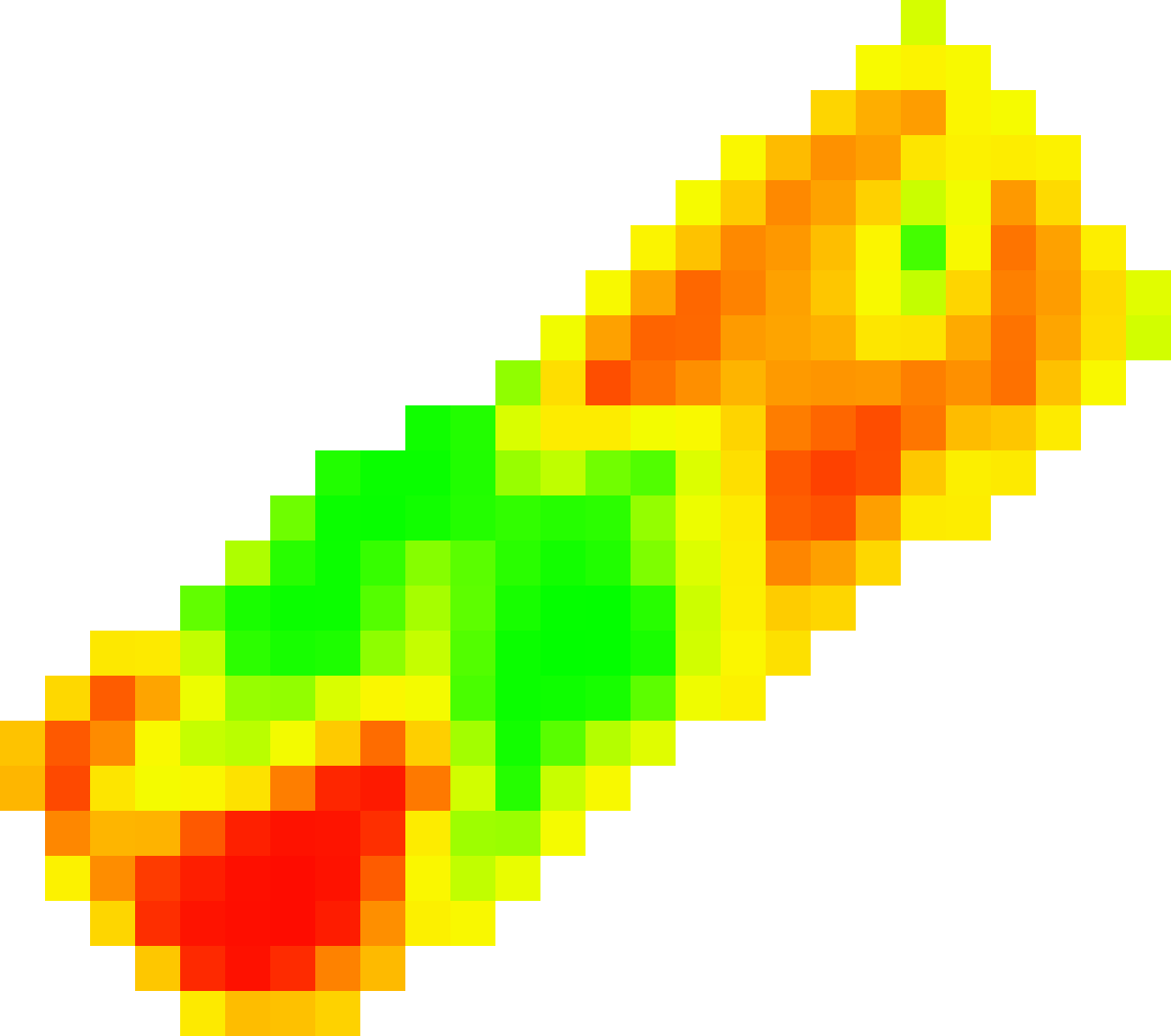}
        \caption{B05 (705 nm)}
        \label{fig:b05}
    \end{subfigure}
    \hspace{0.25em}
    \begin{subfigure}[b]{0.15\textwidth}
        \includegraphics[width=\textwidth]{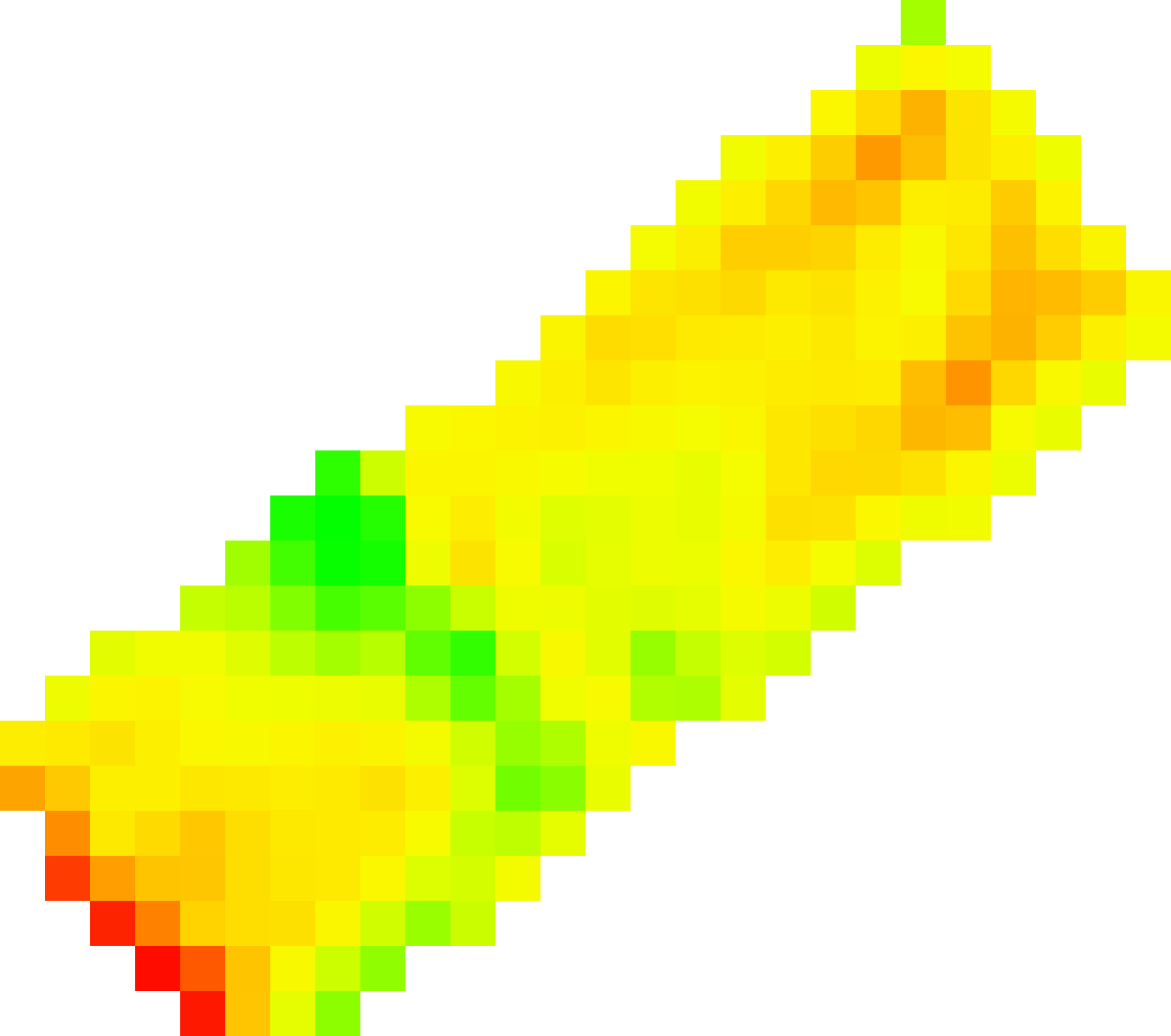}
        \caption{B06 (740 nm)}
        \label{fig:b06}
    \end{subfigure}
    \hspace{0.25em}
    \begin{subfigure}[b]{0.15\textwidth}
        \includegraphics[width=\textwidth]{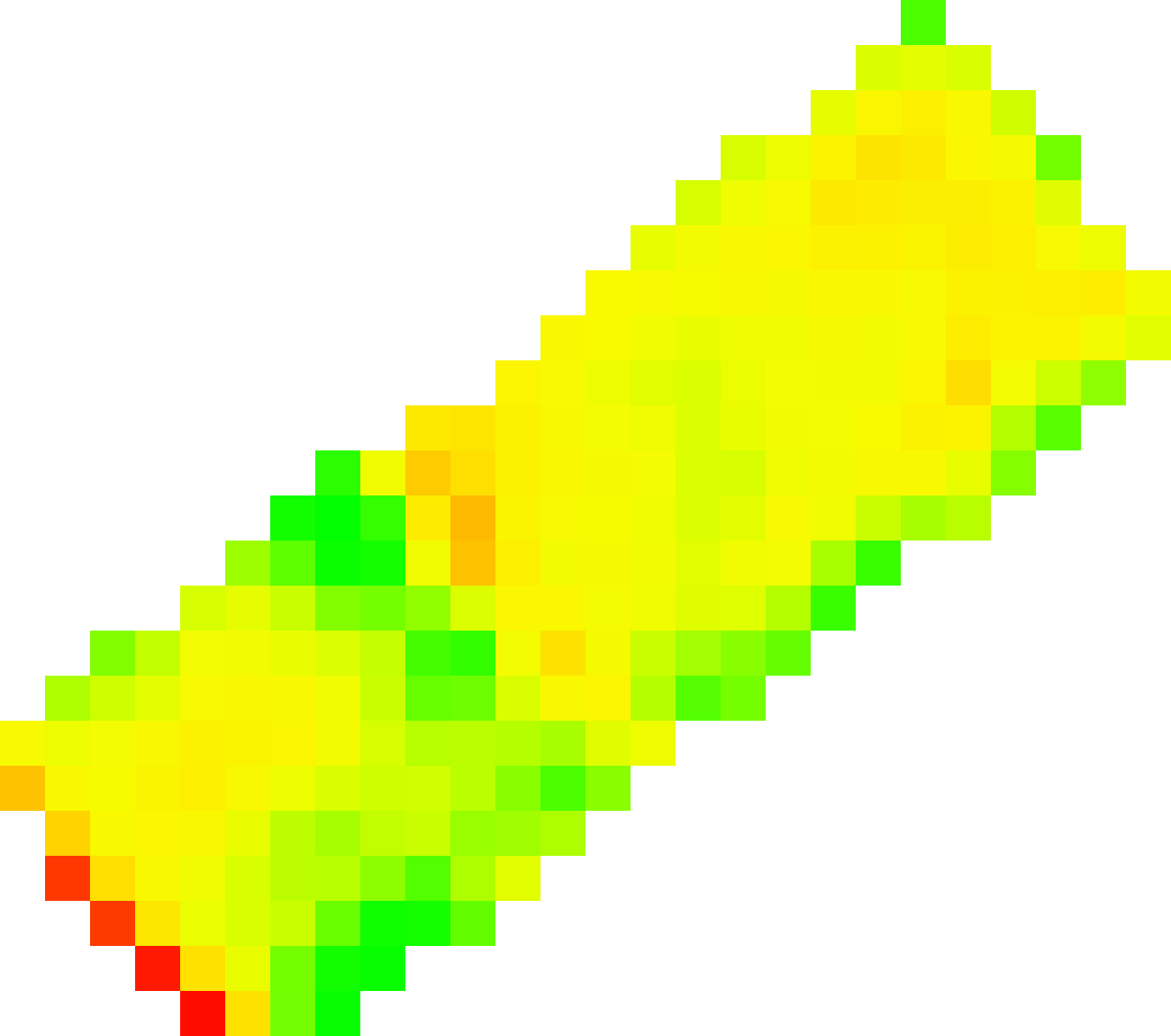}
        \caption{B07 (783 nm)}
        \label{fig:b07}
    \end{subfigure}
    
    \vspace{1em}
    
    \begin{subfigure}[b]{0.15\textwidth}
        \includegraphics[width=\textwidth]{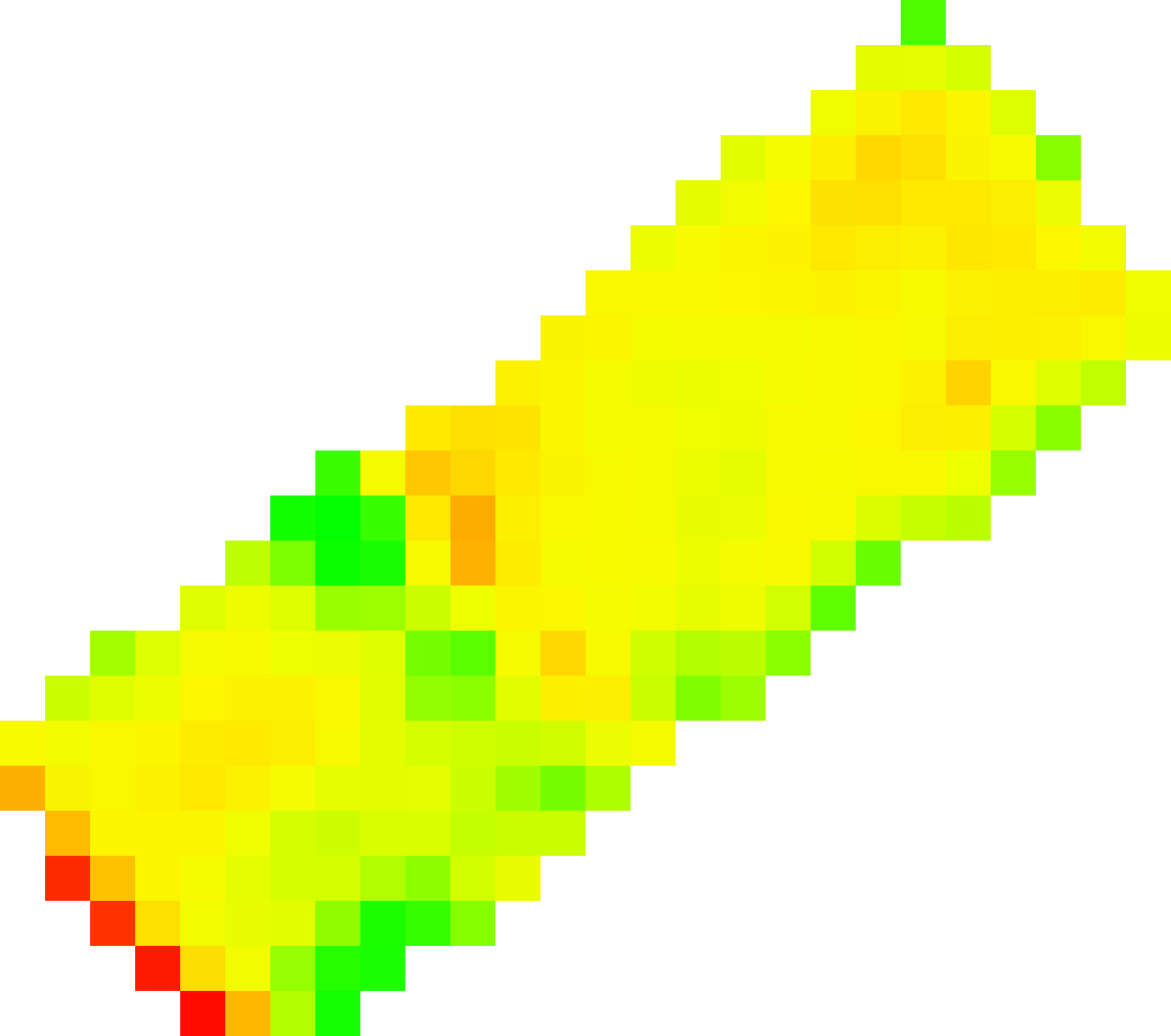}
        \caption{B8A (865 nm)}
        \label{fig:b8a}
    \end{subfigure}
    \hspace{0.25em}
    \begin{subfigure}[b]{0.15\textwidth}
        \includegraphics[width=\textwidth]{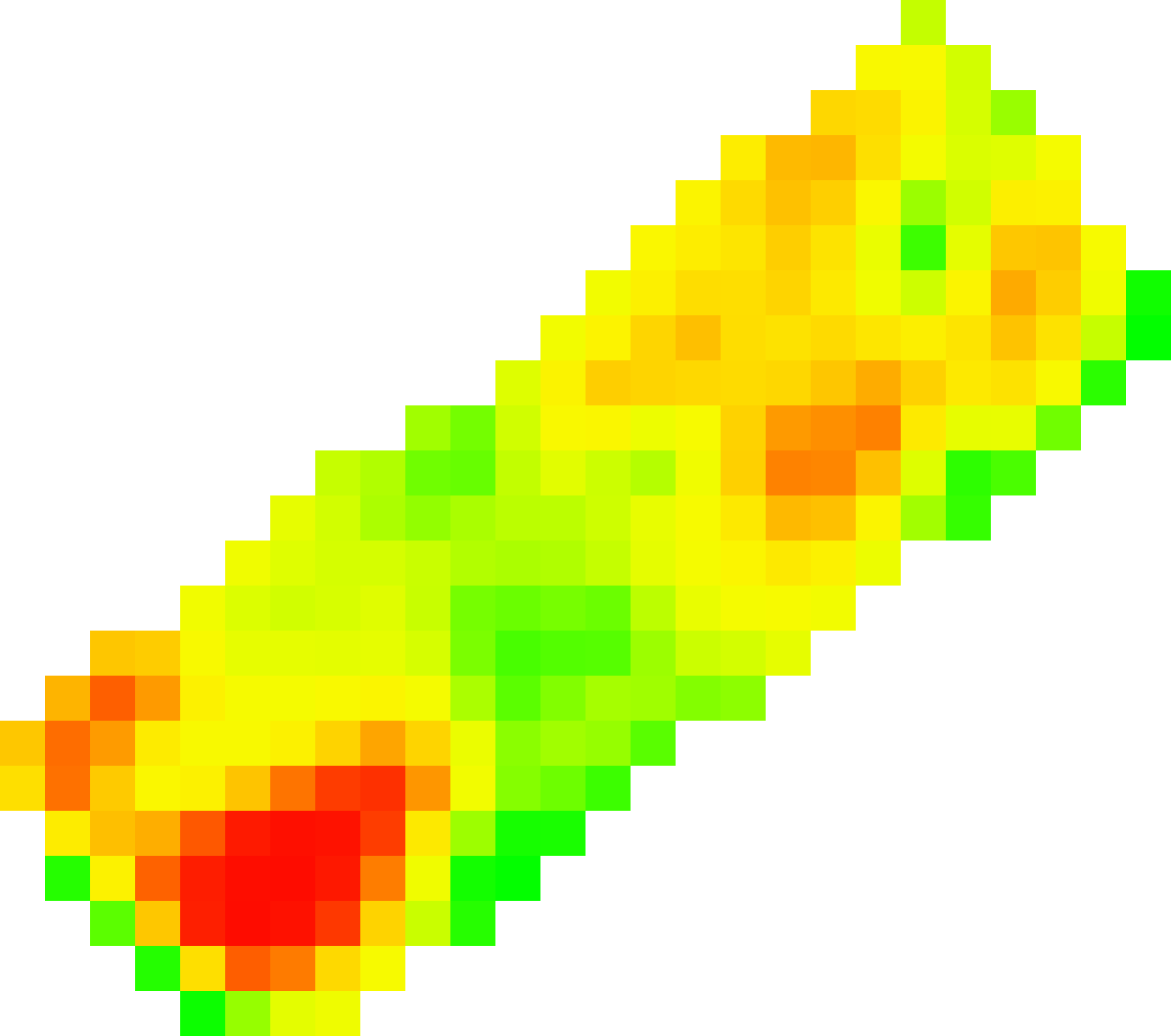}
        \caption{B11 (1610 nm)}
        \label{fig:b11}
    \end{subfigure}
    \hspace{0.25em}
    \begin{subfigure}[b]{0.15\textwidth}
        \includegraphics[width=\textwidth]{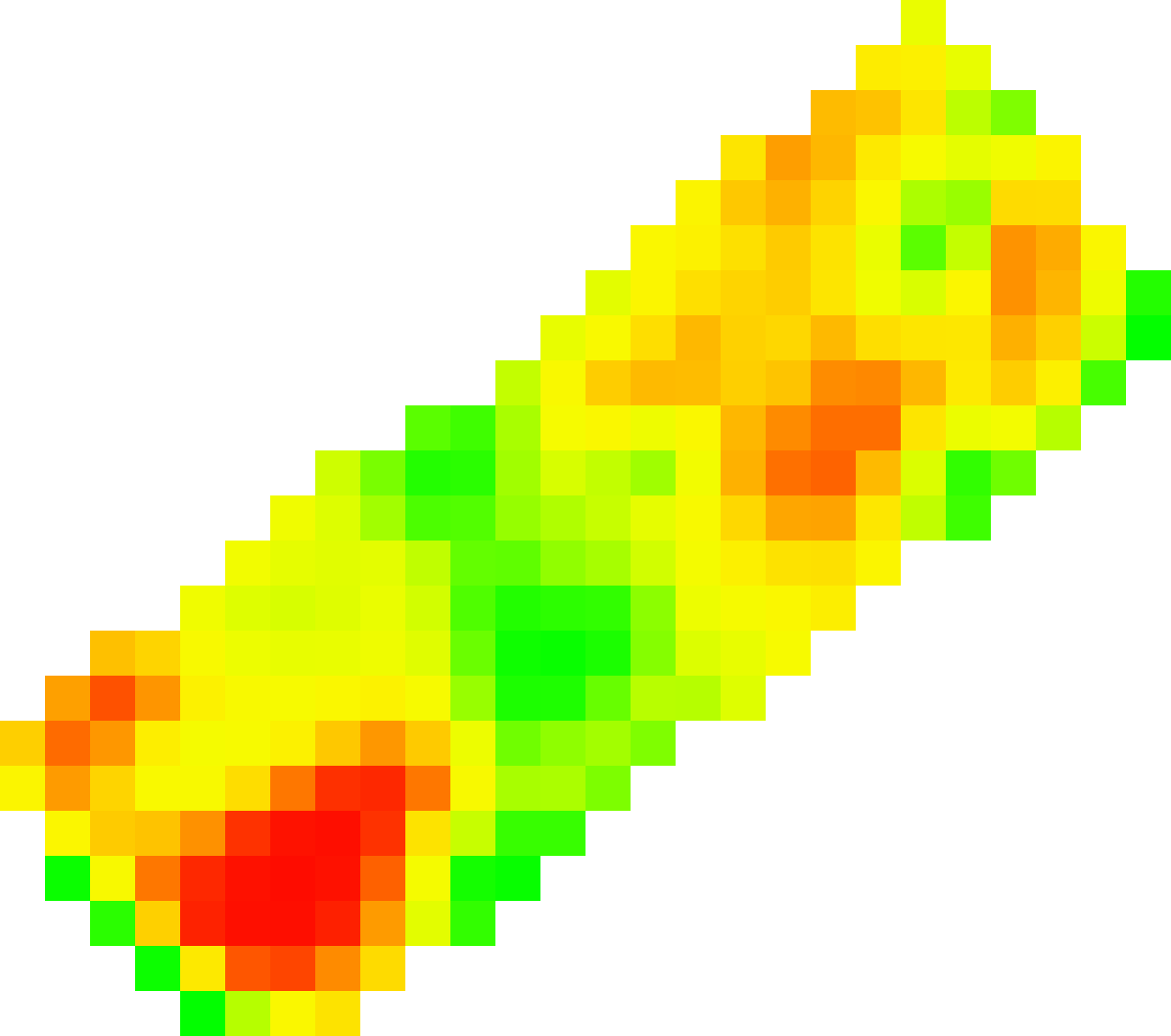}
        \caption{B12 (2190 nm)}
        \label{fig:b12}
    \end{subfigure}
        \vspace{1em}
    \begin{subfigure}[b]{0.5\textwidth}
        \centering
        \includegraphics[width=0.35\textwidth]{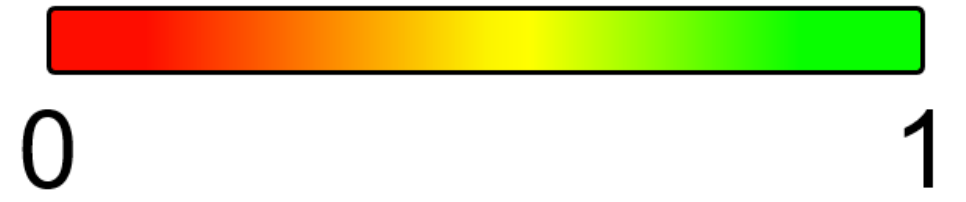}
    \end{subfigure}
    \caption{Sentinel-2 multispectral bands for an example block, with each panel showing a different band. Pixel values have been normalised to highlight within-field variation.}
    \label{fig:sentinel_bands}
\end{figure}

The Sentinel-2 data processing pipeline used the QGIS Python API to transform the coordinate reference system (CRS) of the Sentinel-2 products to match the CRS of the ground-truth shapefile representing the farms of interest. The ground-truth data was then used to extract the 76 sampled blocks from the Sentinel-2 imagery, ensuring that each pixel within the block geometries was labelled with both disease status and variety. Figure \ref{fig:sentinel_bands} shows an example block across Sentinel-2 bands, highlighting within-field variation captured in the multispectral data. Each labelled pixel within the shapefile geometries was subsequently utilised as a separate observation. Table \ref{tab1} summarises the number of ground-truth pixel observations at 20m resolution for each sugarcane variety.

\begin{table}[h]
\fontsize{8}{8}\selectfont
\caption{Ground truth pixel observations at a 20m resolution for RSD detection \label{tab1}}
\begin{tabular}{p{1.5cm}p{2cm}p{2cm}p{1cm}}
\hline\hline
\textbf{Variety} & \textbf{RSD Positive} & \textbf{RSD Negative} & \textbf{Total}  \\
\hline
Q200 & 145 & 389 & 534 \\
\midrule
Q208 & 869 & 649 & 1518 \\
\midrule
Q240 & 766 & 573 & 1339 \\
\midrule
Q253 & 886 & 1769 & 2655 \\
\midrule
SRA14 & 88 & 89 & 177 \\
\midrule
 & 2754 & 3469 & 6223 \\
\hline\hline
\end{tabular}
\end{table}

\begin{figure*}[H]
  \centering
  \begin{subfigure}[b]{0.32\textwidth}
    \includegraphics[width=\textwidth]{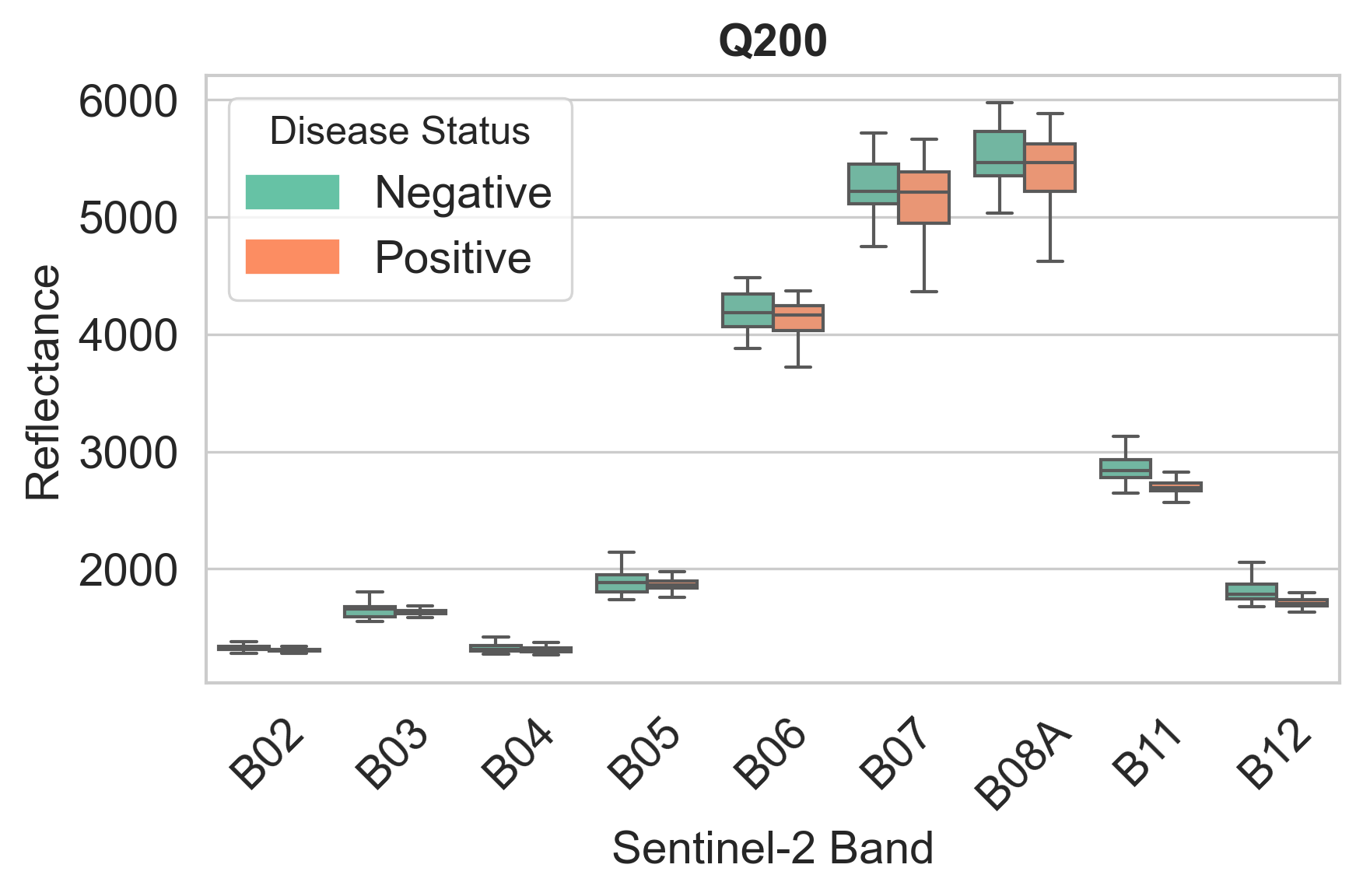}
    \label{fig:reflectance_var1}
  \end{subfigure}
  \hfill
  \begin{subfigure}[b]{0.32\textwidth}
    \includegraphics[width=\textwidth]{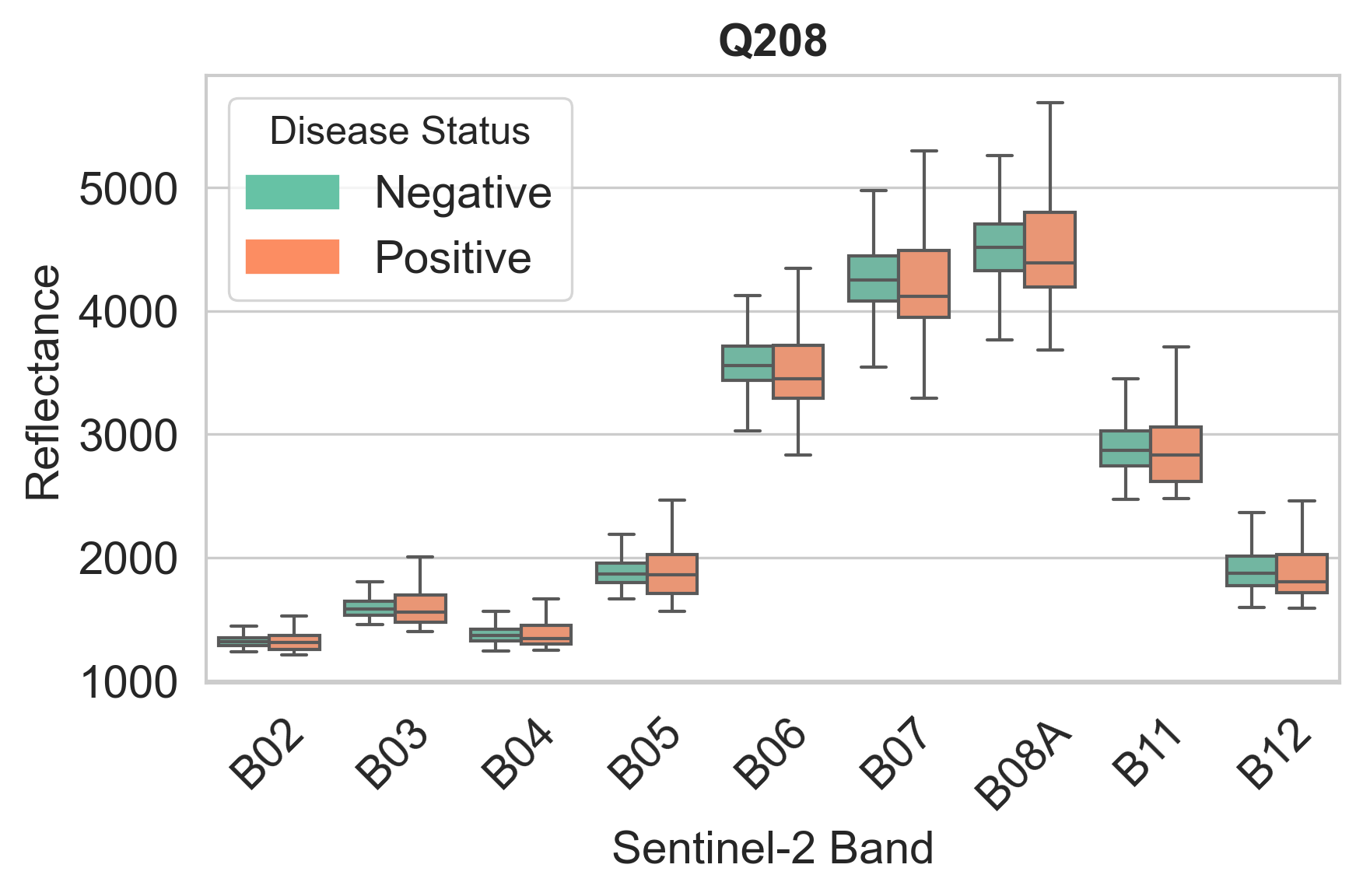}
    \label{fig:reflectance_var2}
  \end{subfigure}
  \hfill
  \begin{subfigure}[b]{0.32\textwidth}
    \includegraphics[width=\textwidth]{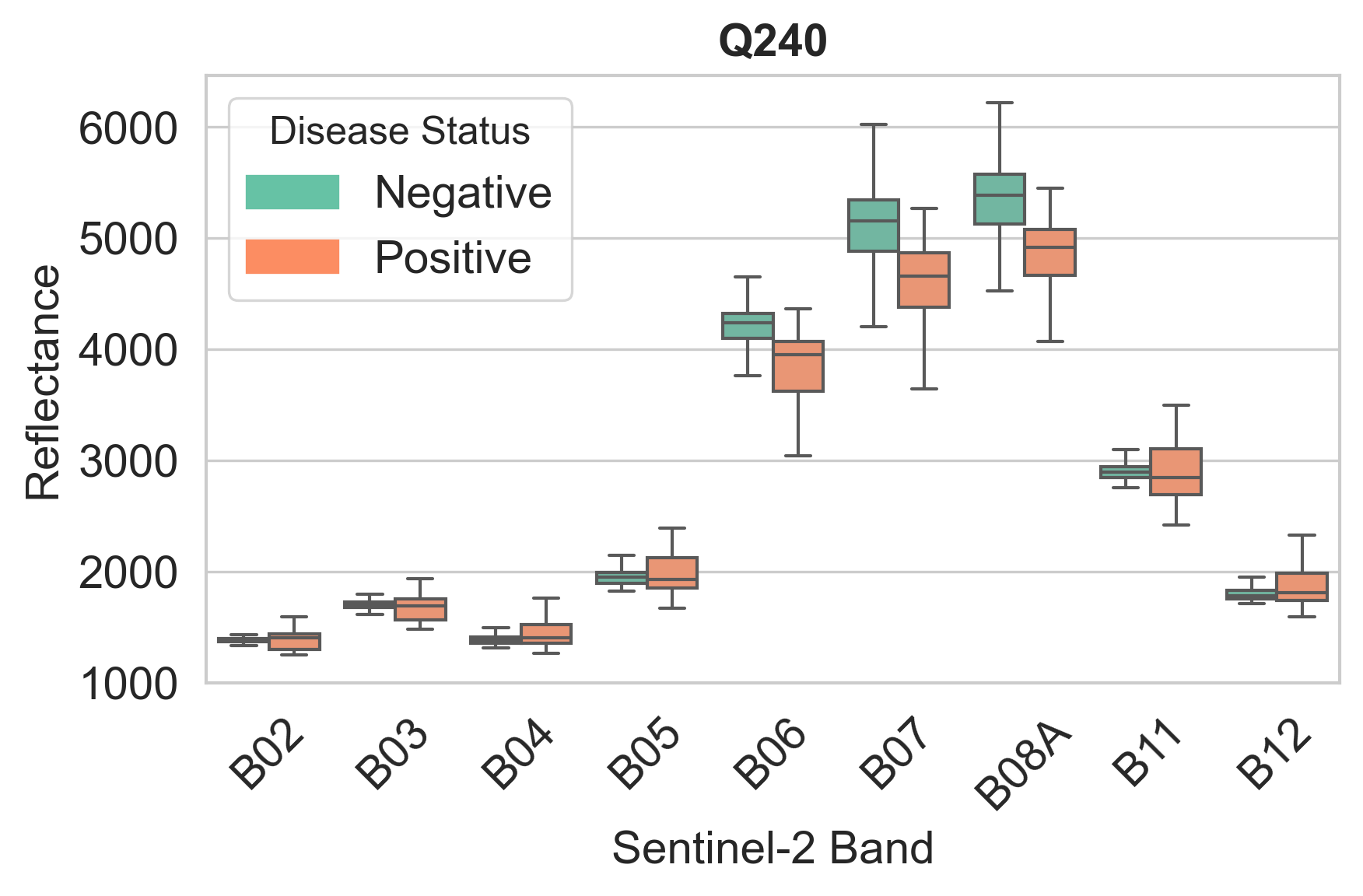}
    \label{fig:reflectance_var3}
  \end{subfigure}
  
  \begin{subfigure}[b]{0.32\textwidth}
    \includegraphics[width=\textwidth]{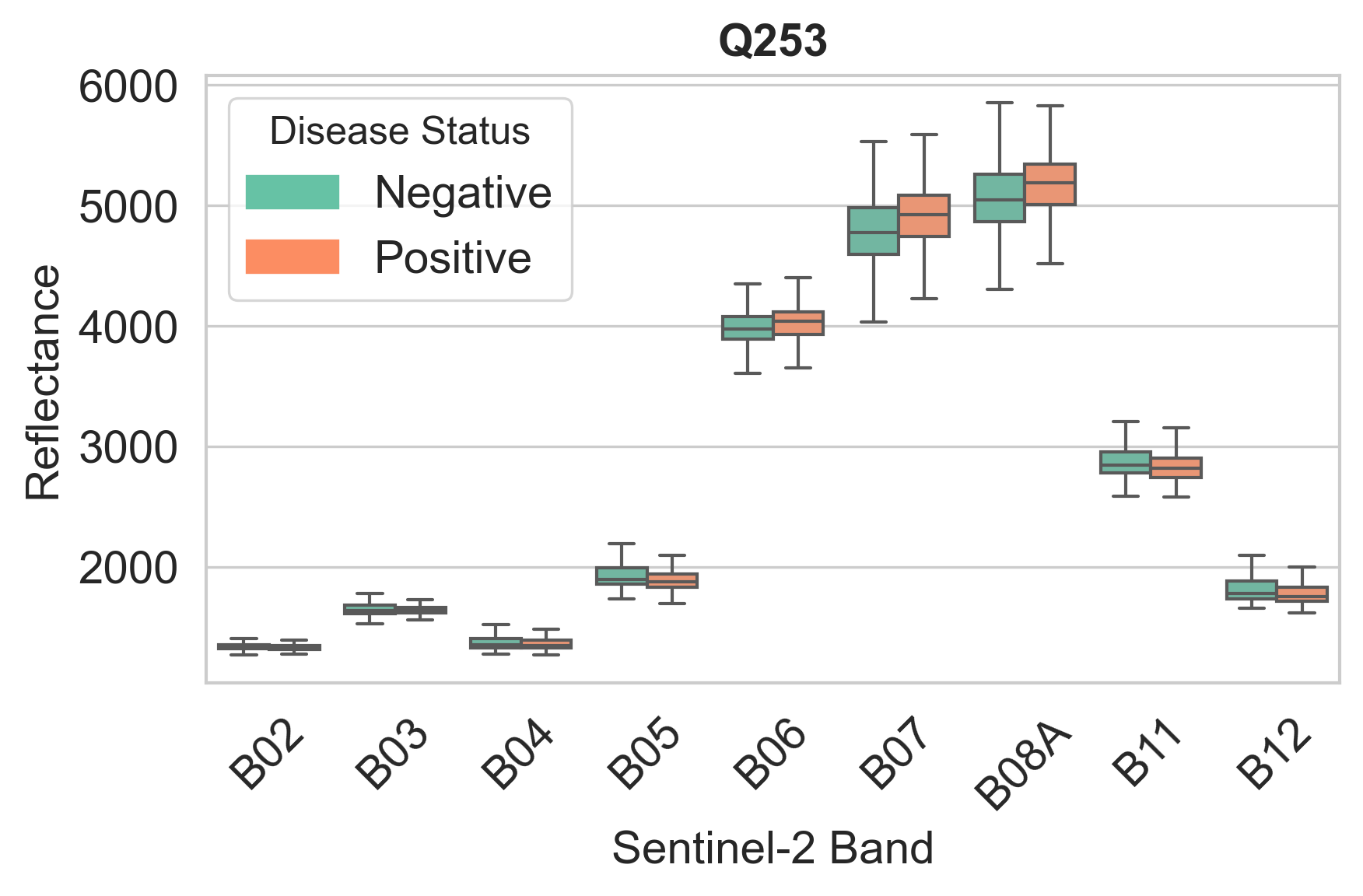}
    \label{fig:reflectance_var4}
  \end{subfigure}
  \hfill
  \begin{subfigure}[b]{0.32\textwidth}
    \includegraphics[width=\textwidth]{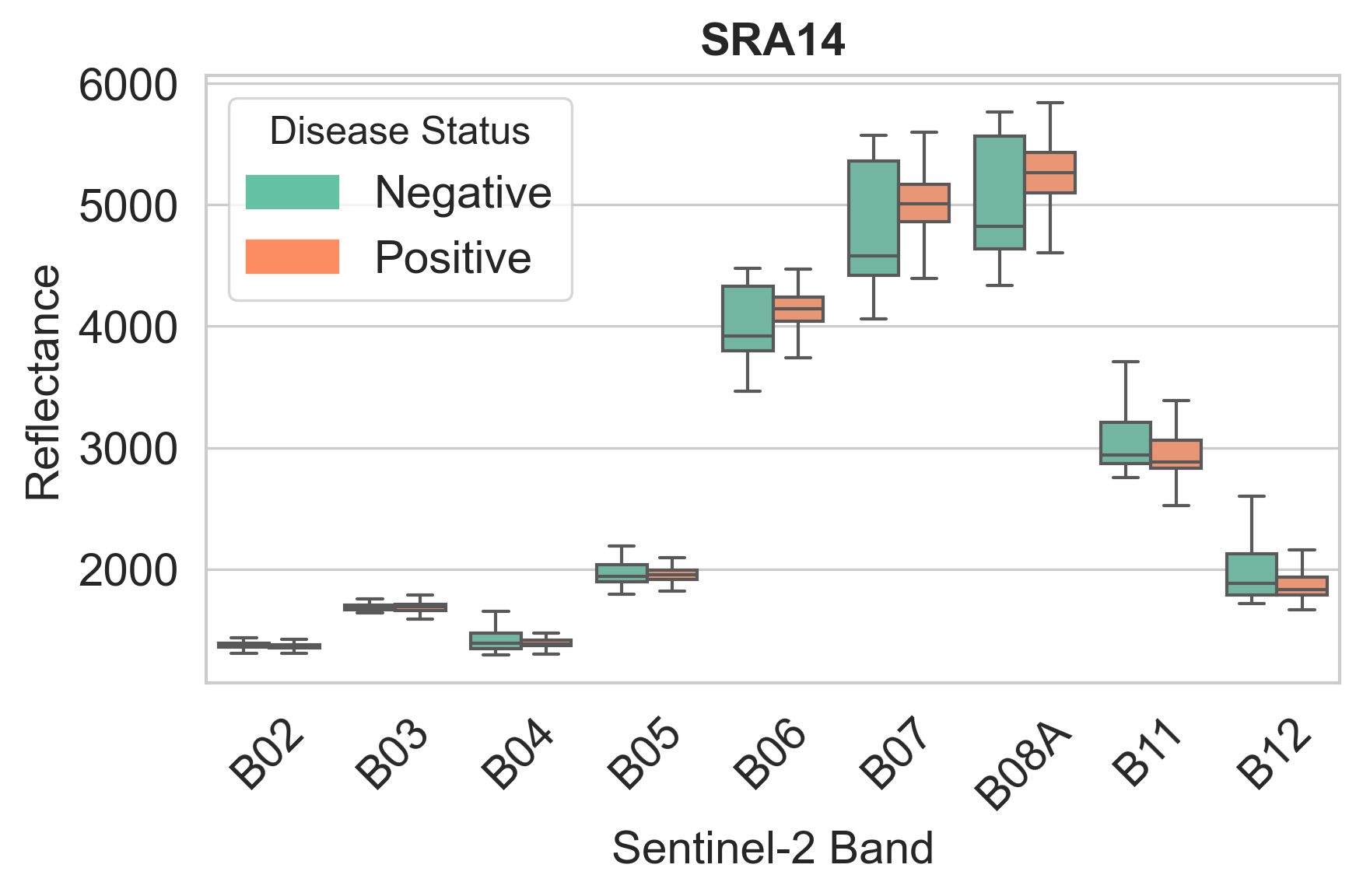}
    \label{fig:reflectance_var5}
  \end{subfigure}
  \hfill
  \begin{subfigure}[b]{0.32\textwidth}
    \includegraphics[width=\textwidth]{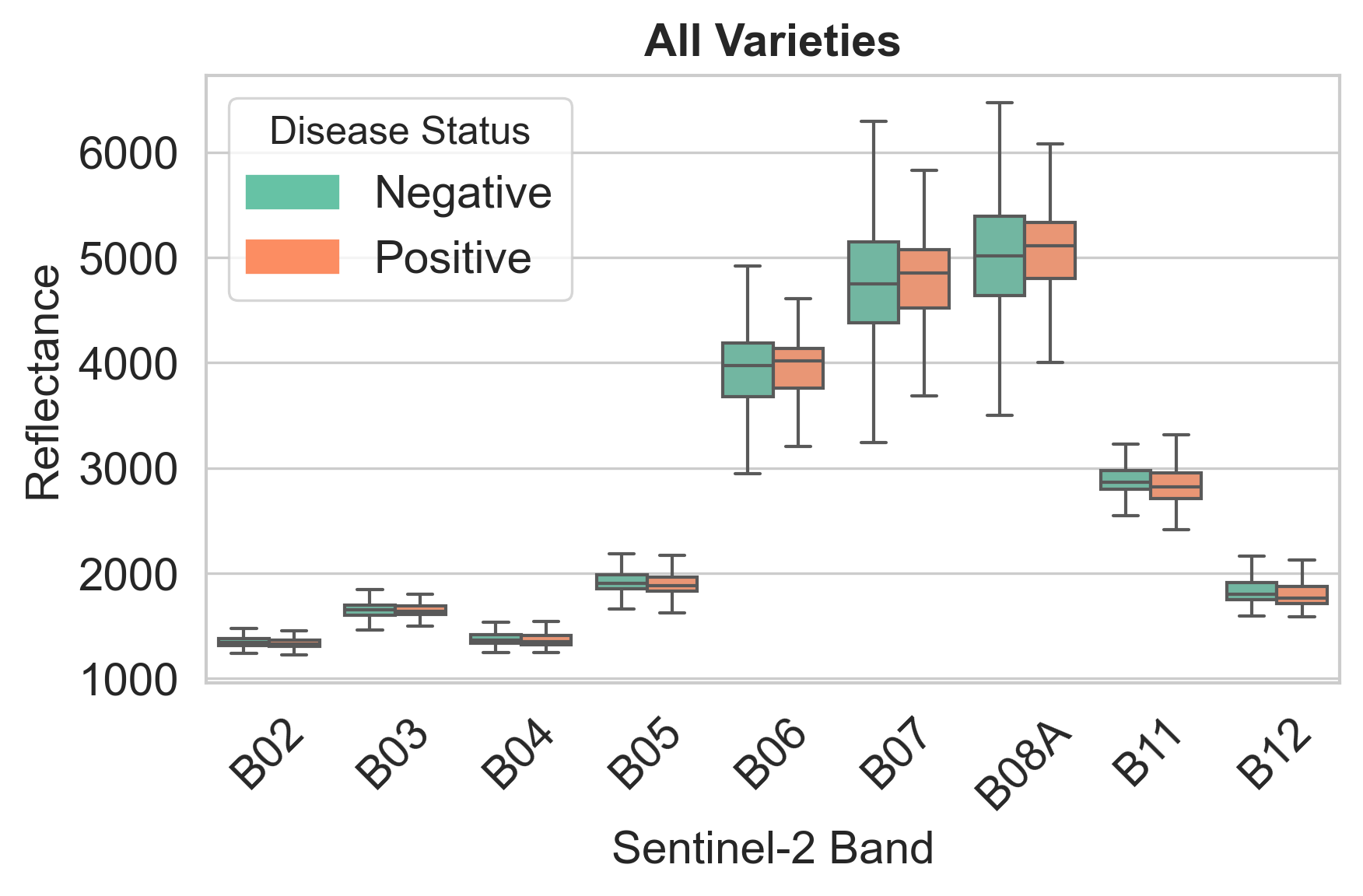}
    \label{fig:reflectance_overall}
  \end{subfigure}

  \caption{Reflectance distributions across Sentinel-2 bands for five sugarcane varieties and an overall panel. Outliers are omitted for clarity. Healthy and diseased crops are colour-coded (green = healthy, orange = diseased).}
  \label{fig:reflectance_distributions}
\end{figure*}

The distributions of reflectance values across Sentinel-2 bands vary with both disease status and variety. Figure \ref{fig:reflectance_distributions} illustrates the spectral response of each variety. For example, in variety Q240, RSD-positive pixels generally exhibit lower reflectance in the red-edge and near-infrared regions (bands 6–8) compared with RSD-negative pixels, whereas varieties such as SRA14 and Q253 show higher median reflectance for RSD-positive pixels.

\begin{table*}[!H]
\begin{center}
\fontsize{8}{8}\selectfont
\captionsetup{type=figure}
\captionof{table}{Vegetation indices utilised in this study for RSD detection.\label{tab2}}
\begin{tabular}{>{\centering\arraybackslash}p{0.25\textwidth}>{\centering\arraybackslash}p{0.35\textwidth}>{\centering\arraybackslash}p{0.35\textwidth}}
\toprule
\textbf{Reference}	& \textbf{Vegetation Index}	& \textbf{Formula}\\

\midrule
\citet{RN73} & Normalized Difference Vegetation Index (NDVI) & $\frac{NIR - RED}{NIR + RED}$\\
\midrule
\citet{RN74} & Atmospherically Resistant Vegetation Index (ARVI) & $\frac{NIR - (RED-BLUE)}{NIR + (RED-BLUE)}$\\
\midrule
\citet{RN75} & Simple ratio index (SRI) & $\frac{NIR}{RED}$\\
\midrule
\citet{RN76} & Plant Senescence Reflectance index (PSRI) & $\frac{R_{680} - R_{500}}{R_{750}}$\\
\midrule
\cite{RN77} & Ratio Vegetation Index (RVI) & $\frac{RED}{NIR}$\\
\midrule
\citet{RN78} & Normalized difference Water index (NDWI)	& $\frac{GREEN - NIR}{GREEN + NIR}$\\
\midrule
\citet{RN79} & Normalized Difference Moiseture Index (NDMI)	& $\frac{NIR - SWIR}{NIR + SWIR}$\\
\midrule
\citet{RN80} & Normalized Green Red Difference Index (NGRDI) & $\frac{GREEN - RED}{GREEN + RED}$\\
\midrule
\citet{RN82} & Visible Atmospherically Resistant Index (VARI)	& $\frac{GREEN - RED}{GREEN+RED - BLUE}$\\
\midrule
 \citet{VI1} & Simple Ratio 860/550 & $\frac{R_{800}}{R_{550}}$ \\
\midrule
\citet{RN18, RN19} & Disease-Water Stress Index 1 (DWSI-1) & $\frac{R800}{R1660}$\\
\midrule
\citet{RN18, RN19} & Disease-Water Stress Index 2 (DWSI-2) & $\frac{R1660}{R550}$\\
\midrule
\citet{RN18, RN19} & Disease-Water Stress Index 3 (DWSI-3) & $\frac{R1660}{R680}$\\
\midrule
\citet{RN18, RN19} & Disease-Water Stress Index 4 (DWSI-4) & $\frac{R550}{R680}$\\
\midrule
\citet{RN18, RN19} & Disease-Water Stress Index 5 (DWSI-5) & $\frac{R800 + R550}{R1660 + R680}$\\
\midrule
\citet{RN73} & Green Blue NDVI (GBNDVI) & $\frac{NIR - (Blue + Green)}{NIR + Blue + Green)}$\\
\midrule
This Study & Disease-Water Stress Index 6 (DWSI-6) & $\frac{B12 + B11}{B8A}$ \\
\midrule
This Study & Disease-Water Stress Index 7 (DWSI-7) & $\frac{B12 + B11}{B4}$ \\
\midrule
This Study & Disease-Water Stress Index 8 (DWSI-8) & $\frac{B12 + B11}{B4 + B5}$ \\
 
\hline\hline
\end{tabular}
\end{center}
\end{table*}

\subsection{Vegetation Indices}
The causal bacteria of RSD infects the xylem vessels responsible for water transport in sugarcane, reducing water absorption and resulting in vegetation stress \citep{RN474, RSD3}. To detect these internal symptoms, vegetation indices were selected that measure key indicators of plant health, as shown in Table \ref{tab2}. Several indices were chosen to assess vegetation moisture content \citep{RN79, RN18}, with the intention highlighting reduced water uptake caused by RSD infection of the xylem. Others focus on assessing symptoms of plant stress \citep{RN76, RN19, RN73}, indicated by changes in photosynthetic activity or chlorophyll levels. In addition to vegetation indices, previous research has demonstrated that the raw Sentinel-2’s bands in the Short-wave Infrared Region (SWIR) are sensitive to moisture absorption in vegetation \citep{RN500, RN501, RN502}, while bands in the Near-Infrared (NIR) region have been previously used as indicators of vegetation stress \citep{stress}, making both potential predictors for RSD. Together, the indices and bands provide a comprehensive net to capture the expected physiological impacts of RSD on sugarcane.

To optimise class separation, strategic modifications of vegetation indices were explored. The alteration of existing indices was guided by first identifying statistically significant bands for diagnosing RSD with an independent t-test (significance level = 0.05) for each variety and spectral band, with Bonferroni's correction applied to adjust for type I error. Although statistically significant bands varied between varieties, bands B08A and B12 were consistent across most varieties, supporting the connection between vegetation moisture and stress as key indicators for diagnosing RSD. With the identification of statistically significant bands, opportunities arise to modify key vegetation indices to enhance class separation. Among the most relevant indices are the Disease Water Stress Indices (DWSIs), which have proven effective in detecting sugarcane disease through satellite-based remote sensing \citet{RN19}. The DWSIs were designed to target water absorption at R1660 (Sentinel-2's B11). Incorporating both B11 and B12, given their vegetation moisture absorption sensitivity \citep{RN500, RN501, RN502} and B12's statistical significance, led to modified versions of these indices for Sentinel-2, DWSI 6 and 7. Additionally, DWSI 7 was further modified to include the red-edge band, B5, as the red-edge region is sensitive to stress-related changes in vegetation and has previously been effective for vegetation classification \citep{RN503, RN504}. This modification resulted in the creation of DWSI 8, designed to better capture stress-related signals in vegetation as a result of RSD.

The spectral bands of the satellite image captured on the 27th February 2022 were utilised to compute the 19 vegetation indices listed in Table \ref{tab2} with the QGIS Python API for each pixel in the dataset. A visualisation of several key vegetation indices for an example block is shown in Figure~\ref{fig:vi_block}, illustrating patterns relevant to sugarcane disease detection.

\begin{figure}[!htbp]
    \centering

    \begin{subfigure}[b]{0.15\textwidth}
        \includegraphics[width=\textwidth]{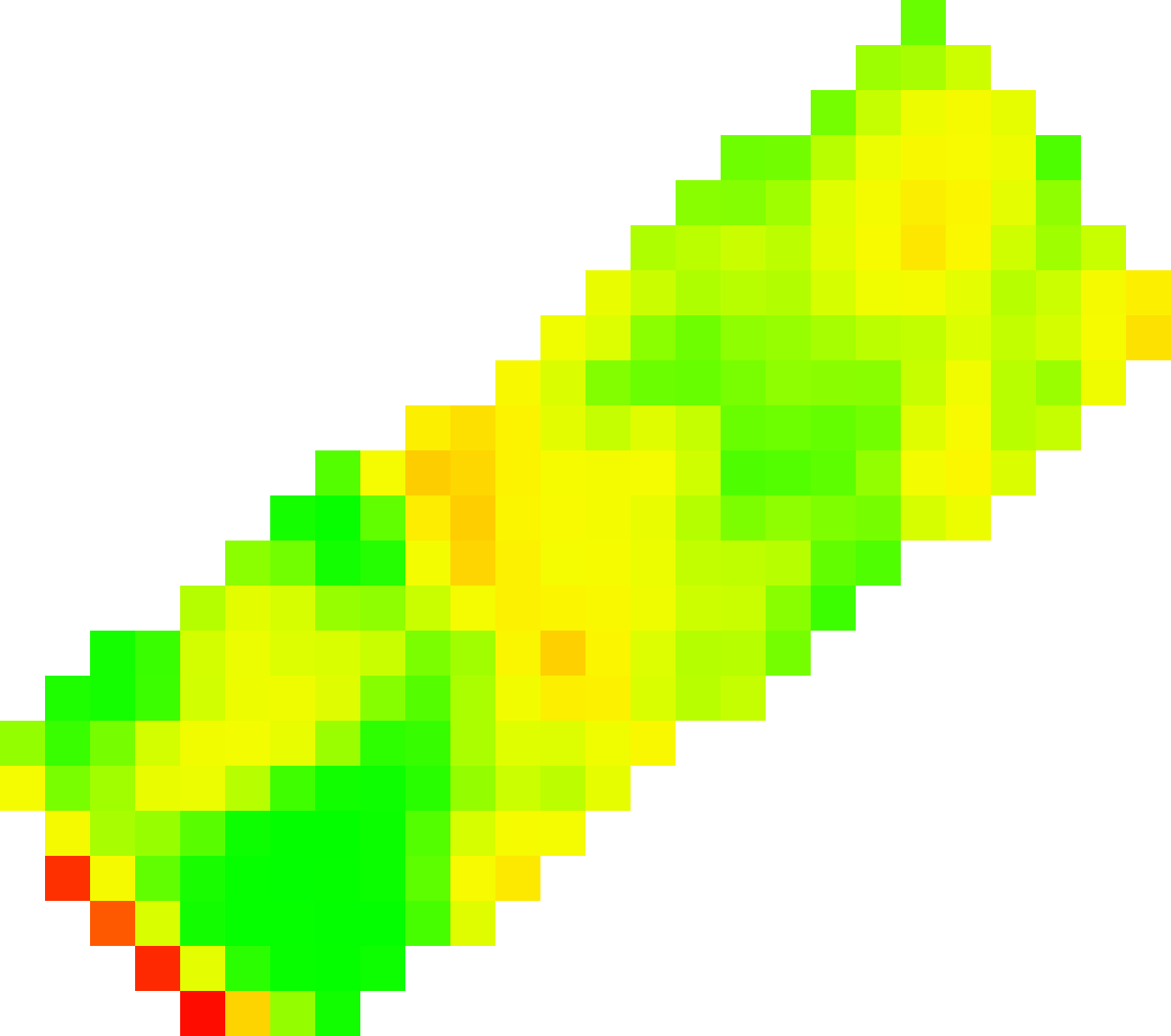}
        \caption{DWSI-1}
        \label{fig:DWSI-1}
    \end{subfigure}
    \hspace{0.25em}
    \begin{subfigure}[b]{0.15\textwidth}
        \includegraphics[width=\textwidth]{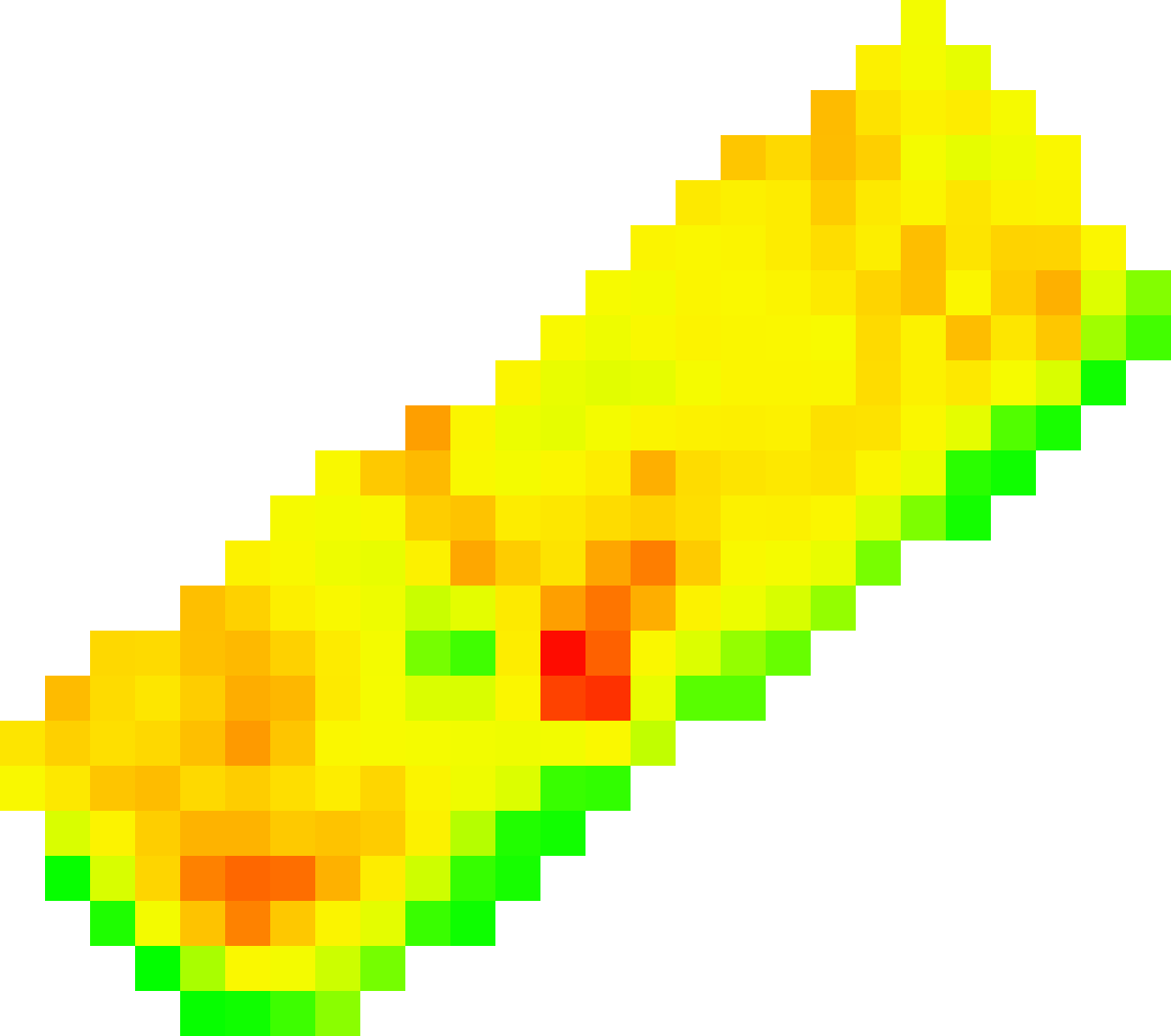}
        \caption{DWSI-2}
        \label{fig:DWSI-2}
    \end{subfigure}
    \hspace{0.25em}
    \begin{subfigure}[b]{0.145\textwidth}
        \includegraphics[width=\textwidth]{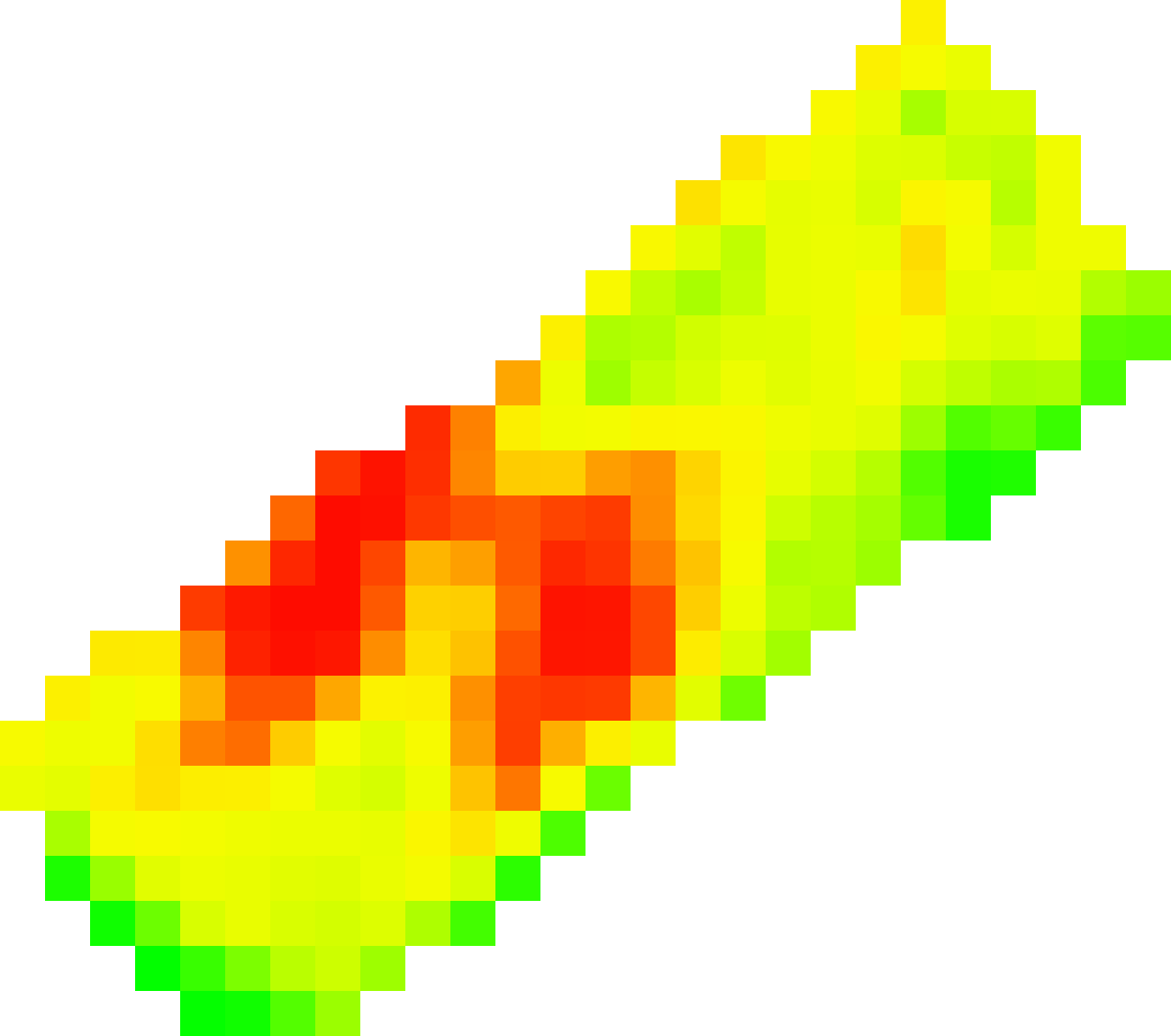}
        \caption{DWSI-3}
        \label{fig:DWSI-3}
    \end{subfigure}
    \vspace{1em}
    
    \begin{subfigure}[b]{0.15\textwidth}
        \includegraphics[width=\textwidth]{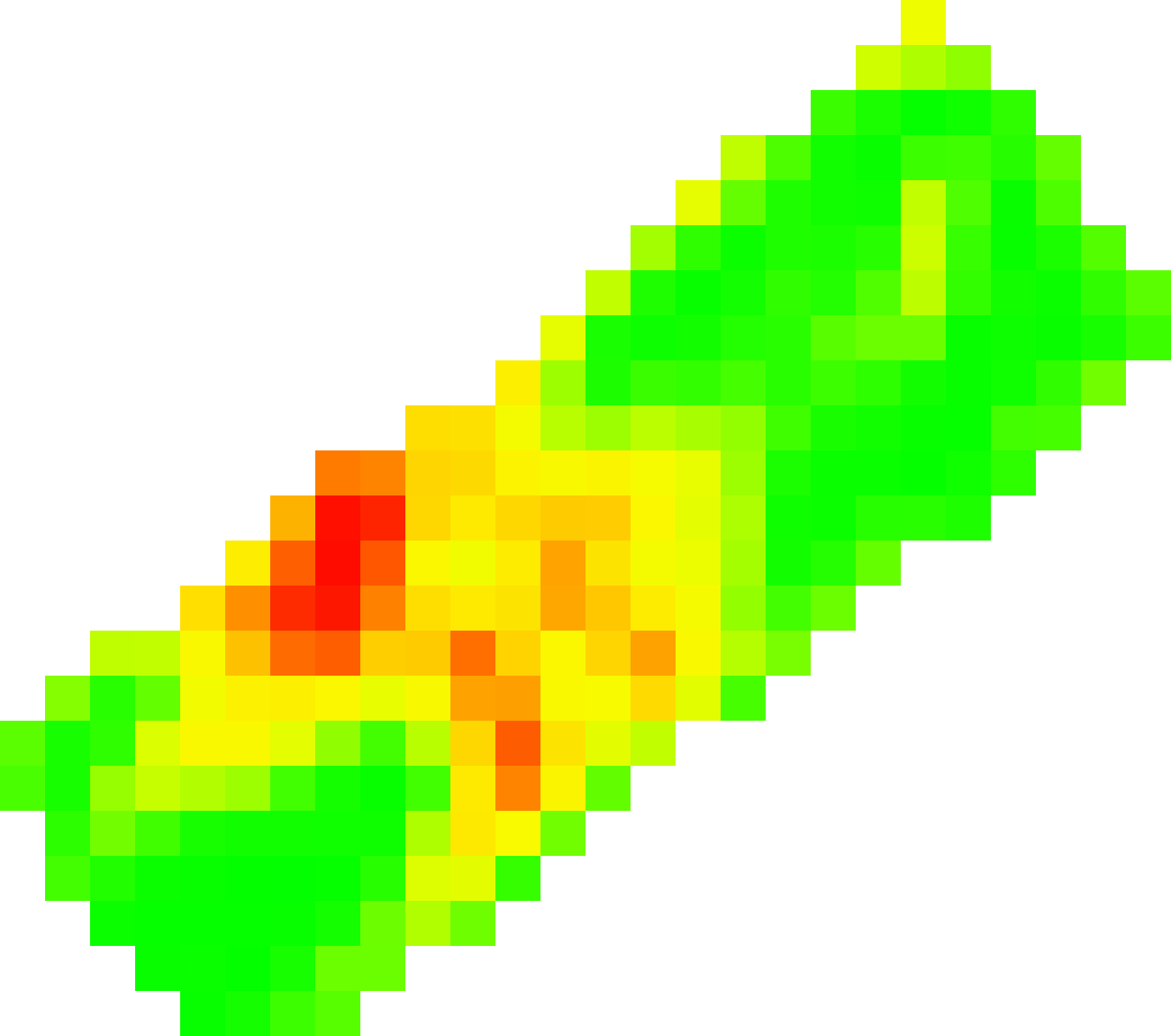}
        \caption{DWSI-4}
        \label{fig:DWSI-4}
    \end{subfigure}
    \hspace{0.25em}
    \begin{subfigure}[b]{0.15\textwidth}
        \includegraphics[width=\textwidth]{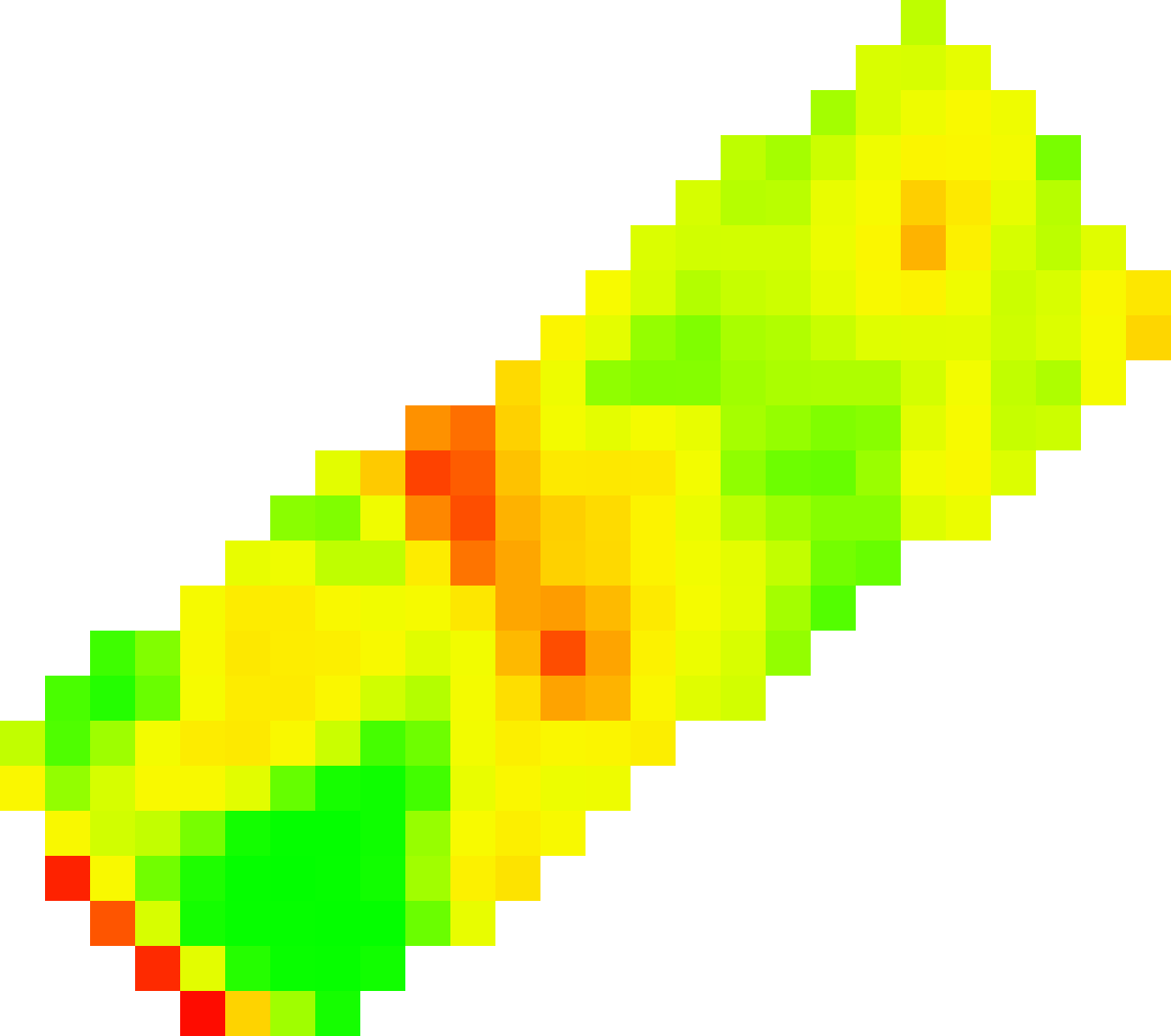}
        \caption{DWSI-5}
        \label{fig:DWSI-5}
    \end{subfigure}
    \hspace{0.25em}
    \begin{subfigure}[b]{0.15\textwidth}
        \includegraphics[width=\textwidth]{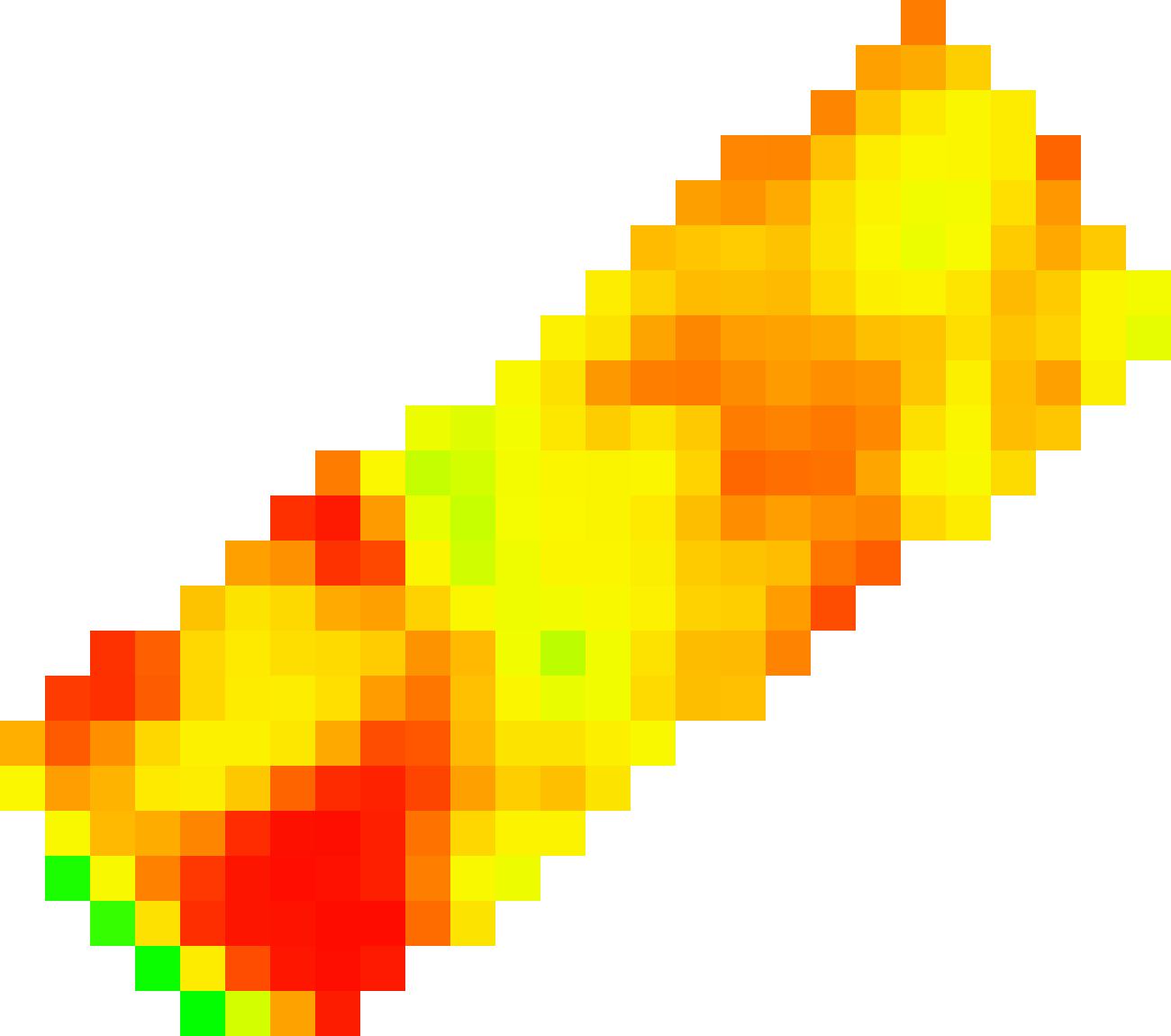}
        \caption{DWSI-6}
        \label{fig:DWSI-6}
    \end{subfigure}
    
        \vspace{1em}
    
    \begin{subfigure}[b]{0.15\textwidth}
        \includegraphics[width=\textwidth]{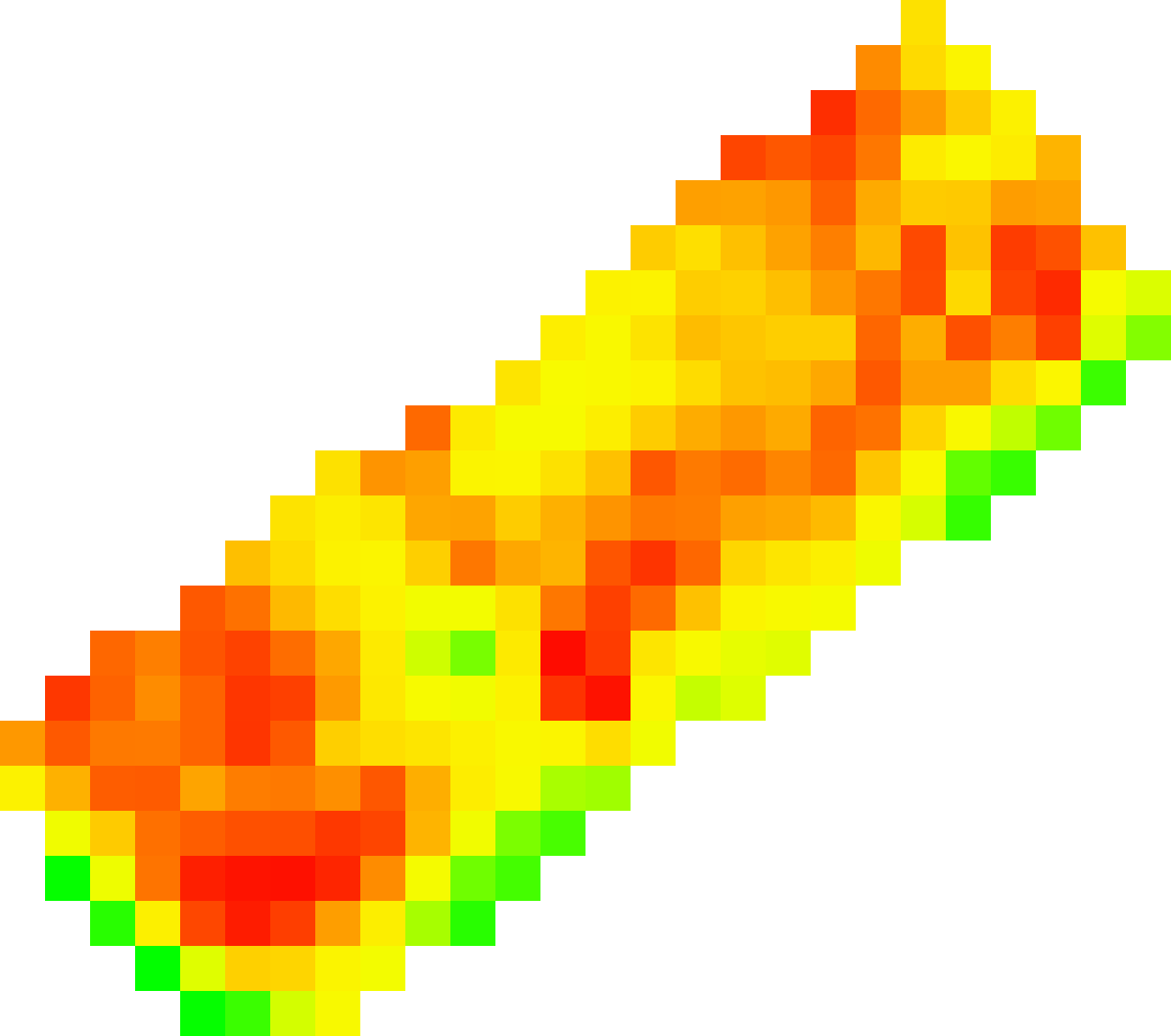}
        \caption{DWSI-7}
        \label{fig:DWSI-7}
    \end{subfigure}
    \hspace{0.25em}
    \begin{subfigure}[b]{0.15\textwidth}
        \includegraphics[width=\textwidth]{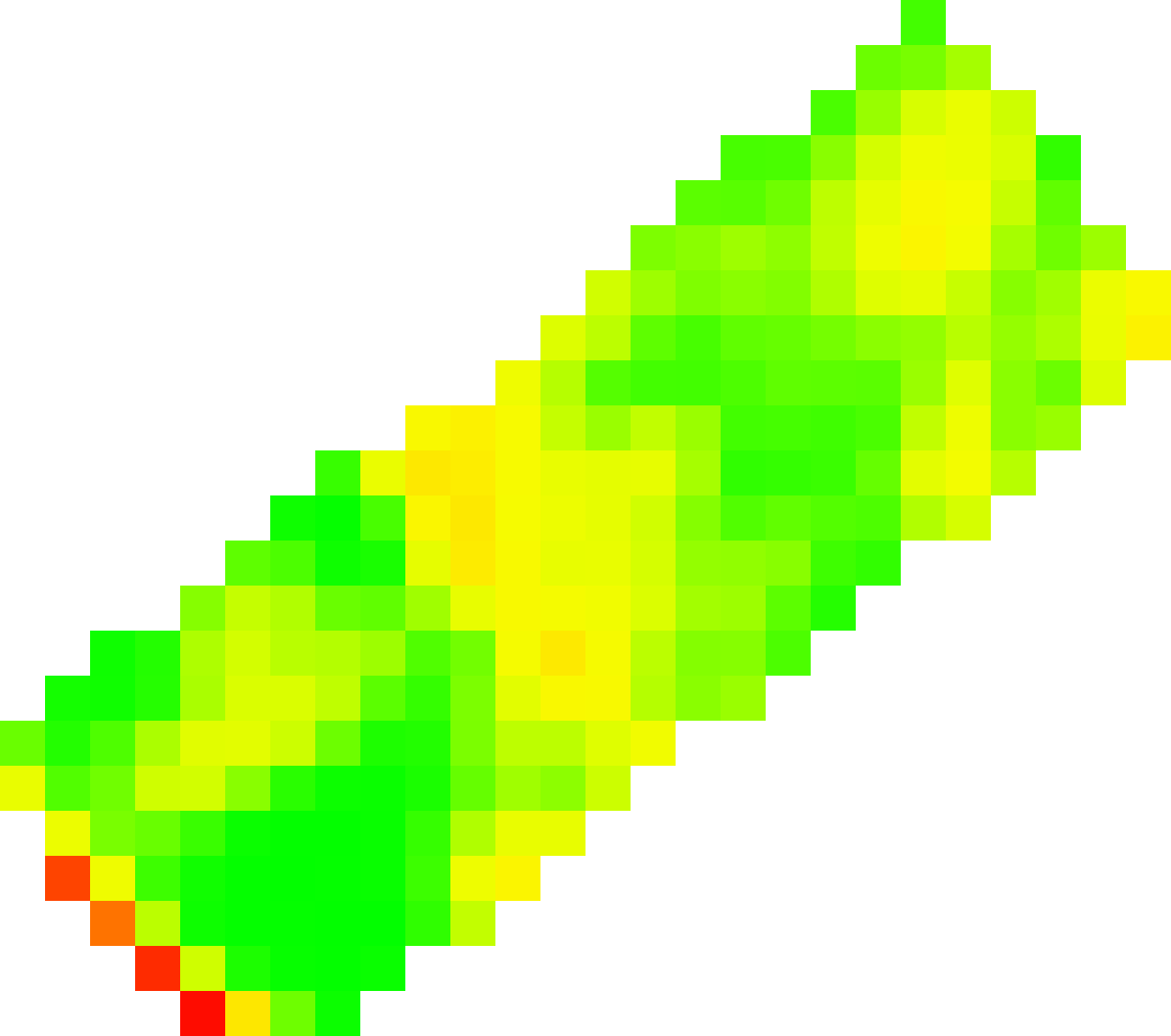}
        \caption{NDMI}
        \label{fig:NDMI}
    \end{subfigure}
    \hspace{0.25em}
    \begin{subfigure}[b]{0.15\textwidth}
        \includegraphics[width=\textwidth]{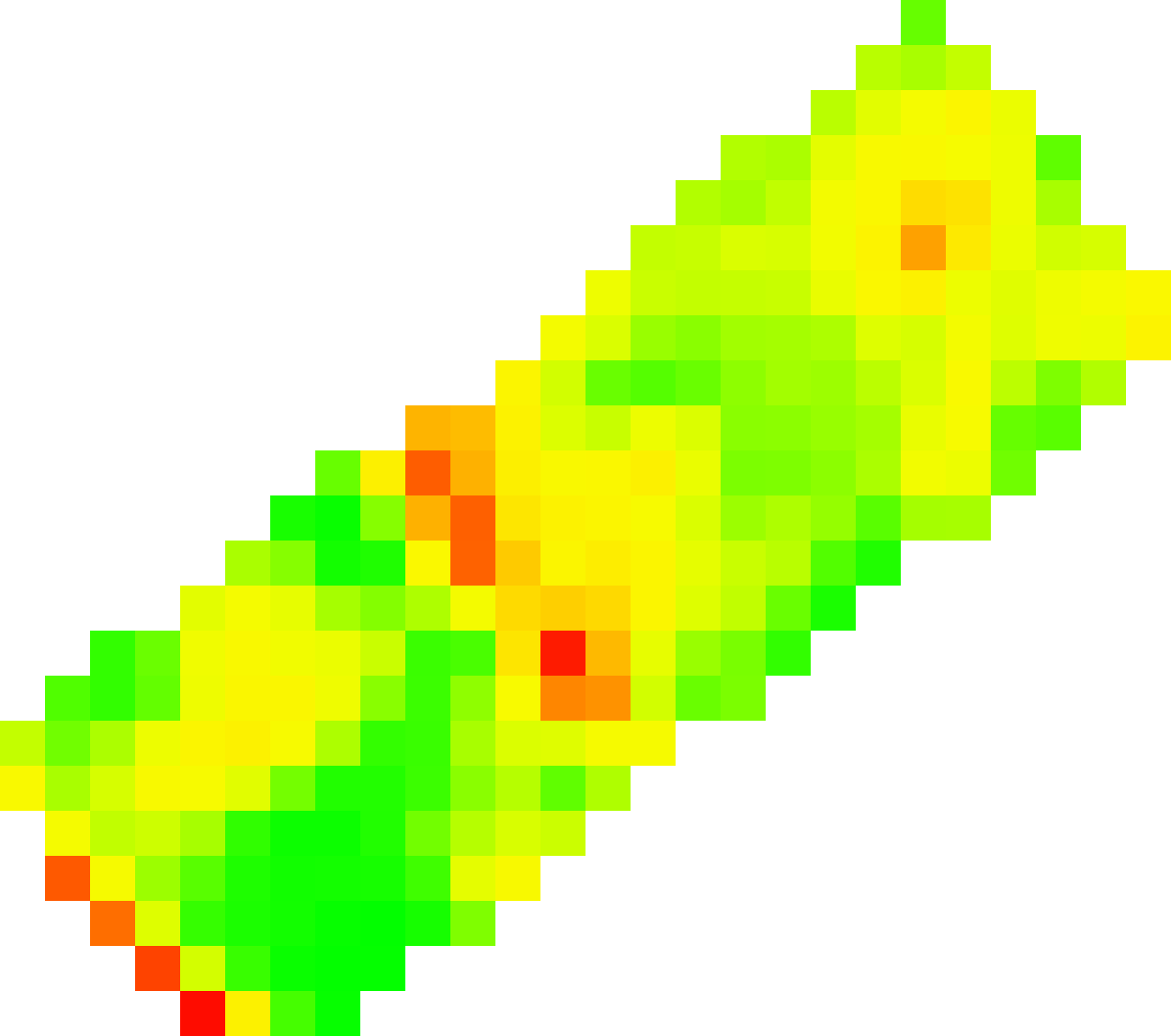}
        \caption{NDVI}
        \label{fig:NDVI}
    \end{subfigure}
        \begin{subfigure}[b]{0.5\textwidth}
        \centering
        \includegraphics[width=0.35\textwidth]{colorbar.pdf}
    \end{subfigure}
    \caption{Nine vegetation indices for a sample block that are historically significant for vegetation monitoring or were found to be important for detecting RSD. Pixel values have been normalised to highlight within-field variation.}
\end{figure}

\vspace{-2em}

\subsection{Machine Learning Algorithm Development}
In this study, we compared five machine learning algorithms to predict the RSD status of a pixel. These are Random Forest (RF), Logistic Regression (LR), Quadratic Discriminant Analysis (QDA), Gradient Boosting (GB) and Radial Basis Function Support Vector Machine (SVM-RBF). Tree-based methods, such as RF and GB, were included due to their previous success in similar remote sensing studies. These models are particularly effective for handling high-dimensional data, non-linearity, and interactions among features, making them well-suited for RSD detection. SVM-RBF kernel was included due to its ability to effectively model complex non-linear relationships, which are often present in multispectral data. LR was included due to its simplicity and interpretability. LR is useful for assessing whether the problem could be adequately solved with linear methods without resorting to more complex models. QDA was tested as it is capable of modelling quadratic decision boundaries, which can be useful for capturing more complex class separations that may exist in the data.

Given that the sugarcane variety can be potentially unknown to the users, two sets of models were developed, one for when the sugarcane variety is known and the other when its unknown. The final dataset seen in Table \ref{tab1} is largely unbalance, therefore, the dataset was down-sampled to achieve equal class distribution, to mitigate potential bias.

\begin{figure*}[H!]
    \centering
    \includegraphics[width=1\textwidth]{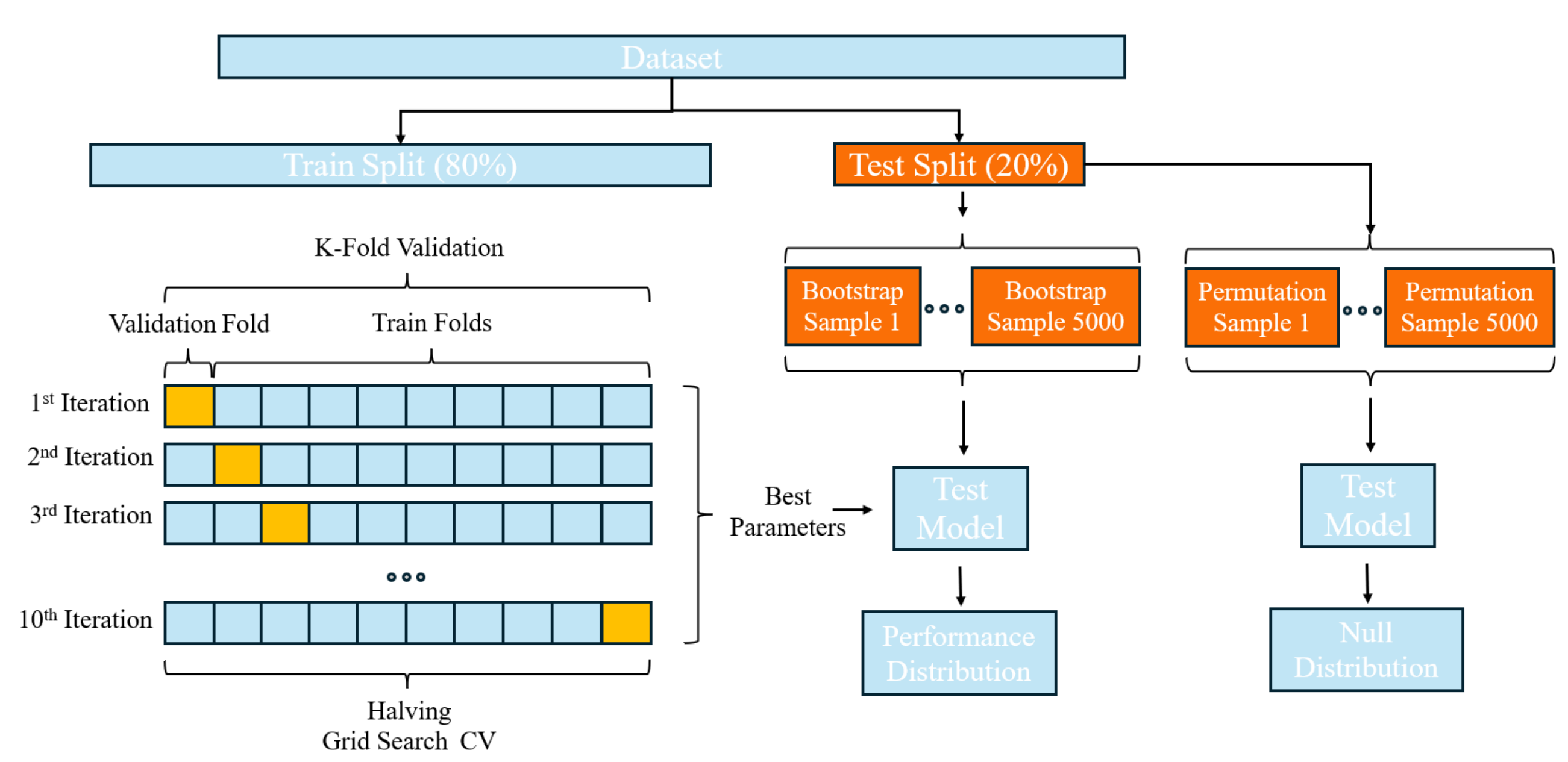} 
    \caption{Visual guide to the machine learning workflow. The dataset was split into training and testing sets. The training set underwent 10-fold cross-validation to select the best model. Model performance was then assessed by bootstrapping the test set 5,000 times, followed by permutation testing.}
    \label{fig:ml_pipeline}
\end{figure*}

The dataset was partitioned into 80\% for training and 20\% for testing. Standard scaling was applied to the entire dataset, with the scaling parameters fitted exclusively on the training set. Hyperparameters were tuned using 10-fold cross-Validation with "Halving Grid Search" on the 80\% training set. HalvingGridSearchCV is a resource-efficient version of GridSearchCV from the scikit-learn library for hyperparameter tuning. HalvingGridSearchCV is an implementation of the successive halving algorithm, introduced by \citet{HGS2}, which starts with many candidate parameter combinations and progressively eliminates the least promising ones, allocating more resources as the number of iterations increases \citep{HGS}. Table \ref{exp_params} provides a summary of the key parameters adjusted, including the search space explored.

The model's performance was evaluated by bootstrapping the unseen test set with 5000 bootstrap samples, generating distributions for the chosen performance metrics, accuracy, precision, and recall \citep{Ting2010}. Accuracy provides an overall measure of correct predictions, while precision and recall offer insights into the model's ability to correctly identify positive cases and avoid false positives, respectively. In the context of large-scale disease detection, evaluating false positives and false negatives is critical to ensuring an appropriate balance is achieved. High false positive rates could lead to unnecessary interventions by agronomists, wasting time, diverting resources, and reducing the economic benefit of the system. After model evaluation, each machine learning algorithm was trained with 10-fold cross-validation on the entire dataset, to find the most appropriate hyperparameters for the entire dataset, after the performance has been determined. A visualisation of this machine learning workflow is shown in Figure~\ref{fig:ml_pipeline}.
 
\begin{table}[H]
\fontsize{8}{8}\selectfont
\caption{Summary of hyperparameters search space utilised in halving grid search.  \label{exp_params}}
\begin{tabular}{p{0.8cm}p{1.5cm}p{5cm}}
\hline\hline
\textbf{Model} & \textbf{Parameter} & \textbf{Search Space} \\
\hline
\multirow{3}{*}{SVM} & \multirow{2}{*}{C} & 0.1, 1, 10, 100, 1000, 10000, 100000, 1000000  \\
& gamma & 0.001, 0.01, 0.1, 1, 10, 100  \\
\midrule
\multirow{2}{*}{RF} & n\_estimators & 50, 100, 200, 500, 1000, 1500  \\
& max\_depth & None, 10, 15, 20, 30  \\
\midrule
\multirow{3}{*}{GB} & n\_estimators & 50, 100, 200, 500, 1000, 1500  \\
& learning\_rate & 0.01, 0.1, 0.2, 0.3  \\
& max\_depth & 3, 5, 7, 10  \\
\midrule
\multirow{2}{*}{LR} & penalty & 'l1', 'l2'  \\
& solver & 'liblinear', 'saga'  \\
& gamma & 0.001, 0.01, 0.1, 1, 10, 100  \\
\midrule
QDA & N/A & N/A  \\

\hline\hline
\end{tabular}
\end{table}
The Gini importance of the final RF and GB models was extracted to identify the most important vegetation indices and Sentinel-2 multispectral bands for RSD detection. Additionally, to indicate important features for RSD detection, independent t-tests (significance level = 0.05) were performed for each vegetation index, applying Bonferroni's correction to adjust for type I error. Although the results are included in the appendix, this approach does not account for correlations among the indices or interactions between them. Therefore, the findings should be interpreted with caution.

To assess the significance of the ML algorithms' ability to detect RSD, a permutation test was conducted. The pixel labels were shuffled, and the models were retrained using the previously determined hyperparameters. This process was repeated 1000 times to generate a null distribution of accuracy, reflecting model performance when there is no true association between the features and disease status. By comparing the original performance distribution to the null distribution, the statistical significance of the observed model performance can be evaluated.

\subsection{Software, Libraries \& Experimental Environment}

This study was conducted with several software tools and libraries for data processing, analysis, and model development. QGIS 3.36.0 was employed for examining satellite imagery and utilizing its Python API to calculate vegetation indices, perform spatial joins, and extract tabular data from the spatial layers. Visual Studio Code version 1.89.1 was utilised as the programming IDE. Python 3.9.18 and a list of its libraries were utilised, including the following: matplotlib 3.7.2, seaborn 0.12.2, numpy 1.24.3, pandas 2.0.3, scikit-learn 1.3.0 and scipy 1.11.1.

All experiments were conducted on a system running Microsoft Windows 11 Home. The hardware configuration included an Intel(R) Core (TM) i7-8700 CPU operating at 3.20 GHz with 6 cores and 12 logical processors, supported by 16 GB of installed physical memory.

\section{Results}
\begin{figure*}[htbp]
    \centering
    
    % Support Vector Machine
    \begin{subfigure}[b]{0.28\textwidth}
        \includegraphics[width=\textwidth]{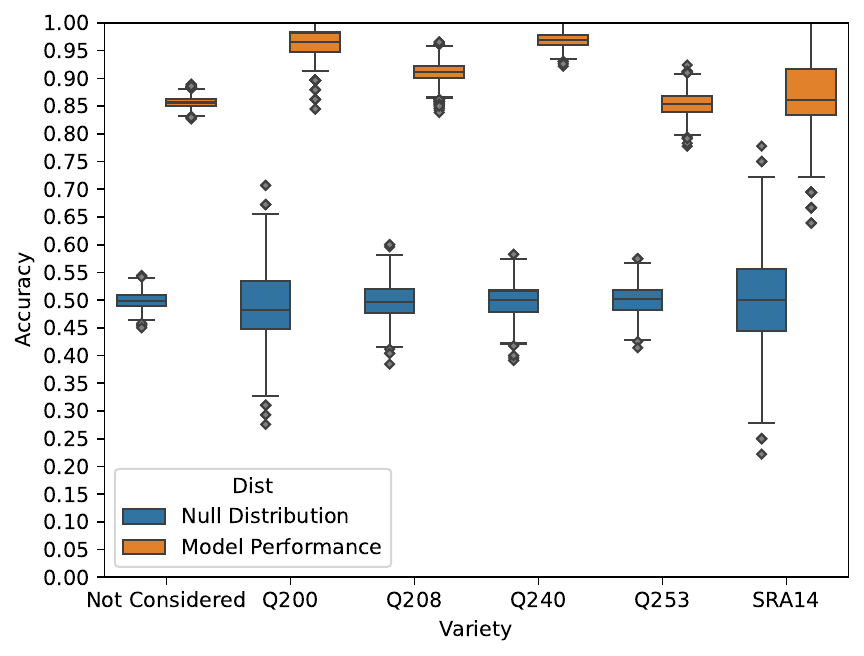}
        \caption{SVM - Accuracy}
    \end{subfigure}
    \hfill
    \begin{subfigure}[b]{0.28\textwidth}
        \includegraphics[width=\textwidth]{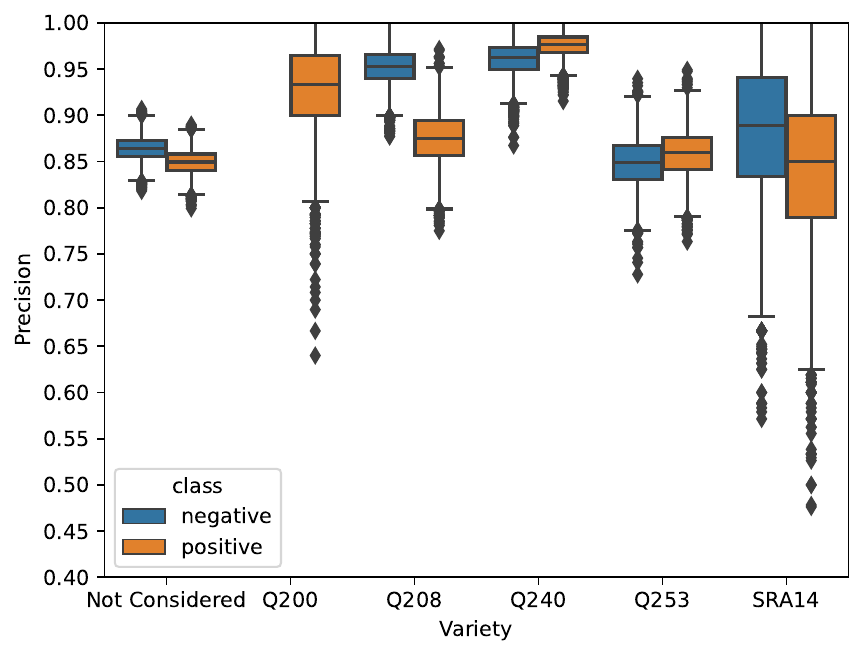}
        \caption{SVM - Precision}
    \end{subfigure}
    \hfill
    \begin{subfigure}[b]{0.28\textwidth}
        \includegraphics[width=\textwidth]{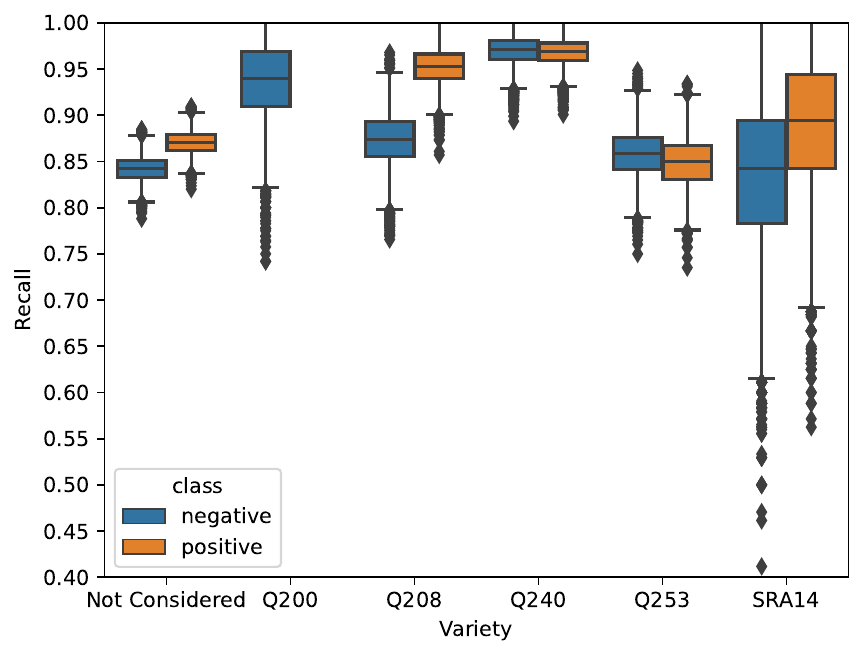}
        \caption{SVM - Recall}
    \end{subfigure}
    
    \vspace{0.5em}
    
    % Gradient Boosting
    \begin{subfigure}[b]{0.28\textwidth}
        \includegraphics[width=\textwidth]{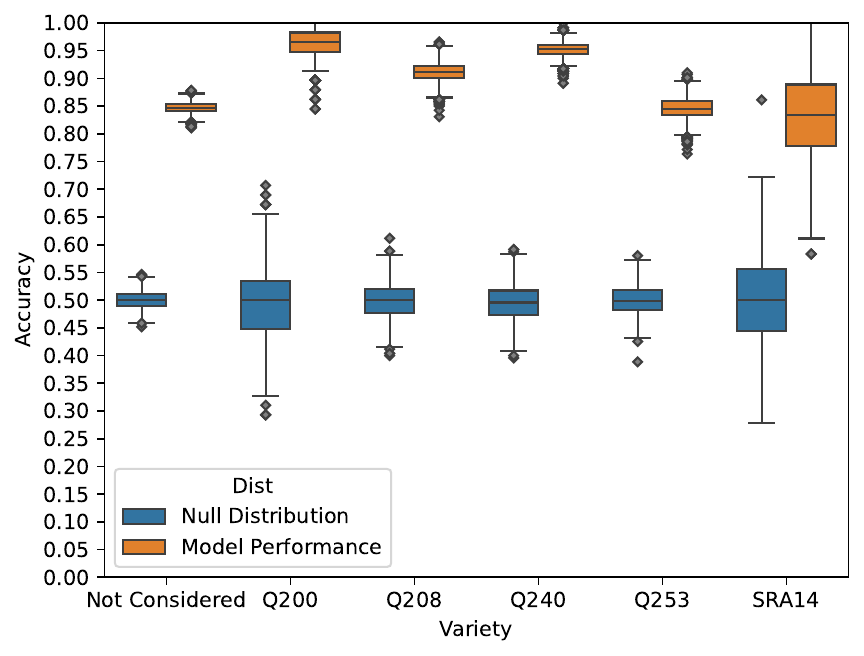}
        \caption{GB - Accuracy}
    \end{subfigure}
    \hfill
    \begin{subfigure}[b]{0.28\textwidth}
        \includegraphics[width=\textwidth]{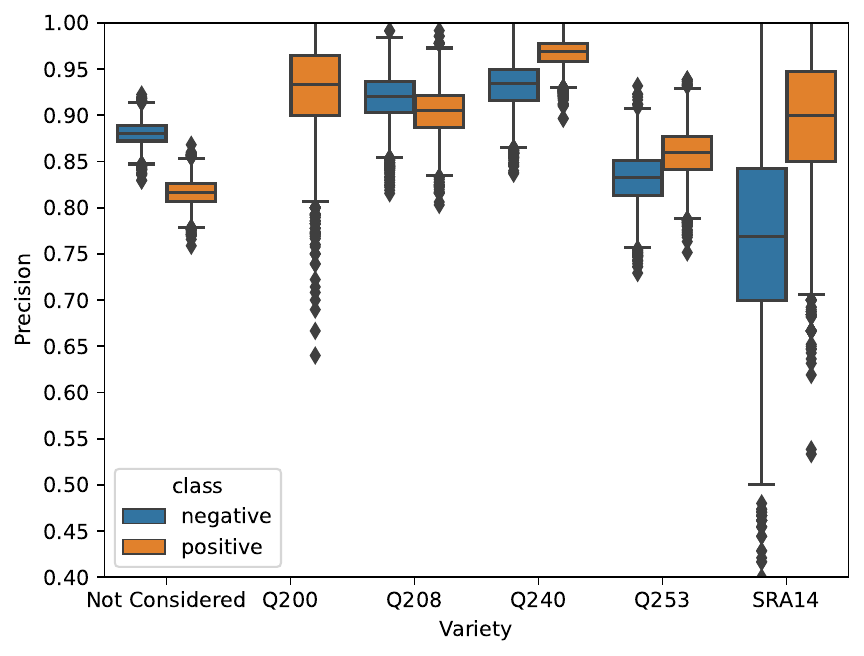}
        \caption{GB - Precision}
    \end{subfigure}
    \hfill
    \begin{subfigure}[b]{0.28\textwidth}
        \includegraphics[width=\textwidth]{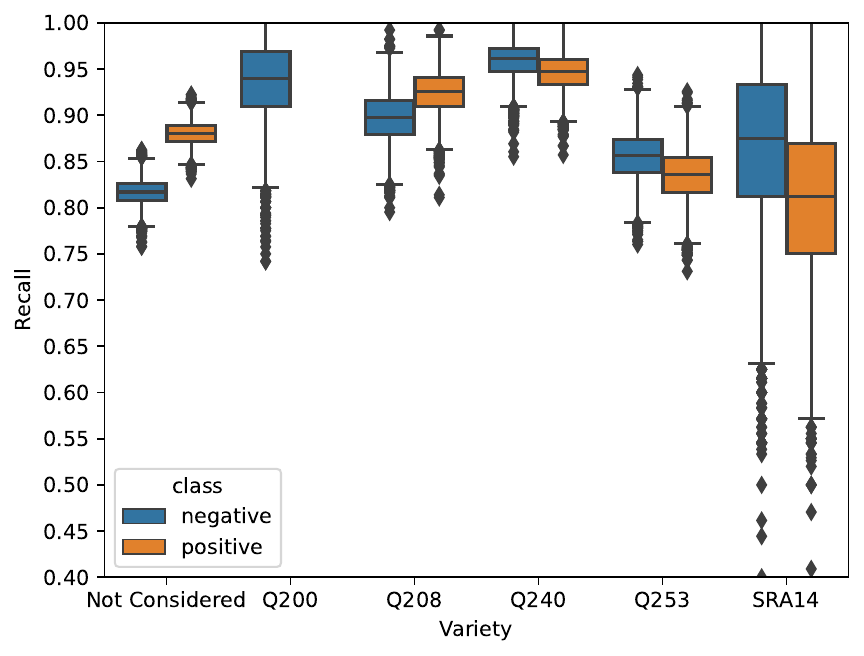}
        \caption{GB - Recall}
    \end{subfigure}
    
    \vspace{0.5em}
    
    % Random Forest
    \begin{subfigure}[b]{0.28\textwidth}
        \includegraphics[width=\textwidth]{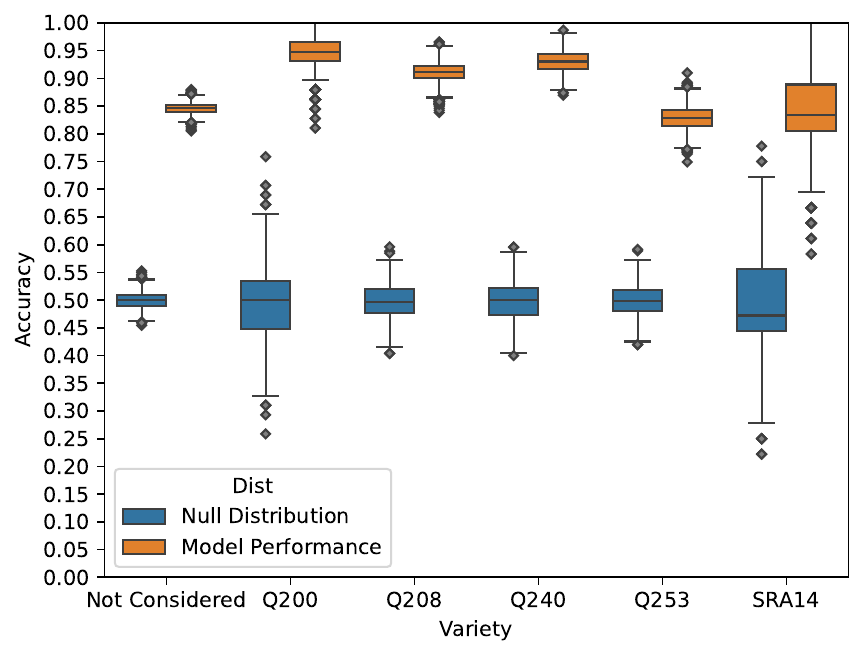}
        \caption{RF - Accuracy}
    \end{subfigure}
    \hfill
    \begin{subfigure}[b]{0.28\textwidth}
        \includegraphics[width=\textwidth]{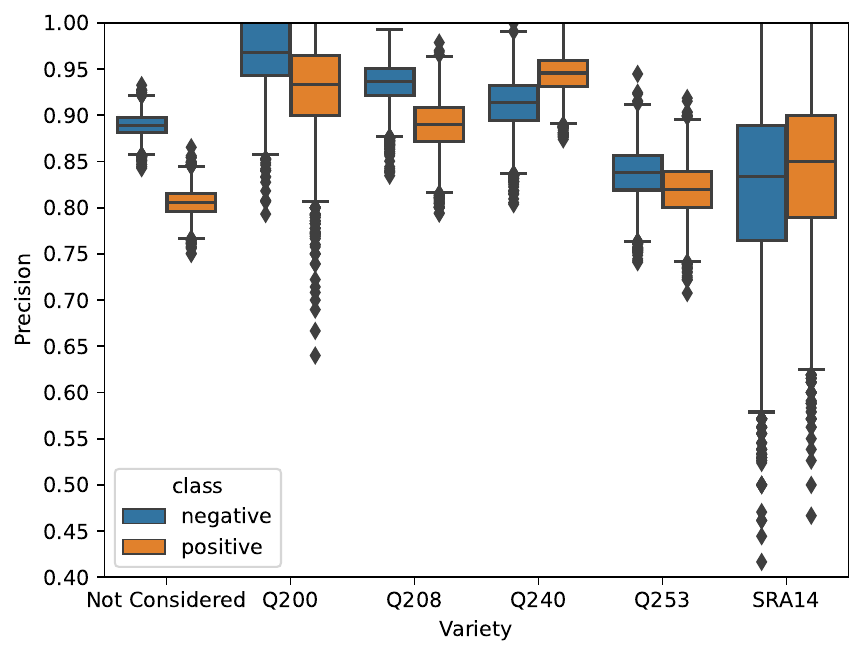}
        \caption{RF - Precision}
    \end{subfigure}
    \hfill
    \begin{subfigure}[b]{0.28\textwidth}
        \includegraphics[width=\textwidth]{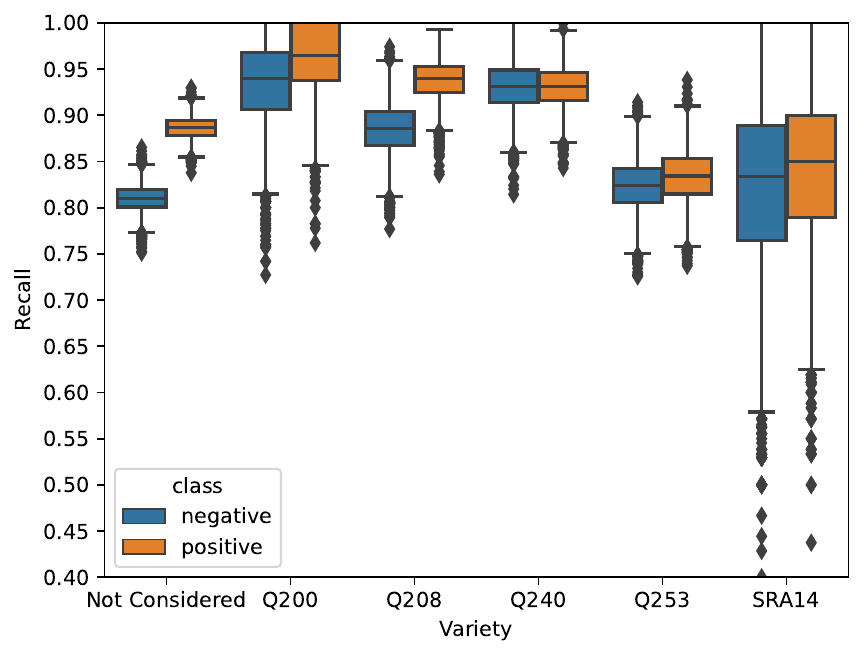}
        \caption{RF - Recall}
    \end{subfigure}
    
    \vspace{0.5em}
    
    % Quadratic Discriminant Analysis
    \begin{subfigure}[b]{0.28\textwidth}
        \includegraphics[width=\textwidth]{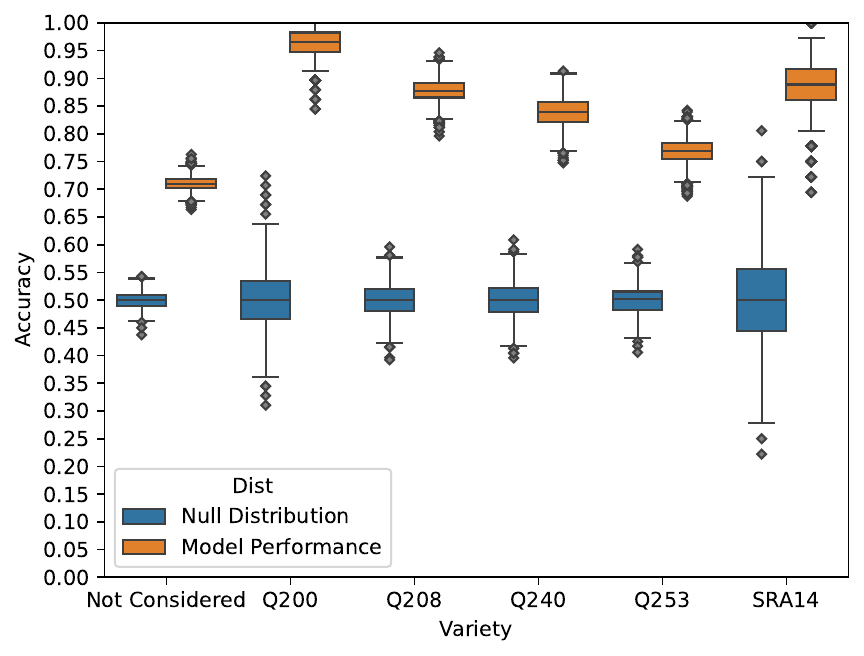}
        \caption{QDA - Accuracy}
    \end{subfigure}
    \hfill
    \begin{subfigure}[b]{0.28\textwidth}
        \includegraphics[width=\textwidth]{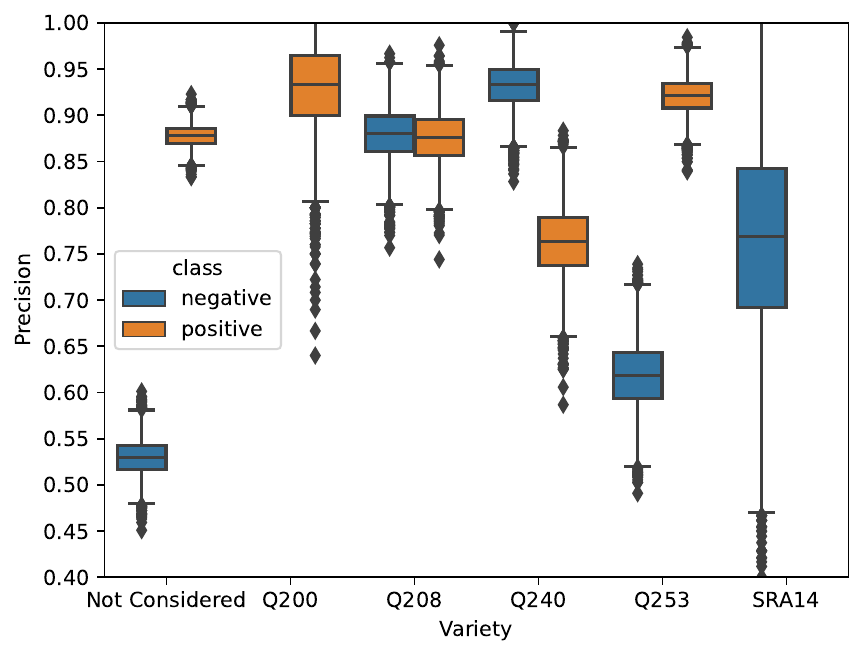}
        \caption{QDA - Precision}
    \end{subfigure}
    \hfill
    \begin{subfigure}[b]{0.28\textwidth}
        \includegraphics[width=\textwidth]{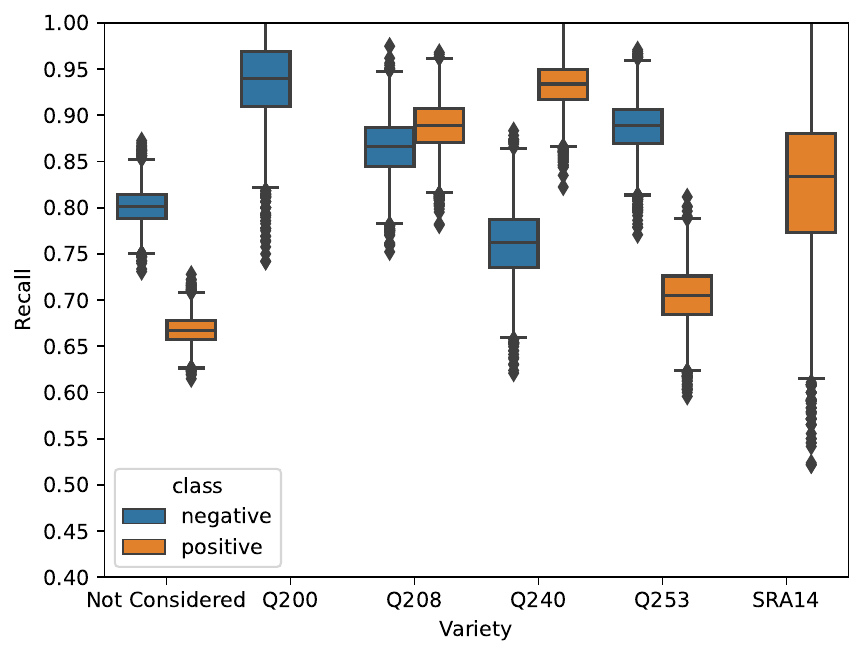}
        \caption{QDA - Recall}
    \end{subfigure}
    
    \vspace{0.5em}
    
    % Logistic Regression
    \begin{subfigure}[b]{0.28\textwidth}
        \includegraphics[width=\textwidth]{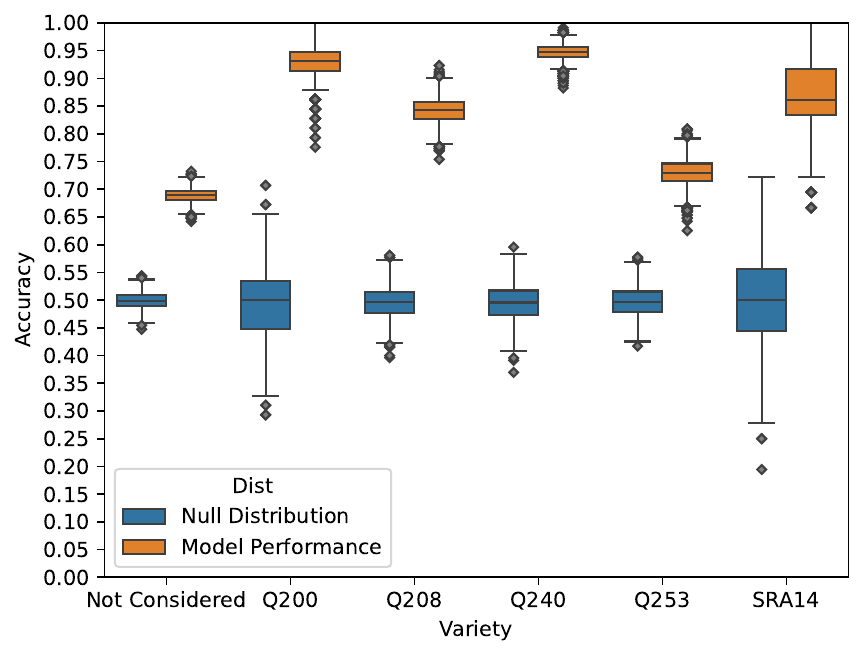}
        \caption{LR - Accuracy}
    \end{subfigure}
    \hfill
    \begin{subfigure}[b]{0.28\textwidth}
        \includegraphics[width=\textwidth]{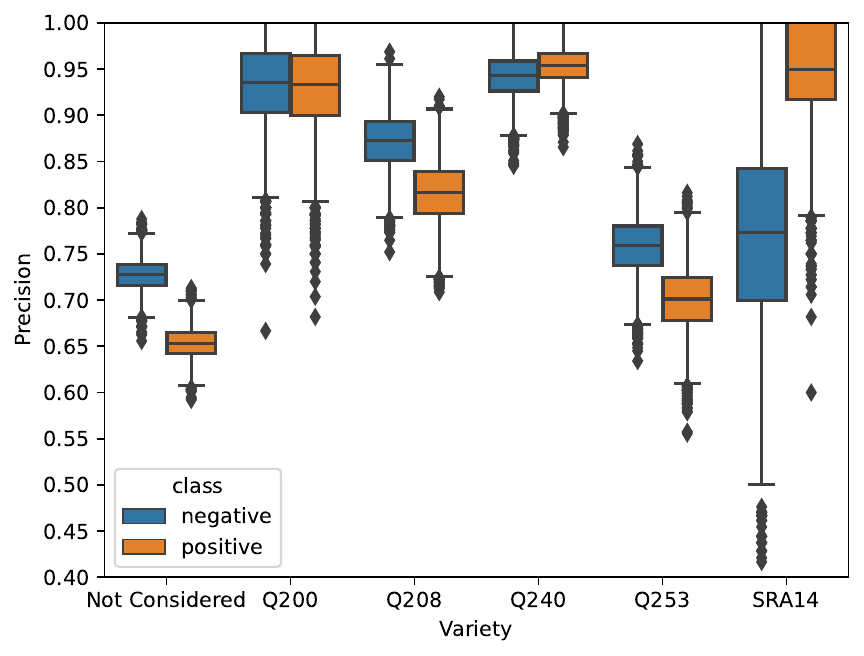}
        \caption{LR - Precision}
    \end{subfigure}
    \hfill
    \begin{subfigure}[b]{0.28\textwidth}
        \includegraphics[width=\textwidth]{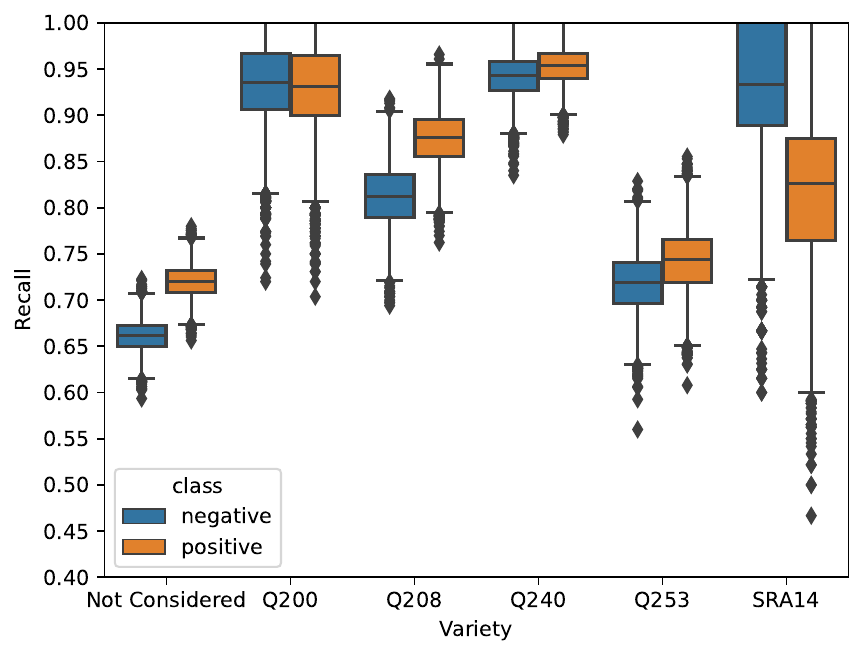}
        \caption{LR - Recall}
    \end{subfigure}

    \caption{Bootstrapping distribution of accuracy, precision, and recall for different ML algorithms diagnosing RSD in several sugarcane varieties. The accuracy distribution is compared with a null distribution generated with permutations to indicate significance of model performance for that specific variety. \label{fig3:mainfig}}
\end{figure*}

The classification and permutation testing results (Figure \ref{fig3:mainfig}) reveal a clear separation between the null distribution and the model accuracy distribution for all machine learning algorithms and varieties, with the exception of SRA14. This demonstrates that the models significantly outperformed random chance. In contrast, the overlap of distributions for SRA14 across all models suggests that its performance was not substantially different from what would be expected by chance, indicating a weaker predictive power compared to models for the other varieties.

Three machine learning algorithms demonstrated over 80\% accuracy across all varieties for RSD detection. SVM-RBF performed the best overall achieving between 85.64\% and 96.55\% accuracy depending on the variety (Figure \ref{fig3:mainfig}). SVM-RBF achieved the highest classification accuracy when variety was not considered (85.64\%) and for several specific varieties including Q208 (91.15\%), Q240 (96.96\%), and Q253 (85.35\%). Similarly, GB and RF demonstrated consistently high performance across all varieties achieving 83.33\% to 94.83\% and 83.33\% to 96.55\% accuracy respectively, depending on the variety. All three models performed particularly well for Q200, Q208 and Q240 achieving between 91.15\% and 96.55\% accuracy depending on the specific variety and model. GB and RF tied for the highest median overall accuracy on Q208, though both models demonstrated less balanced performance in terms of precision and recall when compared to SVM-RBF. Additionally, GB matched the best median overall accuracy for Q200. A complete table of results for all models and varieties can be seen in Table \ref{tab3} in Appendix A.

In contrast, LR and QDA showed varying performance depending on the specific variety. These two models performed particularly poorly for variety Q253 and underperformed compared to the other models on Q240. However, they excelled with other varieties. QDA, in particular, emerged as the top-performing model for variety SRA14, achieving a median overall accuracy of 88.89\%. Additionally, for Q200, QDA, SVM-RBF, and GB all tied as the best-performing models, each yielding identical classification reports with a median overall accuracy of 96.55\%. 

Variety-specific models for all machine learning algorithms performed as well as, or better than, the variety-agnostic models. Models specific to Q200, Q208 and Q240 outperformed the variety agnostic model, whereas those specific to Q253 and SRA14 showed no significant improvement. RSD in varieties Q200 and Q240 was consistently classified with higher accuracy compared to the other varieties. Conversely, Q253 consistently showed lower classification accuracy, while SRA14 exhibited significant variability in performance. Notably, a closer analysis of positive class precision revealed that, in 4 out of 5 algorithms, varieties Q200 and Q208, or the variety-agnostic model, yielded lower precision for the positive class.

The hyperparameters and overall accuracies of the final models trained on the entire dataset are reported in Table \ref{tab4} in Appendix B now that an appropriate estimate of performance has been determined. The outcomes largely align with the previously observed accuracies, which is expected if the test set utilised was representative of the sample distribution. In this final evaluation, there were no ties for best performance across varieties. SVM-RBF maintained its position as the top-performing model for most varieties and QDA continued to outperform other models for SRA14. 

The feature importance for diagnosing RSD revealed some variation between the GB and RF classifiers, however, both generally agreed upon several key findings. Notably, when sugarcane variety was not considered, both classifiers identified DWSI-6 as the most influential feature, and identified DWSI-7, DWSI-2, and spectral bands 2, 5, 6, and 11 among the top 10 most important features. Analysing feature importance for RSD diagnosis with separate models for each sugarcane variety indicated variations between both varieties and classifiers. No single feature appeared in the top 10 importance rankings across all varieties for both classifiers. However, across a majority of instances DWSI-6, DWSI-3, Band 2, 5, 6, 7 and 11 were the most frequently occurring features in the top 10 importance. A complete description of feature importance can be seen in Figure \ref{fig2:mainfig} in Appendix C. The t-test to identify statistically significant vegetation indices generally supports similar findings. A summary of these statistical significances is available in Appendix D.

\section{Discussion}
This study demonstrates that ML algorithms can effectively classify RSD across several varieties with freely available satellite-based multispectral data. Consistent with previous studies, RF, SVM, and GB variants demonstrated strong performance \citep{RN20, RN224, RN225}. Notably, the SVM with RBF kernel slightly outperformed both RF and GB in this context. This may be attributed to the SVM's ability to effectively model complex non-linear decision boundaries in high-dimensional settings with limited training samples. Under these conditions tree-based methods may be less effective at capturing complex relationships \citep{MPROB}. Additionally, the underlying structure of the disease-related spectral data may be more effectively separated by a smooth, non-linear decision boundary, potentially contributing to the stronger performance observed with SVM using a RBF kernel. While previous sugarcane disease detection studies have primarily focused on drone-based spectroscopy \citep{RN20, RN224, RN92}, handheld spectroscopy \citep{RN15, RN225} or hyperspectral satellite data \citep{RN18, RN19}, our findings extend this knowledge by demonstrating comparable, if not superior, effectiveness of classifying a sugarcane disease with multispectral satellite data. For example, \citet{RN224} reported classification accuracies between 86\% and 90\% for orange and brown rust with multispectral drone data. In comparison, our study achieved accuracies ranging from 83\% and 97\% accuracy for RF and SVM, depending on the variety, with freely available multispectral data. This demonstrates that our approach, utilising cost-effective large-scale satellite data, can achieve comparable or superior performance. Moreover, our study outperformed the only previous study detecting asymptomatic sugarcane disease \citep{RN15}. This represents a significant advancement in early-stage disease detection and large-scale monitoring, where timely intervention is crucial for mitigating potential crop yield loss.

Vegetation Indices (VI) and spectral bands sensitive to vegetation moisture and stress, were consistently the most influential features, reinforcing the established understanding that RSD diminishes water uptake and heightens plant stress \citep{RN11, RSD3, RSD2, RSD1}. Consistent importance of the features DWSI-7, DWSI-6, DWSI-3, Band 5, 6 and 7, clearly reflects this. The trend was particularly pronounced in the variety-agnostic models, potentially due to the variation introduced by the different sugarcane varieties, forcing the model to focus on the most generalised disease signals common across all varieties. In contrast, feature importance varied among variety-specific models, reflecting the distinct biophysical characteristics of each variety and their unique responses to disease presence. The importance of DWSI-7 and DWSI-6 across the different models highlights that adapting existing DWSI to include the other statistically significant moisture absorption band contributed to the success of these models. Interestingly, Band 2 which is centred in the blue part of the visible spectrum was considered to be consistently important, despite RSD being considered as an asymptomatic disease universally by experts. While water and stress-related indices proved valuable, popular general health indicators like the Normalized Difference Vegetation Index (NDVI) performed poorly, potentially due to the asymptomatic nature of RSD. 

Across all machine learning algorithms, variety-specific models were as or more effective than variety-agnostic ones. Performance of RSD detection varied between sugarcane varieties potentially arising from the physical differences between the varieties or external factors correlated with variety selection. Variables such as soil type, rainfall, and land topology may influence variety choice and introduce variability in the observed spectral reflectance. Consequently, future studies should incorporate soil data, rainfall, and land topology as predictors to assess their impact on RSD diagnosis and determine whether these factors explain variation that affects class separability. Notably, variety SRA14 typically exhibited lower classification rates and larger variation, compared to other varieties with model performance that was not significantly greater than random chance. This may be attributed to a smaller sample size than the other varieties (see Table \ref{tab1}) and perhaps future studies with a larger dataset will find a significant performance difference.

Despite the high classification accuracies achieved by the models, there were several limitations of the study. Positive class precision was consistently lower than negative class precision for models classifying RSD in Q200 and Q208, indicating a higher rate of false positives across the algorithms. This may have arisen from the assumption that RSD infection in a block was uniformly distributed by labeling all pixels within a block with the overall disease status of that block. This could result in regions with little or no RSD being mislabeled as 'RSD Positive', thereby affecting model training. Future studies should avoid this assumption where possible and implement a higher resolution sampling method. Interestingly, this observation coincides with the only two varieties identified by Sugar Research Australia as partially resistant to RSD \citep{RSD4, RSD5}. Partial resistance might reduce the spread of RSD sufficiently to create patches of non-infected areas within otherwise infected blocks. However, to verify this hypothesis, further research with higher resolution sampling methods would be necessary. 

This study focused on developing models and a prototype system for RSD detection within the Herbert region in February 2022. Future work should assess the influence of other environmental and agronomic predictors that could not be accounted for in this study to ensure broader applicability in the industry. This includes exploring the impacts of factors such as temperature, humidity, sunlight duration, flowering, other sugarcane varieties, simultaneous disease infections, and precipitation on the performance of classification models. Furthermore, RSD detection should be trialled at different growth stages of sugarcane using time-series data to improve its applicability to the industry. Given their proven success in previous time-series classification tasks, exploring the use of techniques such as recurrent neural networks or transformers for this application represents a promising direction for future research. Additionally, comparing various principal component analysis algorithms could further optimise vegetation index characteristics while addressing and mitigating issues such as overfitting.

\section{Conclusion}
In this study, we demonstrated that RSD can be classified for several varieties with ML algorithms from freely available satellite-based spectroscopy data. The best-performing machine learning algorithm was Support Vector Machine with a Radial Basis Kernel achieving between 85.64\% and 96.55\% accuracy depending on the variety. The performance of all machine learning algorithms was found to be significantly better than the null distribution for all varieties except SRA14. Inline with current literature, the inclusion of sugarcane variety information and vegetation indices improved classification rates of disease. Additionally, several varieties were found to be more challenging to classify for RSD compared to others. These promising initial results, coupled with the efficiency of classifying 76 blocks within minutes as opposed to months, indicates the potential benefits of implementing a large-scale health monitoring system using satellites and machine learning. Given that RSD is one of the most impactful diseases in sugarcane cultivation, this research significantly advances sugarcane disease management, with substantial practical applications. Providing farmers with a more accurate understanding of where RSD is present enables them to alter harvester paths to minimise disease spread, and source alternative clean, disease-free seeds from other blocks, or reliable suppliers, to ensure healthier crop cycles. Additionally, this will allow the implementation of strict hygiene protocols on infected blocks to prevent the disease from contaminating others. These applications indicate the potential of this research to support sustainable and efficient sugarcane cultivation.

\section{Acknowledgements}
We extend our sincere appreciation to HCPSL for their invaluable contribution in collecting and providing the ground-truth data essential for this study. Special thanks to Lawrence Di Bella, Rod Nielson, Adam Royle and Rhiannan Harragon from HCPSL for their industry knowledge and support in the ground truth data collection process.

\printcredits

\section{Funding}
This work was supported by Australia's Economic Accelerator Seed grant provided by the Australian Government, Department of Education

\clearpage
\onecolumn
\renewcommand{\thetable}{\textbf{\arabic{table}}} % Bold the table number
\renewcommand{\tablename}{\textbf{Table}} % Bold the word "Table"
\renewcommand{\figurename}{\textbf{Figure}} % Bold the word "Table"
\appendix
\renewcommand{\arraystretch}{0.58}

\section*{Appendix A}
\label{b:appendix}
\begin{center}
\noindent
\fontsize{8}{8}\selectfont
\captionsetup{type=figure}
\captionof{table}{\label{tab3} Median classification metrics of bootstrap distributions for ML algorithms.}
\begin{tabular}{p{2cm}p{2cm}p{2cm}p{2cm}p{2cm}p{2cm}}
\hline\hline
 \textbf{Model} & \textbf{Variety} & \textbf{Class} & \textbf{Precision} & \textbf{Recall} & \textbf{Accuracy}   \\
\hline
\multirow{15}{*}{SVM: RBF} & \multirow{2}{*}{Q200} & Positive & 93.33\% & 100\% &  \multirow{2}{*}{96.55\%} \\
 &  & Negative & 100\% & 93.39\% & \\
   \cmidrule(lr){2-6}
 & \multirow{2}{*}{Q208} & Positive & 87.5\% & 95.31\% & \multirow{2}{*}{91.15\%} \\
 &  & Negative & 95.31\% & 87.41\% & \\
   \cmidrule(lr){2-6}
 & \multirow{2}{*}{Q240} & Positive & 97.69\% & 96.95\% & \multirow{2}{*}{96.96\%} \\
 & & Negative & 96.22\% & 97.14\% &  \\
   \cmidrule(lr){2-6}
 & \multirow{2}{*}{Q253} & Positive & 85.95\% & 84.97\% & \multirow{2}{*}{85.35\%} \\
 &  & Negative & 84.86\% & 85.88\% & \\
   \cmidrule(lr){2-6}
 & \multirow{2}{*}{SRA14} & Positive & 85.00\% & 89.47\% & \multirow{2}{*}{86.11\%} \\
 &  & Negative & 88.88\% & 84.21\% & \\
   \cmidrule(lr){2-6}
 & \multirow{2}{*}{Not Considered} & Positive & 84.94\% & 87.06\% & \multirow{2}{*}{85.64\%} \\
 &  & Negative & 86.45\% & 84.24\% & \\
\midrule
\multirow{15}{*}{GB} & \multirow{2}{*}{Q200} & Positive & 93.33\% & 100.00\% & \multirow{2}{*}{96.55\%} \\
 &  & Negative & 100.00\% & 93.39\% &  \\
   \cmidrule(lr){2-6}
 & \multirow{2}{*}{Q208} & Positive & 90.48\% & 92.59\% & \multirow{2}{*}{91.15\%} \\
 &  & Negative & 92.06\% & 89.78\% &  \\
   \cmidrule(lr){2-6}
 & \multirow{2}{*}{Q240} & Positive & 96.90\% & 94.74\% & \multirow{2}{*}{95.22\%} \\
 &  & Negative & 93.39\% & 96.12\% &  \\
   \cmidrule(lr){2-6}
 & \multirow{2}{*}{Q253} & Positive & 85.96\% & 83.59\% & \multirow{2}{*}{84.51\%} \\
 &  & Negative & 83.23\% & 85.63\% &  \\
   \cmidrule(lr){2-6}
 & \multirow{2}{*}{SRA14} & Positive & 90.00\% & 81.25\% & \multirow{2}{*}{83.33\%} \\
 &  & Negative & 76.92\% & 87.5\% &  \\
   \cmidrule(lr){2-6}
 & \multirow{2}{*}{Not Considered} & Positive & 81.67\% & 88.03\% & \multirow{2}{*}{84.73\%} \\
 &  & Negative & 88.05\% & 81.71\% & \\
\midrule
\multirow{15}{*}{RF} & \multirow{2}{*}{Q200} & Positive & 93.33\% & 96.43\% & \multirow{2}{*}{94.83\%} \\
 &   & Negative & 96.77\% & 93.94\% &  \\
  \cmidrule(lr){2-6}
 & \multirow{2}{*}{Q208} & Positive & 88.98\% & 93.94\% & \multirow{2}{*}{91.15\%} \\
 &  & Negative & 93.66\% & 88.57\% &  \\
  \cmidrule(lr){2-6}
 & \multirow{2}{*}{Q240} & Positive & 94.57\% & 93.13\% & \multirow{2}{*}{93.04\%} \\
 &  & Negative & 91.43\% & 93.13\% & \\
  \cmidrule(lr){2-6}
 & \multirow{2}{*}{Q253} & Positive & 82.01\% & 83.43\% & \multirow{2}{*}{82.82\%} \\
 &  & Negative & 83.82\% & 82.42\% &  \\
  \cmidrule(lr){2-6}
 & \multirow{2}{*}{SRA14} & Positive & 85.00\% & 85.00\% & \multirow{2}{*}{83.33\%} \\
 &  & Negative & 83.33\% & 83.33\% &  \\
  \cmidrule(lr){2-6}
 & \multirow{2}{*}{Not Considered} & Positive & 80.58\% & 88.67\% &\multirow{2}{*}{84.59\%} \\
 &  & Negative & 88.92\% & 80.99\% &  \\
\midrule
\multirow{15}{*}{QDA} & \multirow{2}{*}{Q200} & Positive & 93.33\% & 100.00\% & \multirow{2}{*}{96.55\%} \\
 &  & Negative & 100.00\% & 93.39\% &  \\
 \cmidrule(lr){2-6}
 & \multirow{2}{*}{Q208} & Positive & 87.59\% & 88.89\% & \multirow{2}{*}{87.69\%} \\
 & & Negative & 88.00\% & 86.61\% &  \\
 \cmidrule(lr){2-6}
 & \multirow{2}{*}{Q240} & Positive & 76.38\% & 93.39\% & \multirow{2}{*}{83.91\%} \\
 &  & Negative & 93.33\% & 76.19\% &  \\
 \cmidrule(lr){2-6}
 & \multirow{2}{*}{Q253} & Positive & 92.18\% & 70.54\% & \multirow{2}{*}{76.90\%} \\
 &  & Negative & 61.80\% & 88.89\% &  \\
 \cmidrule(lr){2-6}
 & \multirow{2}{*}{SRA14} & Positive & 100.00\% & 83.33\% & \multirow{2}{*}{88.89\%} \\
 & & Negative & 76.92 \% & 100.00\% &  \\
 \cmidrule(lr){2-6}
 & \multirow{2}{*}{Not Considered} & Positive & 87.79\% & 66.74\% & \multirow{2}{*}{71.01\%} \\
 & & Negative & 52.99\% & 80.17\% &  \\
\midrule
\multirow{15}{*}{LR} & \multirow{2}{*}{Q200} & Positive & 93.33\% & 93.10\% & \multirow{2}{*}{93.10\%}\\
 &  & Negative & 93.55\% & 93.55\% & \\
  \cmidrule(lr){2-6}
 & \multirow{2}{*}{Q208} & Positive & 81.69\% & 87.60\% & \multirow{2}{*}{84.23\%} \\
 &  & Negative & 87.29\% & 81.20\% &  \\
  \cmidrule(lr){2-6}
 & \multirow{2}{*}{Q240} & Positive & 95.39\% & 95.35\% & \multirow{2}{*}{94.78\%} \\
 & & Negative & 94.29\% & 94.29\% &  \\
  \cmidrule(lr){2-6}
 & \multirow{2}{*}{Q253} & Positive & 70.12\% & 74.37\% & \multirow{2}{*}{72.96\%} \\
 &  & Negative & 75.90\% & 71.88\% &  \\
  \cmidrule(lr){2-6}
 & \multirow{2}{*}{SRA14} & Positive & 95.00\% & 82.61\% & \multirow{2}{*}{86.11\%} \\
 & & Negative & 77.27\% & 93.33\% & \\
  \cmidrule(lr){2-6}
 & \multirow{2}{*}{Not Considered} & Positive & 65.33\% & 72.03\% & \multirow{2}{*}{68.90\%} \\
 & & Negative & 72.73\% & 66.13\% & \\
\hline\hline
\end{tabular}
\end{center}

\newpage
\section*{Appendix B}
\begin{center}
\noindent
\fontsize{8}{8}\selectfont
\captionsetup{type=figure}
\captionof{table}{\label{tab4} Optimal hyper-parameters and mean classification accuracy for different machine learning algorithms and varieties determined with 10-fold cross-validation across the entire dataset.}
\begin{tabular}{p{2cm}p{2cm}p{7cm}p{2cm}}
\hline\hline
 \textbf{Model} & \textbf{Variety} & \textbf{Optimal hyper-parameters} & \textbf{Accuracy}   \\
\hline
\multirow{6}{*}{SVM: RBF} & Q200 & 'C': 1000, 'gamma': 0.001 & 99.17\% \\

 & Q208 & 'C': 1000, 'gamma': 0.01 & 90.14\% \\
 & Q240 & 'C': 10, 'gamma': 0.1 & 96.72\%  \\
 & Q253 & 'C': 10, 'gamma': 1 & 85.39\% \\
 & SRA14 & 'C': 1000, 'gamma': 0.01 & 89.02\% \\
 & Not Considered & 'C': 10, 'gamma': 0.1 & 84.59\% \\
\midrule
\multirow{6}{*}{GB} & Q200 & 'learning rate': 0.1, 'max depth': 3, 'estimators': 200 & 95.00\% \\
 & Q208 & 'learning rate': 0.2, 'max depth': 3, 'estimators': 500 & 88.68\% \\
 & Q240 & 'learning rate': 0.3, 'max depth': 3, 'estimators': 1000 & 95.17\%  \\
 & Q253 & 'learning rate': 0.1, 'max depth': 5, 'estimators': 500 & 82.46\% \\
 & SRA14 & 'learning rate': 0.3, 'max depth': 3, 'estimators': 500 & 87.04\% \\
 & Not Considered & 'learning rate': 0.1, 'max depth': 5, 'estimators': 1500 & 84.17\% \\
\midrule
\multirow{6}{*}{RF} & Q200 & 'max depth': None, 'estimators': 200 & 95.83\% \\
 & Q208 & 'max depth': 30, 'estimators': 1000 & 89.75\% \\
 & Q240 & 'max depth': 30, 'estimators': 200 & 94.81\%  \\
 & Q253 & 'max depth': 20, 'estimators': 1000 & 85.19\% \\
 & SRA14 & 'max depth': None, 'estimators': 200 & 88.79\% \\
 & Not Considered & 'max depth': 20, 'estimators': 200 & 84.26\% \\
\midrule
\multirow{6}{*}{QDA} & Q200 & N/A & 97.24\% \\
 & Q208 & N/A & 89.13\% \\
 & Q240 & N/A & 87.09\%  \\
 & Q253 & N/A & 76.58\% \\
 & SRA14 & N/A & 92.71\% \\
 & Not Considered & N/A & 69.46\% \\
\midrule
\multirow{6}{*}{LR} & Q200 & 'C': 100, 'penalty': 'l1', 'solver': 'liblinear' & 93.33\% \\
 & Q208 & 'C': 100, 'penalty': 'l1', 'solver': 'liblinear' & 84.31\% \\
 & Q240 & 'C': 100, 'penalty': 'l1', 'solver': 'liblinear' & 95.24\%  \\
 & Q253 & 'C': 100, 'penalty': 'l1', 'solver': 'liblinear' & 74.34\% \\
 & SRA14 & 'C': 100, 'penalty': 'l1', 'solver': 'liblinear' & 82.19\% \\
 & Not Considered & 'C': 100, 'penalty': 'l1', 'solver': 'liblinear' & 68.73\% \\
\hline\hline
\end{tabular}
\end{center}

\newpage
\FloatBarrier
\section*{Appendix C}

\noindent

\begin{figure}[h!]
\begin{center}
\begin{subfigure}[b]{\textwidth}  
    \includegraphics[width=\textwidth, height=0.58\textwidth]{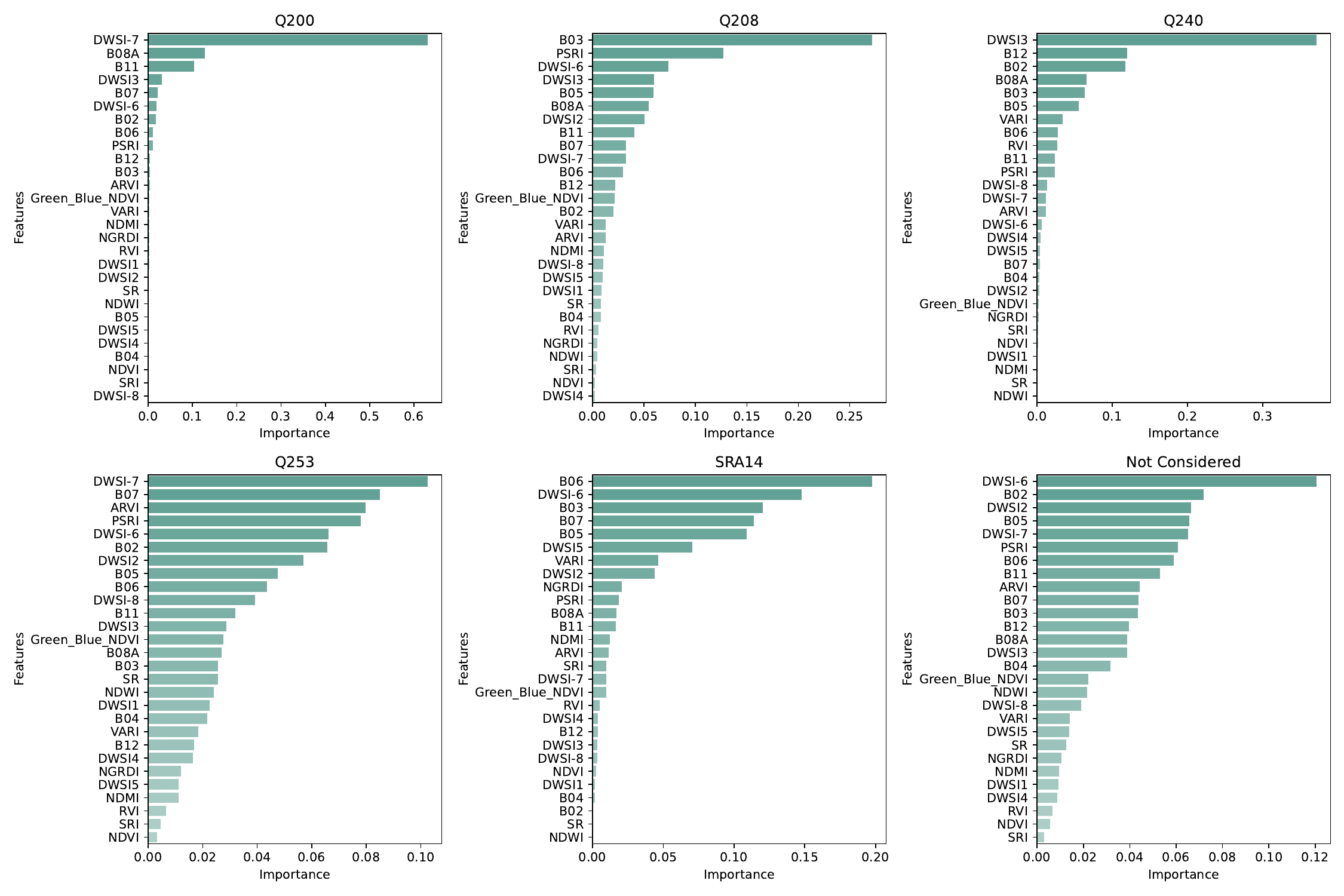}
    \caption{GB classifier}
    \label{fig2:subfig1}
\end{subfigure}
\vspace{1em}  
\noindent
\begin{subfigure}[b]{\textwidth} 
    \includegraphics[width=\textwidth, height=0.58\textwidth]{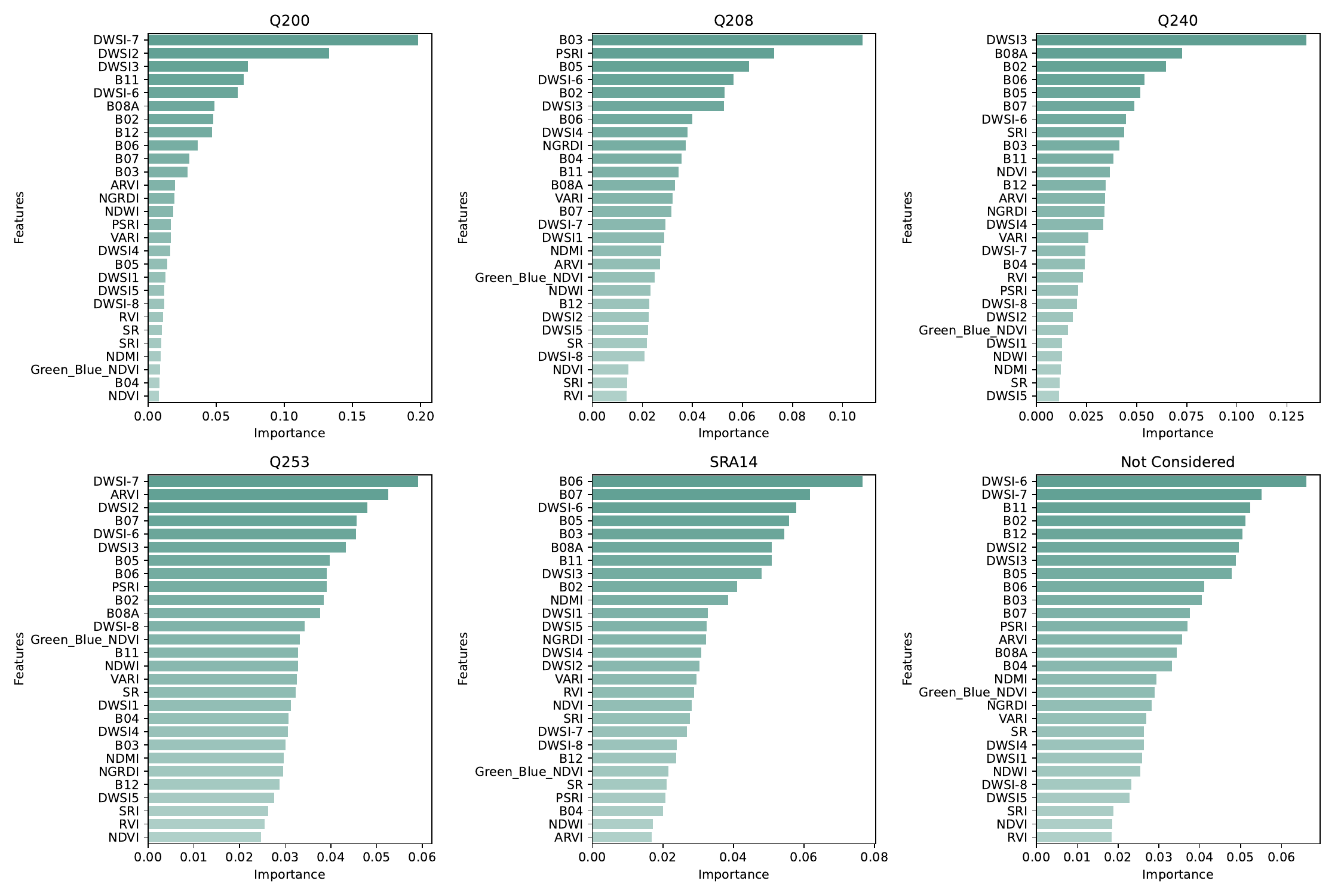}
    \caption{RF classifier}
    \label{fig2:subfig2}
\end{subfigure}

% Caption for the entire figure
\captionsetup{justification=raggedright} % Left-align the caption
\captionof{figure}{Feature importance of different sugarcane varieties}
\label{fig2:mainfig}
\end{center}
\end{figure}

\newpage
\FloatBarrier
\section*{Appendix D}
\label{D:appendix}
\begin{center}
\noindent
\fontsize{8}{8}\selectfont
\captionsetup{type=figure}
\captionof{table}{\label{statSum} Summary of statistical significance from t-tests comparing healthy and diseased samples across different sugarcane varieties.'.': adjusted p-value > 0.05, '*': 0.05 >= adjusted p-value > 0.01, '**': adjusted p-value <= 0.01}
\begin{tabular}{lccccc}
\hline
\textbf{Vegetation Index} & \textbf{Q200} & \textbf{Q208} & \textbf{Q240} & \textbf{Q253} & \textbf{SRA14} \\
\hline
ARVI               & .   & .   & **  & **  & .   \\
DWSI1              & .   & *   & **  & **  & .   \\
DWSI2              & **  & .   & .   & **  & .   \\
DWSI3              & **  & **  & **  & .   & **  \\
DWSI4              & .   & .   & **  & **  & .   \\
DWSI5              & .   & .   & **  & **  & .   \\
NDMI               & .   & .   & **  & **  & .   \\
NDVI               & .   & .   & **  & **  & .   \\
NDWI               & .   & .   & **  & **  & .   \\
NGRDI              & .   & .   & **  & **  & .   \\
PSRI               & .   & **  & **  & .   & .   \\
RVI                & .   & .   & **  & **  & .   \\
SRI                & .   & .   & **  & **  & .   \\
SR                 & .   & .   & **  & **  & .   \\
VARI               & .   & .   & **  & **  & .   \\
Green Blue NDVI  & .   & .   & **  & **  & .   \\
DWSI 6             & **  & **  & **  & .   & *   \\
DWSI 7             & **  & **  & **  & .   & .   \\
DWSI 8             & .   & .   & **  & **  & .   \\
\hline\hline
\end{tabular}
\end{center}

\twocolumn

%% Loading bibliography style file
% \bibliographystyle{model1-num-names}
\bibliographystyle{cas-model2-names}

% Loading bibliography database
\bibliography{cas-refs}

\begin{thebibliography}{51}
\expandafter\ifx\csname natexlab\endcsname\relax\def\natexlab#1{#1}\fi
\providecommand{\url}[1]{\texttt{#1}}
\providecommand{\href}[2]{#2}
\providecommand{\path}[1]{#1}
\providecommand{\DOIprefix}{ }
\providecommand{\ArXivprefix}{arXiv:}
\providecommand{\URLprefix}{URL: }
\providecommand{\Pubmedprefix}{pmid:}
\providecommand{\doi}[1]{\href{http://dx.doi.org/#1}{\path{#1}}}
\providecommand{\Pubmed}[1]{\href{pmid:#1}{\path{#1}}}
\providecommand{\bibinfo}[2]{#2}
\ifx\xfnm\relax \def\xfnm[#1]{\unskip,\space#1}\fi
%Type = Article
\bibitem[{Abdel-Rahman et~al.(2010)Abdel-Rahman, Ahmed, van~den Berg and
  Way}]{RN96}
\bibinfo{author}{Abdel-Rahman, E.}, \bibinfo{author}{Ahmed, F.},
  \bibinfo{author}{van~den Berg, M.}, \bibinfo{author}{Way, M.},
  \bibinfo{year}{2010}.
\newblock \bibinfo{title}{Potential of spectroscopic data sets for sugarcane
  thrips (fulmekiola serrata kobus) damage detection}.
\newblock \bibinfo{journal}{International Journal of Remote Sensing}
  \bibinfo{volume}{31}, \bibinfo{pages}{4199--4216}.
\newblock \DOIprefix\doi{10.1080/01431160903241981}.
%Type = Article
\bibitem[{Apan et~al.(2003)Apan, Held, Phinn and Markley}]{RN19}
\bibinfo{author}{Apan, A.}, \bibinfo{author}{Held, A.}, \bibinfo{author}{Phinn,
  S.}, \bibinfo{author}{Markley, J.}, \bibinfo{year}{2003}.
\newblock \bibinfo{title}{Formulation and assessment of narrow-band vegetation
  indices from eo1 hyperion imagery for discriminating sugarcane disease}.
\newblock \bibinfo{journal}{Proceedings of the Spatial Sciences Conference} .
%Type = Article
\bibitem[{Apan et~al.(2004)Apan, Held, Phinn and Markley}]{RN18}
\bibinfo{author}{Apan, A.}, \bibinfo{author}{Held, A.}, \bibinfo{author}{Phinn,
  S.}, \bibinfo{author}{Markley, J.}, \bibinfo{year}{2004}.
\newblock \bibinfo{title}{Detecting sugarcane ‘orange rust’ disease using
  eo-1 hyperion hyperspectral imagery}.
\newblock \bibinfo{journal}{International Journal of Remote Sensing}
  \bibinfo{volume}{25}, \bibinfo{pages}{489--498}.
\newblock \DOIprefix\doi{10.1080/01431160310001618031}.
%Type = Article
\bibitem[{Bailey and Bechet(1986)}]{RN11}
\bibinfo{author}{Bailey, R.}, \bibinfo{author}{Bechet, G.},
  \bibinfo{year}{1986}.
\newblock \bibinfo{title}{Effect of ratoon stunting disease on the yield and
  components of yield of sugarcane under rainfed conditions}.
\newblock \bibinfo{journal}{Proceedings of the South African Sugar
  Technologists Association} \bibinfo{volume}{60}, \bibinfo{pages}{204--210}.
%Type = Article
\bibitem[{Bao et~al.(2024)Bao, Zhou, Bhuiyan, Adhikari, Tuxworth, Ford and
  Gao}]{Bao2}
\bibinfo{author}{Bao, D.}, \bibinfo{author}{Zhou, J.},
  \bibinfo{author}{Bhuiyan, S.A.}, \bibinfo{author}{Adhikari, P.},
  \bibinfo{author}{Tuxworth, G.}, \bibinfo{author}{Ford, R.},
  \bibinfo{author}{Gao, Y.}, \bibinfo{year}{2024}.
\newblock \bibinfo{title}{Early detection of sugarcane smut and mosaic diseases
  via hyperspectral imaging and spectral-spatial attention deep neural
  networks}.
\newblock \bibinfo{journal}{Journal of Agriculture and Food Research}
  \bibinfo{volume}{18}, \bibinfo{pages}{101369}.
\newblock \URLprefix
  \url{https://www.sciencedirect.com/science/article/pii/S266615432400406X},
  \DOIprefix\doi{https://doi.org/10.1016/j.jafr.2024.101369}.
%Type = Article
\bibitem[{Bao et~al.(2021)Bao, Zhou, Bhuiyan, Zia, Ford and Gao}]{RN479}
\bibinfo{author}{Bao, D.}, \bibinfo{author}{Zhou, J.},
  \bibinfo{author}{Bhuiyan, S.A.}, \bibinfo{author}{Zia, A.},
  \bibinfo{author}{Ford, R.}, \bibinfo{author}{Gao, Y.}, \bibinfo{year}{2021}.
\newblock \bibinfo{title}{Early detection of sugarcane smut disease in
  hyperspectral images}.
\newblock \bibinfo{journal}{2021 36th International Conference on Image and
  Vision Computing New Zealand (IVCNZ)} ,
  \bibinfo{pages}{1--6}\DOIprefix\doi{10.1109/IVCNZ54163.2021.9653386}.
%Type = Article
\bibitem[{Carvalho et~al.(2016)Carvalho, {da Silva}, Munhoz,
  Monteiro-Vitorello, Azevedo, Melotto and Camargo}]{RN477}
\bibinfo{author}{Carvalho, G.}, \bibinfo{author}{{da Silva}, T.},
  \bibinfo{author}{Munhoz, A.}, \bibinfo{author}{Monteiro-Vitorello, C.},
  \bibinfo{author}{Azevedo, R.}, \bibinfo{author}{Melotto, M.},
  \bibinfo{author}{Camargo, L.}, \bibinfo{year}{2016}.
\newblock \bibinfo{title}{Development of a qpcr for leifsonia xyli subsp. xyli
  and quantification of the effects of heat treatment of sugarcane cuttings on
  lxx}.
\newblock \bibinfo{journal}{Crop Protection} \bibinfo{volume}{80},
  \bibinfo{pages}{51--55}.
\newblock \DOIprefix\doi{https://doi.org/10.1016/j.cropro.2015.10.029}.
%Type = Article
\bibitem[{Chakraborty et~al.(2024)Chakraborty, Soda, Strachan, Ngo, Bhuiyan,
  Shiddiky and Ford}]{RSD3}
\bibinfo{author}{Chakraborty, M.}, \bibinfo{author}{Soda, N.},
  \bibinfo{author}{Strachan, S.}, \bibinfo{author}{Ngo, C.N.},
  \bibinfo{author}{Bhuiyan, S.A.}, \bibinfo{author}{Shiddiky, M.J.A.},
  \bibinfo{author}{Ford, R.}, \bibinfo{year}{2024}.
\newblock \bibinfo{title}{Ratoon stunting disease of sugarcane: A review
  emphasizing detection strategies and challenges}.
\newblock \bibinfo{journal}{Phytopathology®} \bibinfo{volume}{114},
  \bibinfo{pages}{7--20}.
\newblock \URLprefix \url{https://doi.org/10.1094/PHYTO-05-23-0181-RVW},
  \DOIprefix\doi{10.1094/PHYTO-05-23-0181-RVW}. \bibinfo{note}{pMID: 37530477}.
%Type = Book
\bibitem[{Croft et~al.(2000)Croft, Magarey and Whittle}]{RSD1}
\bibinfo{author}{Croft, B.}, \bibinfo{author}{Magarey, R.},
  \bibinfo{author}{Whittle, P.}, \bibinfo{year}{2000}.
\newblock \bibinfo{title}{Manual of Canegrowing}.
\newblock \bibinfo{publisher}{BSES}.
%Type = Article
\bibitem[{Datt(1998)}]{VI1}
\bibinfo{author}{Datt, B.}, \bibinfo{year}{1998}.
\newblock \bibinfo{title}{Remote sensing of chlorophyll a, chlorophyll b,
  chlorophyll a+b, and total carotenoid content in eucalyptus leaves}.
\newblock \bibinfo{journal}{Remote Sensing of Environment}
  \bibinfo{volume}{66}, \bibinfo{pages}{111--121}.
\newblock \URLprefix
  \url{https://www.sciencedirect.com/science/article/pii/S0034425798000467},
  \DOIprefix\doi{https://doi.org/10.1016/S0034-4257(98)00046-7}.
%Type = Book
\bibitem[{Davis and Bailey(2000)}]{RSD2}
\bibinfo{author}{Davis, M.J.}, \bibinfo{author}{Bailey, R.A.},
  \bibinfo{year}{2000}.
\newblock \bibinfo{title}{A guide to sugarcane diseases}.
\newblock \bibinfo{publisher}{CIRAD and ISSCT}.
%Type = Article
\bibitem[{Davis et~al.(1984)Davis, Gillaspie, Vidaver and Harris}]{RN474}
\bibinfo{author}{Davis, M.J.}, \bibinfo{author}{Gillaspie, A.G.},
  \bibinfo{author}{Vidaver, A.K.}, \bibinfo{author}{Harris, R.W.},
  \bibinfo{year}{1984}.
\newblock \bibinfo{title}{Clavibacter: a new genus containing some
  phytopathogenic coryneform bacteria, including clavibacter xyli subsp. xyli
  sp. nov., subsp. nov. and clavibacter xyli subsp. cynodontis subsp. nov.,
  pathogens that cause ratoon stunting disease of sugarcane and bermudagrass
  stunting disease†}.
\newblock \bibinfo{journal}{International Journal of Systematic and
  Evolutionary Microbiology} \bibinfo{volume}{34}, \bibinfo{pages}{107--117}.
\newblock \DOIprefix\doi{https://doi.org/10.1099/00207713-34-2-107}.
%Type = Misc
\bibitem[{{European Space Agency}()}]{RN46}
\bibinfo{author}{{European Space Agency}}, .
\newblock \bibinfo{title}{Sentinel-2 overview}.
\newblock
  \bibinfo{howpublished}{\url{https://sentinels.copernicus.eu/web/sentinel/missions/sentinel-2/overview}}.
\newblock \bibinfo{note}{Accessed: 01/05/2022}.
%Type = Misc
\bibitem[{Fang and Liang(2014)}]{RN71}
\bibinfo{author}{Fang, H.}, \bibinfo{author}{Liang, S.}, \bibinfo{year}{2014}.
\newblock \bibinfo{title}{Leaf area index models}.
\newblock \DOIprefix\doi{https://doi.org/10.1016/B978-0-12-409548-9.09076-X}.
%Type = Article
\bibitem[{Fegan et~al.(1998)Fegan, Croft, Teakle, Hayward and Smith}]{RN478}
\bibinfo{author}{Fegan, M.}, \bibinfo{author}{Croft, B.J.},
  \bibinfo{author}{Teakle, D.S.}, \bibinfo{author}{Hayward, A.C.},
  \bibinfo{author}{Smith, G.R.}, \bibinfo{year}{1998}.
\newblock \bibinfo{title}{Sensitive and specific detection of clavibacter xyli
  subsp. xyli, causal agent of ratoon stunting disease of sugarcane, with a
  polymerase chain reaction-based assay}.
\newblock \bibinfo{journal}{Plant Pathology} \bibinfo{volume}{47},
  \bibinfo{pages}{495--504}.
\newblock \DOIprefix\doi{https://doi.org/10.1046/j.1365-3059.1998.00255.x}.
%Type = Article
\bibitem[{Fensholt and Sandholt(2003)}]{RN500}
\bibinfo{author}{Fensholt, R.}, \bibinfo{author}{Sandholt, I.},
  \bibinfo{year}{2003}.
\newblock \bibinfo{title}{Derivation of a shortwave infrared water stress index
  from modis near- and shortwave infrared data in a semiarid environment}.
\newblock \bibinfo{journal}{Remote Sensing of Environment}
  \bibinfo{volume}{87}, \bibinfo{pages}{111--121}.
\newblock \URLprefix
  \url{https://www.sciencedirect.com/science/article/pii/S0034425703001895},
  \DOIprefix\doi{https://doi.org/10.1016/j.rse.2003.07.002}.
%Type = Article
\bibitem[{Gao(1996)}]{RN79}
\bibinfo{author}{Gao, B.c.}, \bibinfo{year}{1996}.
\newblock \bibinfo{title}{Ndwi—a normalized difference water index for remote
  sensing of vegetation liquid water from space}.
\newblock \bibinfo{journal}{Remote Sensing of Environment}
  \bibinfo{volume}{58}, \bibinfo{pages}{257--266}.
\newblock \DOIprefix\doi{https://doi.org/10.1016/S0034-4257(96)00067-3}.
%Type = Article
\bibitem[{Genc et~al.(2008)Genc, Genc, Turhan, Smith and Nation}]{RN75}
\bibinfo{author}{Genc, H.}, \bibinfo{author}{Genc, L.},
  \bibinfo{author}{Turhan, H.}, \bibinfo{author}{Smith, S.},
  \bibinfo{author}{Nation, J.}, \bibinfo{year}{2008}.
\newblock \bibinfo{title}{Vegetation indices as indicators of damage by the
  sunn pest (hemiptera: Scutelleridae) to field grown wheat}.
\newblock \bibinfo{journal}{African Journal of Biotechnology}
  \bibinfo{volume}{7}.
%Type = Article
\bibitem[{Ghai et~al.(2014)Ghai, Singh, Martin, McFarlane, van Antwerpen and
  Rutherford}]{RN476}
\bibinfo{author}{Ghai, M.}, \bibinfo{author}{Singh, V.},
  \bibinfo{author}{Martin, L.}, \bibinfo{author}{McFarlane, S.},
  \bibinfo{author}{van Antwerpen, T.}, \bibinfo{author}{Rutherford, R.},
  \bibinfo{year}{2014}.
\newblock \bibinfo{title}{{A rapid and visual loop‐mediated isothermal
  amplification assay to detect Leifsonia xyli subsp. xyli targeting a
  transposase gene}}.
\newblock \bibinfo{journal}{Letters in Applied Microbiology}
  \bibinfo{volume}{59}, \bibinfo{pages}{648--657}.
\newblock \DOIprefix\doi{10.1111/lam.12327}.
%Type = Article
\bibitem[{Gitelson et~al.(2002)Gitelson, Kaufman, Stark and Rundquist}]{RN82}
\bibinfo{author}{Gitelson, A.A.}, \bibinfo{author}{Kaufman, Y.J.},
  \bibinfo{author}{Stark, R.}, \bibinfo{author}{Rundquist, D.},
  \bibinfo{year}{2002}.
\newblock \bibinfo{title}{Novel algorithms for remote estimation of vegetation
  fraction}.
\newblock \bibinfo{journal}{Remote sensing of Environment}
  \bibinfo{volume}{80}, \bibinfo{pages}{76--87}.
%Type = Article
\bibitem[{Grisham et~al.(2010)Grisham, Johnson and Zimba}]{RN15}
\bibinfo{author}{Grisham, M.P.}, \bibinfo{author}{Johnson, R.M.},
  \bibinfo{author}{Zimba, P.V.}, \bibinfo{year}{2010}.
\newblock \bibinfo{title}{Detecting sugarcane yellow leaf virus infection in
  asymptomatic leaves with hyperspectral remote sensing and associated leaf
  pigment changes}.
\newblock \bibinfo{journal}{J Virol Methods} \bibinfo{volume}{167},
  \bibinfo{pages}{140--5}.
\newblock \DOIprefix\doi{10.1016/j.jviromet.2010.03.024}.
%Type = Inproceedings
\bibitem[{Jamieson and Talwalkar(2016)}]{HGS2}
\bibinfo{author}{Jamieson, K.}, \bibinfo{author}{Talwalkar, A.},
  \bibinfo{year}{2016}.
\newblock \bibinfo{title}{Non-stochastic best arm identification and
  hyperparameter optimization}, in: \bibinfo{booktitle}{Artificial intelligence
  and statistics}, \bibinfo{organization}{PMLR}. pp. \bibinfo{pages}{240--248}.
%Type = Inproceedings
\bibitem[{Johansen et~al.(2014)Johansen, Robson, Samson, Sallam, Chandler,
  Eaton, Derby and Jennings}]{RN95}
\bibinfo{author}{Johansen, K.}, \bibinfo{author}{Robson, A.},
  \bibinfo{author}{Samson, P.}, \bibinfo{author}{Sallam, N.},
  \bibinfo{author}{Chandler, K.}, \bibinfo{author}{Eaton, A.},
  \bibinfo{author}{Derby, L.}, \bibinfo{author}{Jennings, J.},
  \bibinfo{year}{2014}.
\newblock \bibinfo{title}{Mapping canegrub damage from high spatial resolution
  satellite imagery}, in: \bibinfo{booktitle}{Proceedings of the 36th
  Conference of the Australian Society of Sugar Cane Technologists, ASSCT
  2014}, pp. \bibinfo{pages}{62--70}.
%Type = Article
\bibitem[{Johansen et~al.(2018)Johansen, Sallam, Robson, Samson, Chandler,
  Derby, Eaton and Jennings}]{RN213}
\bibinfo{author}{Johansen, K.}, \bibinfo{author}{Sallam, N.},
  \bibinfo{author}{Robson, A.}, \bibinfo{author}{Samson, P.},
  \bibinfo{author}{Chandler, K.}, \bibinfo{author}{Derby, L.},
  \bibinfo{author}{Eaton, A.}, \bibinfo{author}{Jennings, J.},
  \bibinfo{year}{2018}.
\newblock \bibinfo{title}{Using geoeye-1 imagery for multi-temporal
  object-based detection of canegrub damage in sugarcane fields in queensland,
  australia}.
\newblock \bibinfo{journal}{GIScience \& Remote Sensing} \bibinfo{volume}{55},
  \bibinfo{pages}{285--305}.
\newblock \DOIprefix\doi{10.1080/15481603.2017.1417691}.
%Type = Article
\bibitem[{Jordan(1969)}]{RN77}
\bibinfo{author}{Jordan, C.F.}, \bibinfo{year}{1969}.
\newblock \bibinfo{title}{Derivation of leaf‐area index from quality of light
  on the forest floor}.
\newblock \bibinfo{journal}{Ecology} \bibinfo{volume}{50},
  \bibinfo{pages}{663--666}.
\newblock \DOIprefix\doi{https://doi.org/10.2307/1936256}.
%Type = Article
\bibitem[{Kaufman and Tanre(1992)}]{RN74}
\bibinfo{author}{Kaufman, Y.J.}, \bibinfo{author}{Tanre, D.},
  \bibinfo{year}{1992}.
\newblock \bibinfo{title}{Atmospherically resistant vegetation index (arvi) for
  eos-modis}.
\newblock \bibinfo{journal}{IEEE transactions on Geoscience and Remote Sensing}
  \bibinfo{volume}{30}, \bibinfo{pages}{261--270}.
\newblock \DOIprefix\doi{10.1109/36.134076}.
%Type = Article
\bibitem[{Kussul et~al.(2017)Kussul, Lavreniuk, Skakun and Shelestov}]{RN503}
\bibinfo{author}{Kussul, N.}, \bibinfo{author}{Lavreniuk, M.},
  \bibinfo{author}{Skakun, S.}, \bibinfo{author}{Shelestov, A.},
  \bibinfo{year}{2017}.
\newblock \bibinfo{title}{Deep learning classification of land cover and crop
  types using remote sensing data}.
\newblock \bibinfo{journal}{IEEE Geoscience and Remote Sensing Letters}
  \bibinfo{volume}{14}, \bibinfo{pages}{778--782}.
%Type = Article
\bibitem[{Liu et~al.(2021)Liu, Qian and Yue}]{RN501}
\bibinfo{author}{Liu, Y.}, \bibinfo{author}{Qian, J.}, \bibinfo{author}{Yue,
  H.}, \bibinfo{year}{2021}.
\newblock \bibinfo{title}{Comprehensive evaluation of sentinel-2 red edge and
  shortwave-infrared bands to estimate soil moisture}.
\newblock \bibinfo{journal}{IEEE Journal of Selected Topics in Applied Earth
  Observations and Remote Sensing} \bibinfo{volume}{14},
  \bibinfo{pages}{7448--7465}.
\newblock \DOIprefix\doi{10.1109/JSTARS.2021.3098513}.
%Type = Article
\bibitem[{Lu et~al.(2020)Lu, Dao, Liu, He and Shang}]{RN560}
\bibinfo{author}{Lu, B.}, \bibinfo{author}{Dao, P.D.}, \bibinfo{author}{Liu,
  J.}, \bibinfo{author}{He, Y.}, \bibinfo{author}{Shang, J.},
  \bibinfo{year}{2020}.
\newblock \bibinfo{title}{Recent advances of hyperspectral imaging technology
  and applications in agriculture}.
\newblock \bibinfo{journal}{Remote Sensing} \bibinfo{volume}{12},
  \bibinfo{pages}{2659}.
\newblock \DOIprefix\doi{https://doi.org/10.3390/rs12162659}.
%Type = Article
\bibitem[{Magarey et~al.(2021)Magarey, McHardie, Hession, Cripps, Burgess,
  Spannagle, Sutherland, Di~Bella, Milla, Millar, Schembri, Baxter,
  Hetherington, Turner, Jakins, Quinn, Kalkhoran, Gibbs and Ngo}]{RN12}
\bibinfo{author}{Magarey, R.}, \bibinfo{author}{McHardie, R.},
  \bibinfo{author}{Hession, M.}, \bibinfo{author}{Cripps, G.},
  \bibinfo{author}{Burgess, D.}, \bibinfo{author}{Spannagle, B.},
  \bibinfo{author}{Sutherland, P.}, \bibinfo{author}{Di~Bella, L.},
  \bibinfo{author}{Milla, R.}, \bibinfo{author}{Millar, F.},
  \bibinfo{author}{Schembri, A.}, \bibinfo{author}{Baxter, D.},
  \bibinfo{author}{Hetherington, M.}, \bibinfo{author}{Turner, M.},
  \bibinfo{author}{Jakins, A.}, \bibinfo{author}{Quinn, B.},
  \bibinfo{author}{Kalkhoran, S.}, \bibinfo{author}{Gibbs, L.},
  \bibinfo{author}{Ngo, C.}, \bibinfo{year}{2021}.
\newblock \bibinfo{title}{Incidence and economic effects of ratoon stunting
  disease on the queensland sugarcane industry : Assct peer-reviewed paper}.
\newblock \bibinfo{journal}{Proceedings of the Australian Society of Sugar Cane
  Technologists} \bibinfo{volume}{volume 42}, \bibinfo{pages}{520--526}.
%Type = Article
\bibitem[{McFeeters(1996)}]{RN78}
\bibinfo{author}{McFeeters, S.K.}, \bibinfo{year}{1996}.
\newblock \bibinfo{title}{The use of the normalized difference water index
  (ndwi) in the delineation of open water features}.
\newblock \bibinfo{journal}{International Journal of Remote Sensing}
  \bibinfo{volume}{17}, \bibinfo{pages}{1425--1432}.
\newblock \DOIprefix\doi{https://doi.org/10.1080/01431169608948714}.
%Type = Article
\bibitem[{Merzlyak et~al.(1999)Merzlyak, Gitelson, Chivkunova and
  Rakitin}]{RN76}
\bibinfo{author}{Merzlyak, M.N.}, \bibinfo{author}{Gitelson, A.A.},
  \bibinfo{author}{Chivkunova, O.B.}, \bibinfo{author}{Rakitin, V.Y.},
  \bibinfo{year}{1999}.
\newblock \bibinfo{title}{Non‐destructive optical detection of pigment
  changes during leaf senescence and fruit ripening}.
\newblock \bibinfo{journal}{Physiologia plantarum} \bibinfo{volume}{106},
  \bibinfo{pages}{135--141}.
\newblock \DOIprefix\doi{https://doi.org/10.1034/j.1399-3054.1999.106119.x}.
%Type = Article
\bibitem[{Moriya et~al.(2017)Moriya, Imai, Tommaselli and Miyoshi}]{RN92}
\bibinfo{author}{Moriya, E.A.S.}, \bibinfo{author}{Imai, N.N.},
  \bibinfo{author}{Tommaselli, A.M.G.}, \bibinfo{author}{Miyoshi, G.T.},
  \bibinfo{year}{2017}.
\newblock \bibinfo{title}{Mapping mosaic virus in sugarcane based on
  hyperspectral images}.
\newblock \bibinfo{journal}{IEEE Journal of Selected Topics in Applied Earth
  Observations and Remote Sensing} \bibinfo{volume}{10},
  \bibinfo{pages}{740--748}.
\newblock \DOIprefix\doi{10.1109/JSTARS.2016.2635482}.
%Type = Book
\bibitem[{Murphy(2022)}]{MPROB}
\bibinfo{author}{Murphy, K.}, \bibinfo{year}{2022}.
\newblock \bibinfo{title}{Probabilistic Machine Learning: An Introduction}.
\newblock Adaptive Computation and Machine Learning series,
  \bibinfo{publisher}{MIT Press}.
\newblock \URLprefix \url{https://books.google.com.au/books?id=wrZNEAAAQBAJ}.
%Type = Article
\bibitem[{Narmilan et~al.(2022)Narmilan, Gonzalez, Salgadoe and Powell}]{RN20}
\bibinfo{author}{Narmilan, A.}, \bibinfo{author}{Gonzalez, F.},
  \bibinfo{author}{Salgadoe, A.S.A.}, \bibinfo{author}{Powell, K.},
  \bibinfo{year}{2022}.
\newblock \bibinfo{title}{Detection of white leaf disease in sugarcane using
  machine learning techniques over uav multispectral images}.
\newblock \bibinfo{journal}{Drones} \bibinfo{volume}{6}, \bibinfo{pages}{230}.
\newblock \DOIprefix\doi{https://doi.org/10.3390/drones6090230}.
%Type = Article
\bibitem[{Ong et~al.(2023)Ong, Jian, Li, Zou, Yin and Ma}]{RN225}
\bibinfo{author}{Ong, P.}, \bibinfo{author}{Jian, J.}, \bibinfo{author}{Li,
  X.}, \bibinfo{author}{Zou, C.}, \bibinfo{author}{Yin, J.},
  \bibinfo{author}{Ma, G.}, \bibinfo{year}{2023}.
\newblock \bibinfo{title}{New approach for sugarcane disease recognition
  through visible and near-infrared spectroscopy and a modified wavelength
  selection method using machine learning models}.
\newblock \bibinfo{journal}{Spectrochimica Acta Part A: Molecular and
  Biomolecular Spectroscopy} \bibinfo{volume}{302}, \bibinfo{pages}{123037}.
\newblock \DOIprefix\doi{https://doi.org/10.1016/j.saa.2023.123037}.
%Type = Article
\bibitem[{Qiu et~al.(2017)Qiu, He, Yin and Liao}]{RN504}
\bibinfo{author}{Qiu, S.}, \bibinfo{author}{He, B.}, \bibinfo{author}{Yin, C.},
  \bibinfo{author}{Liao, Z.}, \bibinfo{year}{2017}.
\newblock \bibinfo{title}{Assessments of sentinel-2 vegetation red-edge
  spectral bands for improving land cover classification}.
\newblock \bibinfo{journal}{The International Archives of the Photogrammetry,
  Remote Sensing and Spatial Information Sciences} \bibinfo{volume}{XLII-2/W7},
  \bibinfo{pages}{871--874}.
\newblock \URLprefix
  \url{https://isprs-archives.copernicus.org/articles/XLII-2-W7/871/2017/},
  \DOIprefix\doi{10.5194/isprs-archives-XLII-2-W7-871-2017}.
%Type = Techreport
\bibitem[{Rouse~Jr et~al.(1973)Rouse~Jr, Haas, Schell and Deering}]{RN73}
\bibinfo{author}{Rouse~Jr, J.W.}, \bibinfo{author}{Haas, R.H.},
  \bibinfo{author}{Schell, J.}, \bibinfo{author}{Deering, D.},
  \bibinfo{year}{1973}.
\newblock \bibinfo{title}{Monitoring the vernal advancement and retrogradation
  (green wave effect) of natural vegetation}.
\newblock \bibinfo{type}{Report}. Remote Sensing Center Texas A\&M University.
%Type = Manual
\bibitem[{{Scikit-learn Developers}(2024)}]{HGS}
\bibinfo{author}{{Scikit-learn Developers}}, \bibinfo{year}{2024}.
\newblock \bibinfo{title}{HalvingGridSearchCV}.
\newblock \URLprefix
  \url{https://scikit-learn.org/stable/modules/generated/sklearn.model_selection.HalvingGridSearchCV.html}.
  \bibinfo{note}{accessed: 2024-09-26}.
%Type = Article
\bibitem[{Segarra et~al.(2020)Segarra, Buchaillot, Araus and Kefauver}]{stress}
\bibinfo{author}{Segarra, J.}, \bibinfo{author}{Buchaillot, M.L.},
  \bibinfo{author}{Araus, J.L.}, \bibinfo{author}{Kefauver, S.C.},
  \bibinfo{year}{2020}.
\newblock \bibinfo{title}{Remote sensing for precision agriculture: Sentinel-2
  improved features and applications}.
\newblock \bibinfo{journal}{Agronomy} \bibinfo{volume}{10},
  \bibinfo{pages}{641}.
%Type = Article
\bibitem[{Simões and {Rios do Amaral}(2023)}]{RN224}
\bibinfo{author}{Simões, I.O.}, \bibinfo{author}{{Rios do Amaral}, L.},
  \bibinfo{year}{2023}.
\newblock \bibinfo{title}{Uav-based multispectral data for sugarcane resistance
  phenotyping of orange and brown rust}.
\newblock \bibinfo{journal}{Smart Agricultural Technology} \bibinfo{volume}{4},
  \bibinfo{pages}{100144}.
\newblock \DOIprefix\doi{https://doi.org/10.1016/j.atech.2022.100144}.
%Type = Article
\bibitem[{Soca-Mu\~{n}oz et~al.(2020)Soca-Mu\~{n}oz, Rodr\`{i}guez-Machado,
  Aday-D\`{i}az, Hern\`{a}ndez-Santana and Orozco-Morales}]{RN216}
\bibinfo{author}{Soca-Mu\~{n}oz, J.L.}, \bibinfo{author}{Rodr\`{i}guez-Machado,
  E.}, \bibinfo{author}{Aday-D\`{i}az, O.},
  \bibinfo{author}{Hern\`{a}ndez-Santana, L.}, \bibinfo{author}{Orozco-Morales,
  R.}, \bibinfo{year}{2020}.
\newblock \bibinfo{title}{Spectral signature of brown rust and orange rust in
  sugarcane}.
\newblock \bibinfo{journal}{Revista Facultad de Ingenier\~A\-a Universidad de
  Antioquia} , \bibinfo{pages}{9 --
  20}\DOIprefix\doi{10.17533/udea.redin.20191042}.
%Type = Book
\bibitem[{{Sugar Research Australia}(2021)}]{RSD4}
\bibinfo{author}{{Sugar Research Australia}}, \bibinfo{year}{2021}.
\newblock \bibinfo{title}{Ratoon Stunting Disease}.
\newblock \bibinfo{publisher}{Sugar Research Australia}.
\newblock \URLprefix
  \url{https://sugarresearch.com.au/sugar_files/2017/03/RSD-Info-Sheet_2021_May-2021.pdf}.
%Type = Book
\bibitem[{{Sugar Research Australia}(2024)}]{RSD5}
\bibinfo{author}{{Sugar Research Australia}}, \bibinfo{year}{2024}.
\newblock \bibinfo{title}{Variety Guide}.
\newblock \bibinfo{publisher}{Sugar Research Australia}.
\newblock \URLprefix
  \url{https://sugarresearch.com.au/sugar_files/2024/06/SRA_Variety-Guide-2024-25-Herbert.pdf}.
%Type = Inbook
\bibitem[{Ting(2010)}]{Ting2010}
\bibinfo{author}{Ting, K.M.}, \bibinfo{year}{2010}.
\newblock \bibinfo{title}{Encyclopedia of Machine Learning}.
  \bibinfo{publisher}{Springer US}, \bibinfo{address}{Boston, MA}. chapter
  \bibinfo{chapter}{Precision and Recall}.
\newblock pp. \bibinfo{pages}{781--781}.
\newblock \URLprefix \url{https://doi.org/10.1007/978-0-387-30164-8_652},
  \DOIprefix\doi{10.1007/978-0-387-30164-8_652}.
%Type = Article
\bibitem[{Tucker(1979)}]{RN80}
\bibinfo{author}{Tucker, C.J.}, \bibinfo{year}{1979}.
\newblock \bibinfo{title}{Red and photographic infrared linear combinations for
  monitoring vegetation}.
\newblock \bibinfo{journal}{Remote Sensing of Environment} \bibinfo{volume}{8},
  \bibinfo{pages}{127--150}.
\newblock \DOIprefix\doi{https://doi.org/10.1016/0034-4257(79)90013-0}.
%Type = Article
\bibitem[{Vargas et~al.(2016)Vargas, Mendoza, Gómez, Rivero and
  Espinosa}]{RN93}
\bibinfo{author}{Vargas, L.A.O.}, \bibinfo{author}{Mendoza, G.G.},
  \bibinfo{author}{Gómez, R.A.}, \bibinfo{author}{Rivero, N.A.},
  \bibinfo{author}{Espinosa, L.Y.}, \bibinfo{year}{2016}.
\newblock \bibinfo{title}{Characterization of diatraea saccharalis in sugarcane
  (saccharum officinarum) with field spectroradiometry}.
\newblock \bibinfo{journal}{International Journal of Environmental \&
  Agriculture Research (IJOEAR)} .
%Type = Article
\bibitem[{Wang et~al.(2017)Wang, Chen, Wu, Tang, Shi, Black and Zhu}]{RN502}
\bibinfo{author}{Wang, C.}, \bibinfo{author}{Chen, J.}, \bibinfo{author}{Wu,
  J.}, \bibinfo{author}{Tang, Y.}, \bibinfo{author}{Shi, P.},
  \bibinfo{author}{Black, T.A.}, \bibinfo{author}{Zhu, K.},
  \bibinfo{year}{2017}.
\newblock \bibinfo{title}{A snow-free vegetation index for improved monitoring
  of vegetation spring green-up date in deciduous ecosystems}.
\newblock \bibinfo{journal}{Remote Sensing of Environment}
  \bibinfo{volume}{196}, \bibinfo{pages}{1--12}.
\newblock \URLprefix
  \url{https://www.sciencedirect.com/science/article/pii/S0034425717301906},
  \DOIprefix\doi{https://doi.org/10.1016/j.rse.2017.04.031}.
%Type = Article
\bibitem[{Waters et~al.(2025)Waters, Chen and {Rahimi Azghadi}}]{Waters}
\bibinfo{author}{Waters, E.K.}, \bibinfo{author}{Chen, C.C.M.},
  \bibinfo{author}{{Rahimi Azghadi}, M.}, \bibinfo{year}{2025}.
\newblock \bibinfo{title}{Sugarcane health monitoring with satellite
  spectroscopy and machine learning: A review}.
\newblock \bibinfo{journal}{Computers and Electronics in Agriculture}
  \bibinfo{volume}{229}, \bibinfo{pages}{109686}.
\newblock \URLprefix
  \url{https://www.sciencedirect.com/science/article/pii/S0168169924010779},
  \DOIprefix\doi{https://doi.org/10.1016/j.compag.2024.109686}.
%Type = Article
\bibitem[{Xue and Su(2017)}]{RN45}
\bibinfo{author}{Xue, J.}, \bibinfo{author}{Su, B.}, \bibinfo{year}{2017}.
\newblock \bibinfo{title}{Significant remote sensing vegetation indices: A
  review of developments and applications}.
\newblock \bibinfo{journal}{Journal of Sensors} \bibinfo{volume}{2017},
  \bibinfo{pages}{1353691}.
\newblock \DOIprefix\doi{10.1155/2017/1353691}.
%Type = Article
\bibitem[{Young et~al.(2016)Young, Kawamata, Ensbey, Lambley and Nock}]{RN475}
\bibinfo{author}{Young, A.J.}, \bibinfo{author}{Kawamata, A.},
  \bibinfo{author}{Ensbey, M.A.}, \bibinfo{author}{Lambley, E.},
  \bibinfo{author}{Nock, C.J.}, \bibinfo{year}{2016}.
\newblock \bibinfo{title}{Efficient diagnosis of ratoon stunting disease of
  sugarcane by quantitative pcr on pooled leaf sheath biopsies}.
\newblock \bibinfo{journal}{Plant Disease} \bibinfo{volume}{100},
  \bibinfo{pages}{2492--2498}.
\newblock \DOIprefix\doi{10.1094/PDIS-06-16-0848-RE}. \bibinfo{note}{pMID:
  30686165}.

\end{thebibliography}

\end{document}